\RequirePackage{fix-cm}
\documentclass[twocolumn]{svjour3}    
\makeatletter\def\makeheadbox{{\hbox to0pt{\vbox{\baselineskip=10dd\hrule\hbox to\hsize{\vrule\kern3pt\vbox{\kern3pt\hbox{\bfseries Submitted to International Journal of Computer Vision, December 2025}\kern3pt}\hfil\kern3pt\vrule}\hrule}\hss}}}

\usepackage{graphicx}
\usepackage{natbib}
\usepackage{url}            
\usepackage{booktabs}       
\usepackage{amsfonts}       
\usepackage{nicefrac}       
\usepackage{microtype}      
\usepackage{graphicx}
\usepackage{enumitem}
\usepackage{bm}
\usepackage{nccmath}
\usepackage{boldline}
\usepackage{booktabs}        
\usepackage{threeparttable}

\usepackage[noend]{algcompatible}

\usepackage{capt-of}
\usepackage{color,soul}
\usepackage[pagebackref=false,breaklinks=true,colorlinks,bookmarks=false,urlcolor=magenta, citecolor=blue]{hyperref}

\usepackage{times}
\usepackage{epsfig}
\usepackage{multirow}
\usepackage{amsmath}
\usepackage{amssymb}
\usepackage{comment}
\usepackage{pifont}
\usepackage[table]{xcolor}
\usepackage{bm}
\usepackage{soul}

\usepackage{stfloats}
\usepackage{cuted}
\usepackage{capt-of}
\usepackage{graphicx}
\usepackage{booktabs}
\usepackage{enumitem}
\usepackage{amsfonts}
\usepackage{textcomp}

\usepackage[ruled,linesnumbered]{algorithm2e}
\usepackage{makecell}
\usepackage{subfigure}
\usepackage{arydshln}
\usepackage{ulem}

\def\cm#1{\checkmark}

\useunder{\underline}{\ul}{}

\usepackage{tikz}

\begin{document}
\begin{sloppypar}
\title{Fine-grained Fragment Retrieval in Multi-modal Long-form Dialogues}

\author{
Hanbo Bi$^{1,2\star}$ \and 
Zhiqiang Yuan$^{1\star\S}$ \and 
Chongyang Li$^{1,2}$ \and 
Qiwei Yan$^{1,2}$ \and 
Zexi Jia$^{1}$ \and 
Jiapei Zhang$^{1}$ \and   
Xiaoyue Duan$^{1}$ \and  
Yingchao Feng$^{2}$ \and
Jinchao Zhang$^{1\dagger}$ \and
Jie Zhou$^{1}$
}

\institute{
Zhiqiang Yuan (yuanzhiqiang19@mails.ucas.edu.cn) \\ 
1. Pattern Recognition Center, WeChat AI, Tencent Inc, China \\
2. Aerospace Information Research Institute, Chinese Academy of Sciences  \\
$\star$ These authors contributed equally.\\
$\S$ Tech Lead.\\
$\dagger$ Corresponding authors.\\
}


\date{Received: date / Accepted: date}

\maketitle

\begin{abstract}
With the widespread adoption of multi-modal communication platforms, long-form dialogues that interleave text and images have become increasingly common. In real-world applications such as customer support, knowledge management, and collaborative tools, users often seek to revisit coherent dialogue fragments related to specific topics or events, rather than isolated utterances or images. To address this need, we propose the task of Fine-grained Fragment Retrieval (FFR), which requires a model to locate semantically relevant fragments composed of multiple utterances and images within multi-modal long-form dialogues. We explore two core FFR settings: (1) FFR within Single-Dialogue, where the model retrieves fragments from a given long-form dialogue; and (2) FFR within Dialogue Corpus, which involves retrieving relevant fragments from a large-scale dialogue corpus, reflecting practical open-domain retrieval scenarios. To support the former, we introduce F$^2$RVLM, a generation-based retrieval model trained with reinforcement learning, which incorporates multi-objective rewards and difficulty-aware curriculum sampling to enhance the consistency and coherence of retrieved fragments. For the latter, we develop FFRS, a two-stage retrieval system that combines offline fragment-level indexing with online retrieval, enabling efficient and accurate retrieval at scale. Specifically, each long-form dialogue is decomposed into minimal semantic units (i.e., fragments), which are encoded into a vector database by a Fragment Embedding Model (FEM). During inference, FEM rapidly recalls Top-K candidate fragments based on a user query, followed by F$^2$RVLM performing fine-grained reasoning to identify the most relevant sub-content within each candidate. To support and evaluate FFR, we construct MLDR, the longest multi-modal dialogue retrieval dataset to date, along with a real-world test set based on WeChat conversations. Extensive experiments on both benchmarks demonstrate that our proposed F$^2$RVLM and FFRS consistently achieve superior performance in both single-dialogue and corpus-level FFR settings. The code and datasets are publicly available at \href{https://github.com/HanboBizl/FFRS.github.io}{FFRS.github.io}.

\end{abstract}

\section{Introduction}

\begin{figure*}[t]
\setlength{\abovecaptionskip}{1pt}
\centering
\includegraphics[width=1.0\linewidth]{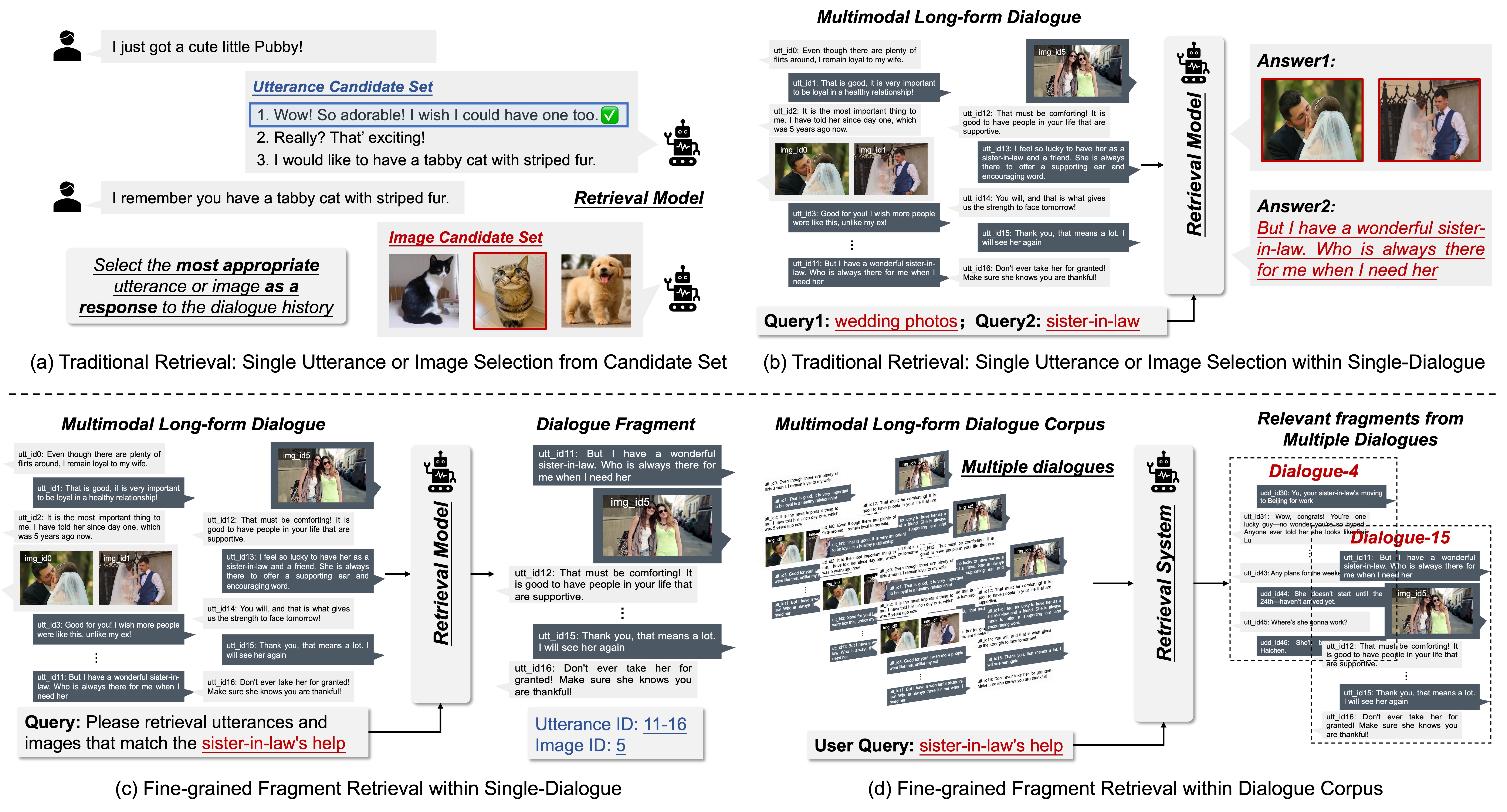}
\caption{\textbf{Comparison of Multi-modal Dialogue Retrieval Paradigms.}
\textbf{(a) Response Selection:} Selecting the most appropriate utterance or image from a predefined candidate pool as a response to the given dialogue history. 
\textbf{(b) Single-item Retrieval:} Retrieving one utterance or image directly from the original multi-modal dialogue context. We introduce the task of \textbf{Fine-grained Fragment Retrieval (FFR)}, which aims to retrieve semantically coherent multi-turn fragments containing both utterances and images, better reflecting real-world information needs in long-form multi-modal conversations.
\textbf{(c) FFR within Single-dialogue:} Given a user query and a multi-modal dialogue, retrieve a single semantically relevant fragment composed of multiple interleaved utterances and images from the same dialogue.
\textbf{(d) FFR within Dialogue Corpus:} Given a user query and a large-scale corpus of dialogues, retrieve multiple relevant fragments from different dialogues, enabling global-level retrieval across conversations and broader content coverage.}
\label{fig:contrast}
\end{figure*}

With the widespread adoption of multi-modal communication platforms such as intelligent customer service systems~\citep{nie2021research,wu2024intelligent}, social messaging applications~\citep{tomar2014maturity,walnycky2015network}, and enterprise-level assistants~\citep{bischoff2002dependable}, massive amounts of dialogue data interwoven with text and images are rapidly accumulating. Efficiently retrieving semantically relevant information from such large-scale multi-modal corpora has become a core capability for enabling knowledge discovery, optimizing user experience, and advancing intelligent human-computer interaction~\citep{linmm}. Compared to traditional text-only dialogues, multi-modal long-form conversations often span dozens or even hundreds of turns, exhibiting strong temporal dependencies, intricate cross-modal semantic interactions, and complex contextual relationships~\citep{chen2024domain, yang2025deep}. These characteristics make retrieving useful information from them far more challenging than from short texts or structured inputs (i.e., documents).

Current paradigms for multi-modal dialogue retrieval generally fall into two categories: (a) selecting the most relevant single utterance or image from a pre-defined candidate set as the dialogue response (see Fig.\ref{fig:contrast}.a)~\citep{yin2024dialclip, bai2025chat}, and (b) directly retrieving an individual utterance or image from the original dialogue context (see Fig.\ref{fig:contrast}.b)~\citep{zang2021photochat}. 
While effective in response prediction and short-context scenarios, existing paradigms struggle in real-world applications. In practice, users rarely aim to retrieve isolated pieces of information; instead, they tend to seek coherent semantic fragments that span multiple dialogue turns and modalities, typically associated with specific topics or events. Examples include requests like “the conversation about the product launch plan from last week” or “the portion involving a particular design image.” These fine-grained information needs underscore a critical gap between existing retrieval paradigms and users’ expectations in navigating complex, long-form multi-modal dialogues.

To fill this gap, we introduce the task of \textbf{Fine-grained Fragment Retrieval (FFR)}, as illustrated in Fig.\ref{fig:contrast}.c-d. In contrast to existing paradigms that focus on retrieving single utterances or isolated images, FFR aims to locate semantically relevant and contextually complete fragments in response to user queries, which typically consist of interleaved text and image turns. By enabling FFR, we seek to transform long-form dialogue content into semantic units that better align with human understanding, thereby significantly improving the efficiency and accuracy of retrieving valuable information from large-scale multi-modal corpora.

As a foundation for studying FFR, we construct MLDR, a large-scale multi-modal long-form dialogue retrieval dataset, where each dialogue covers three distinct topics with an average of 25.45 turns, making it the longest-turn multi-modal dataset to date. To further evaluate generalization in realistic settings, we curate a WeChat-based test set consisting of real-world dialogues with an average of 75.38 turns. Together, these datasets support both controlled benchmarking and open-domain evaluation for fragment retrieval.

Based on these resources, we systematically explore FFR under two representative settings: \textbf{FFR within Single-dialogue}, which evaluates fine-grained fragment retrieval within a single conversation, and \textbf{FFR within Dialogue Corpus}, which reflects large-scale retrieval across multiple dialogues. This dual-setting design not only supports thorough assessment of local and global retrieval performance but also demonstrates strong potential for real-world deployment.

\noindent \textbf{(1) FFR within Single-dialogue.} To develop a retrieval model capable of efficiently performing FFR in multi-modal long-form dialogues, we first explore fragment retrieval within a single dialogue, where the model is expected to locate the most relevant fragment given a user query. To this end, we assess representative vision-language models (VLMs), including embedding models~\citep{radford2021learning, li2023blip} and multi-modal language models (MLLMs)~\citep{jaech2024openai,comanici2025gemini}, utilizing our constructed datasets.
Results reveal that while embedding models offer faster inference, they fall short in semantic precision. Surprisingly, even leading MLLMs, such as Qwen2.5-VL-72B~\citep{wang2024qwen2} and Doubao-Seed-1.6~\citep{guo2025seed1}, frequently retrieve incoherent utterance-image pairs, e.g., mismatched turns or irrelevant visual content, resulting in suboptimal F1 scores in realistic settings. This limitation primarily stems from the gap between the models’ learning objectives and the demands of fragment retrieval: while optimized for generating responses from visual-textual inputs, these models lack explicit supervision to ensure that the retrieved or generated fragments are semantically coherent and contextually aligned with the user query. 

Thus, we introduce \textbf{F$^2$RVLM}, an MLLM-based retrieval model tailored for FFR. F$^2$RVLM follows a two-stage training paradigm: supervised fine-tuning to inject fragment-level retrieval knowledge, followed by GRPO-based reinforcement learning to align retrieval behaviors with human preferences.
We design a multi-objective reward scheme that encourages the generation of fragments with semantic precision and contextual coherence: (i) an F1-based alignment reward encourages accurate matching with ground-truth fragments, penalizing over- and under-retrieval; (ii) a fragment order consistency enhances semantic alignment between selected utterances and images, guiding the model to organize content in a coherent, human-preferred manner. 
Additionally, we incorporate a difficulty-aware curriculum sampling strategy that ranks training samples from easy to hard based on retrieval F1 and prediction entropy, enabling progressive learning of reasoning capabilities in complex, multi-turn conversations. 
Extensive experiments on both the in-domain MLDR and real-domain WeChat-based sets demonstrate that F$^2$RVLM significantly outperforms mainstream VLMs in retrieval accuracy and contextual understanding. 

\noindent \textbf{(2) FFR within Dialogue Corpus.} While F$^2$RVLM demonstrates strong performance in single-dialogue settings, real-world applications often require matching queries against thousands of dialogues. To address this, we further develop \textbf{FFRS,} a two-stage \textbf{F}ine-grained \textbf{F}ragment \textbf{R}etrieval \textbf{S}ystem, that balances semantic precision and retrieval efficiency through an “offline structured indexing and online dual-stage retrieval” paradigm, as illustrated in Fig.\ref{fig:contrast}.d.
\begin{enumerate}
    \item \textbf{Offline Indexing}. 
    We first decompose each dialogue into minimal semantic units, referred to as fragments. Each fragment is then encoded and stored in a vector database, forming an offline-constructed index that supports fast similarity-based retrieval without the need to re-parse the entire dialogue corpus during inference. To ensure high‑quality embeddings, we introduce a Fragment Embedding Model (FEM) trained with a dual-level contrastive learning strategy: (i) at the inter-fragment level, FEM captures global semantic consistency across dialogues; and (ii) at the intra-fragment level, it models fine-grained pragmatic dependencies through multiple Question-Answer (QA) style subtasks constructed within each fragment. This design allows FEM to preserve the semantic integrity of fragment embeddings, thereby supporting accurate and context-aware retrieval.
    \item \textbf{Online Retrieval}. In response to a user query, the system first conducts fast embedding-based retrieval to coarsely recall the Top-K candidate fragments from the offline-constructed index. These candidates are then further refined by F$^2$RVLM, which conducts more precise reasoning to locate sub-content that better matches the query semantics.
\end{enumerate}

By combining embedding-driven coarse recall with generation-driven fine-grained reasoning, FFRS strikes a favorable trade-off between retrieval efficiency and semantic precision. It provides a robust and adaptable framework for fine-grained retrieval across both individual dialogues and large-scale corpora, paving the way for practical deployment of retrieval-oriented multi-modal dialogue systems. Experiments on a real-world WeChat-based corpus further demonstrate that our system achieves efficient and accurate fragment retrieval, highlighting its effectiveness under realistic dialogue scenarios.

Our main contributions are summarized as follows:

\begin{itemize}
\item \textbf{Novel Retrieval Task.} We introduce Fine-grained Fragment Retrieval (FFR), a novel retrieval task that aims to directly locate semantically coherent utterance-image fragments from long-form dialogues, differing from traditional dialogue retrieval that selects the most appropriate individual elements.
\item \textbf{FFR Dataset Construction.} We construct MLDR, the longest-turn multi-modal dialogue retrieval dataset to date, averaging 25.45 turns per dialogue, with each covering three distinct topics. Additionally, we curate a real-world WeChat-based test set averaging 75.38 turns per dialogue to evaluate retrieval generalization in practice.
\item \textbf{FFR within Single-Dialogue.} 
We propose F$^2$RVLM, a reinforcement learning-based retrieval model tailored for single-dialogue FFR. It incorporates multi-objective rewards and difficulty-aware curriculum sampling to progressively enhance the semantic consistency and coherence of retrieved fragments. Experiments on MLDR and WeChat test sets demonstrate that F$^2$RVLM consistently outperforms popular VLMs in retrieval accuracy.
\item \textbf{FFR within Dialogue Corpus.} To scale FFR to a large dialogue corpus, we develop FFRS, a two-stage system that combines embedding-based coarse recall with fine-grained reasoning. Evaluations on real-world WeChat dialogues validate its capability for fast and semantically aligned fragment retrieval in realistic settings.
\end{itemize}

Compared to our original conference paper accepted by AAAI~\citep{bi2025f2rvlm}, which introduced the F$^2$RVLM model for single-dialogue fragment retrieval, this extended work presents four major improvements:
\textbf{(i)} We expand the task scope from single‑dialogue to \textbf{corpus‑level fragment retrieval}, enabling the model to identify semantically consistent fragments across multiple dialogues, an essential step toward scalable real‑world applications;
\textbf{(ii)} We develop FFRS, a \textbf{two-stage retrieval system} that follows the “offline indexing + online retrieval” paradigm, combining embedding-based coarse recall with fine-grained reasoning, thereby overcoming the inefficiency of prior generation-based methods that rely on exhaustive per-dialogue inference;
\textbf{(iii)} To support FFRS, we introduce a \textbf{Fragment Embedding Model (FEM)} that encodes fine-grained multimodal dialogue fragments into dense representations, enabling efficient fragment-level indexing and fast coarse retrieval across large corpora;
\textbf{(iv)} We integrate F$^2$RVLM into FFRS as the \textbf{fine-grained reasoning module}, forming a unified framework that supports both intra- and inter-dialogue retrieval, thereby enhancing its applicability to multimodal personal assistants and knowledge-centric retrieval systems.


\section{Related Work}
\subsection{Multi-modal Dialogue Datasets}
Recent advances in vision-language modeling have accelerated progress in multi-modal dialogue understanding~\citep{meng2020openvidial}. Existing dialogue datasets fall into two main types: (1) Image-grounded datasets~\citep{mostafazadeh2017image, shuster2020image, zheng2022mmchat,lin2023tiktalk} consist of dialogues explicitly constructed around a given image, often collected via crowdsourcing. While well-aligned, they lack the natural heterogeneity of real conversations, where not all utterances refer to images. (2) Image-sharing datasets~\citep{zang2021photochat, lee2021constructing,feng2023mmdialog,lee2024dialogcc} address this by capturing more spontaneous visual usage. For instance, MMDialog~\citep{feng2023mmdialog} collects over 1M social media conversations with images dispersed across turns, while DialogCC~\citep{lee2024dialogcc} enriches textual dialogues by inserting suitable images. However, these corpora still consist mostly of short, single-topic dialogues and lack the long-range, multi-topic structure of real-world interactions. To address this gap, we construct a novel large-scale dataset designed to capture multi-topic transitions and long-range dependencies in multi-modal dialogues, thereby providing a more realistic and challenging testbed for fine-grained retrieval and contextual understanding.

\subsection{Vision Language Models for Retrieval}
Vision-language models (VLMs) have made substantial progress in cross-modal alignment, especially in image-text retrieval. Current research can be broadly categorized into three paradigms: dual-encoder embedding models, multi-modal large language models (MLLMs), and LLM-based embedding models.

\noindent \textbf{Dual-Encoder Embedding Models.}
These models adopt separate encoders for visual and textual inputs and learn a joint embedding space to support scalable similarity-based retrieval. Representative models include CLIP~\citep{radford2021learning}, trained on 400M image-text pairs via contrastive learning, and ALIGN~\citep{jia2021scaling}, which scales to 1.8B noisy pairs to improve multilingual generalization. Extensions such as UniIR~\citep{wei2024uniir} and UniVL-DR~\citep{liuuniversal} adapt this framework to various retrieval tasks. However, these models are primarily optimized for short, context-independent queries. When applied to semantically rich and long-form multi-modal scenarios, such as dialogue-level retrieval, they often struggle to preserve coherence and context alignment.

\noindent \textbf{Multi-modal LLMs.}
MLLMs~\citep{liu2023visual, achiam2023gpt} extend large language models (LLMs) with visual encoders, enabling them to process multi-modal inputs and generate context-aware responses. Recent works~\citep{wang2024qwen2, jaech2024openai, comanici2025gemini, guo2025seed1, chen2024internvl} demonstrate strong performance in tasks that require reasoning across modalities and dialogue history. These models are especially suitable for complex settings involving long-form conversations, user-specific intents, and implicit cross-turn semantics. Their generative nature enables fine-grained, query-driven interpretation, but also poses challenges in inference latency during large-scale retrieval.

\noindent \textbf{LLM-based Embedding Models.}
A recent trend explores using large language models (LLMs) to directly generate embeddings for retrieval tasks. In the text retrieval domain, studies~\citep{su2023one, wang2024improving, meng2024sfr, behnamghader2024llm2vec} have shown that combining instruction tuning with LLMs enables unified representations for text and document retrieval. These models typically process user input through LLMs and extract the hidden states from the final Transformer layer, using either mean pooling or selecting the last token as the final embedding. Inspired by these advances, researchers have extended this approach to the multi-modal setting. Specifically, recent works~\citep{meng2025vlm2vec, zhang2024gme, jiang2024vlm2vec, liu2025lamra} leverage MLLMs to produce unified representations across different input modalities. For example, VLM2Vec-v2~\citep{meng2025vlm2vec} supports embedding generation for text, image, video, and document inputs in a unified manner, while GME~\citep{zhang2024gme} enables retrieval across unimodal, cross-modal, and hybrid-modal scenarios. 

This work adopts a two-stage retrieval framework to address fine-grained fragment retrieval in large-scale dialogue corpora. Specifically, LLM-based embedding model are utilized for fast coarse retrieval, followed by MLLM-based query-guided reasoning to precisely locate relevant content.

\subsection{MLLMs with Reinforcement Learning}
Recent advances such as OpenAI's o1~\citep{jaech2024openai} and DeepSeek-R1~\citep{guo2025deepseek} have shifted the focus of large language models (LLMs) toward enhancing reasoning capabilities through reinforcement learning (RL), giving rise to novel RL paradigms like GRPO~\citep{shao2024deepseekmath}, DAPO~\citep{yu2025dapo}, and VAPO~\citep{yue2025vapo}. This trend is rapidly extending to the Multi-modal domain~\citep{shen2025vlm,wang2025vl}. LMM-R1~\citep{peng2025lmm} proposes a rule-based two-stage RL framework to enhance multi-modal reasoning. Reason-RFT~\citep{tan2025reason} leverages supervised fine-tuning (SFT) and chain-of-thought (CoT) data to initialize the RL process. Vision-R1~\citep{huang2025vision} adopts progressive reflection suppression during GRPO training to mitigate overthinking after cold starts. Visual-RFT~\citep{liu2025visual} combines GRPO with verifiable reward design to improve visual reasoning, while Video-R1~\citep{feng2025video} further extends this line by introducing T-GRPO to enhance temporal understanding in video-language tasks. In this work, we extend the application of RL to multi-modal dialogue content retrieval, aiming to improve fine-grained reasoning and fragment localization over long-form image-text contexts.

\section{Task Definition}

\noindent \textbf{FFR within Single-Dialogue.}
Fine-grained fragment retrieval (FFR) can be formulated as a structured prediction task over long-form, multi-turn dialogues that include both textual utterances and images. Given a multi-modal dialogue $D = \{(u_1, m_1), (u_2, m_2), \dots, (u_T, m_T)\}$, where each turn $t$ consists of a speaker $u_t$ and a message $m_t$ that may contain text or images, and a user-issued query $q$, the objective is to retrieve a subset of utterance and image IDs that are semantically relevant to the query. To enable the VLMs to perceive and understand the semantic and temporal structure of each turn, explicit structural markers are inserted into the dialogue: \texttt{<|utt\_id\_start|>}...\texttt{<|utt\_id\_end|>} to mark each utterance with a unique ID, while \texttt{<|img\_id\_start|>}...\texttt{<|img\_id\_end|>} marks each embedded image. The retrieval process can be formalized as:
\begin{align} \label{equation:2}
\hat{Y} = \mathcal{F}(D, q) = (\hat{I}_{\text{utt}}, \hat{I}_{\text{img}})
\end{align}
where $\hat{I}_{\text{utt}}$ and $\hat{I}_{\text{img}}$ denote the predicted sets of relevant utterance and image IDs, respectively.

\noindent \textbf{FFR within Dialogue Corpus.}
In the corpus-level setting, FFR aims to identify all fragments across a large-scale multi-dialogue collection that are semantically relevant to a user-issued query. Formally, given a query $q$ and a corpus of long-form multi-modal dialogues $\mathcal{D} = \{D_1, D_2, \dots, D_N\}$,
where each dialogue $D_i = \{(u_1^{i}, m_1^{i}), \dots, (u_{T_i}^{i}, m_{T_i}^{i})\}$ consists of turns with speaker $u_t^{i}$ and message $m_t^{i}$ (which may contain text and/or images), the goal is to retrieve a set of fragment-level identifiers that are semantically aligned with $q$, regardless of their source dialogue. The task can be formalized as:
\begin{align} \label{equation:2_1}
\hat{Y} = \mathcal{F}(\mathcal{D}, q) = \left\{ (\hat{d}^{j}, \hat{t}^{j}) \mid j = 1, \dots, M \right\}
\end{align}
where each prediction $(\hat{d}^j, \hat{t}^j)$ denotes the dialogue ID and turn ID of the $j$-th retrieved fragment, and $M$ is the total number of predicted relevant fragments.

Unlike single-dialogue retrieval, which focuses on local reasoning within a given context, the corpus-level task requires global fragment-level matching across heterogeneous dialogues. This setting presents greater challenges in scalability, semantic sparsity, and noise robustness, as relevant fragments may be sparsely distributed across distant dialogues with diverse styles and topics.

\begin{figure*}[t]
\setlength{\abovecaptionskip}{1pt}
\centering
\includegraphics[width=1.0\linewidth]{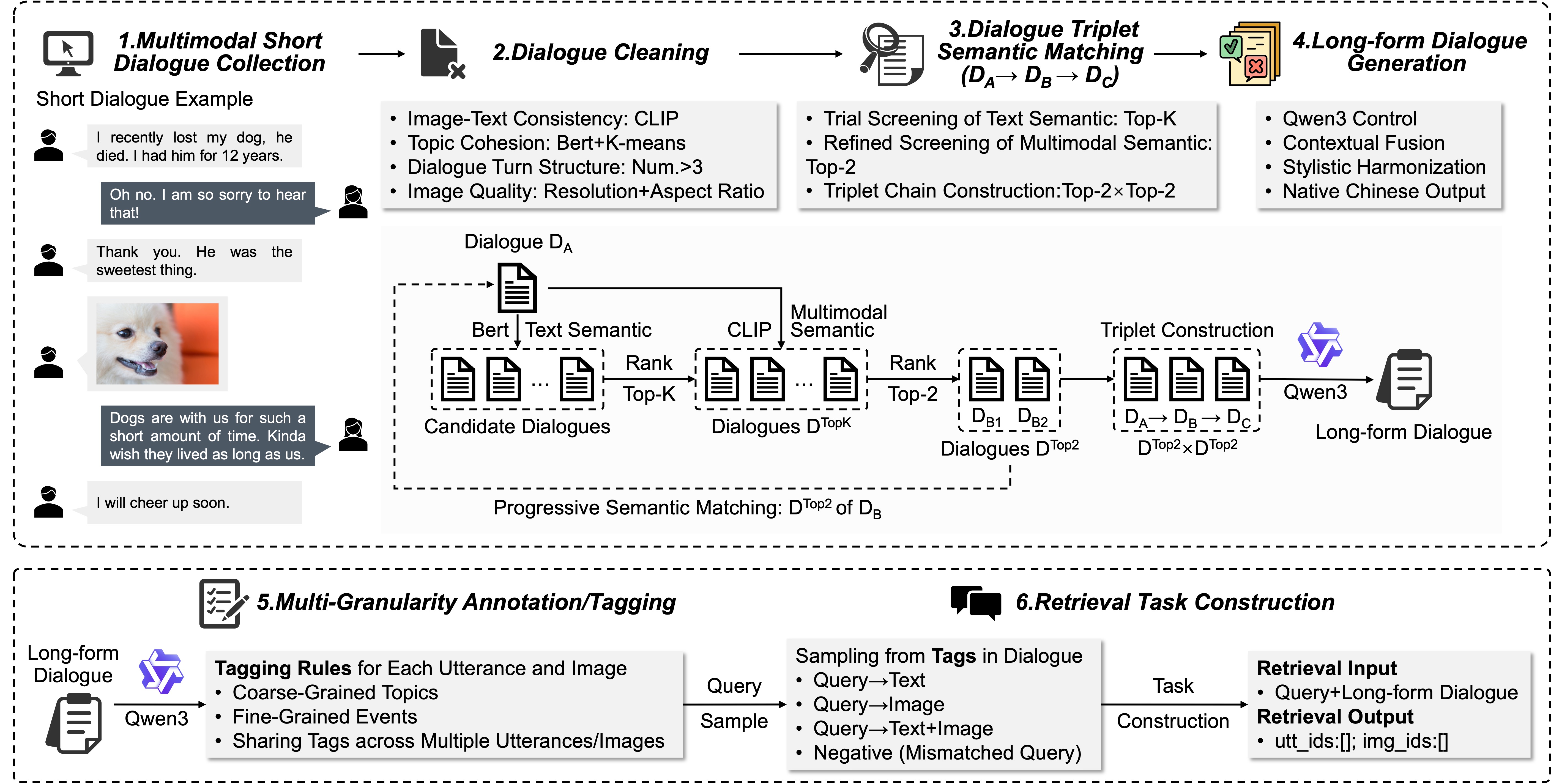}
\caption{\textbf{Overview of the MLDR Construction Pipeline.} It begins with the collection of multi-modal short dialogues, followed by dialogue cleaning, semantic matching, and triplet construction. Long-form dialogue generation is then performed utilizing Qwen3, followed by multi-granularity annotation for semantic tagging. Finally, diverse query types are sampled to formulate the fine-grained retrieval task.}
\label{fig:1}
\end{figure*}

\section{Multi-modal Long-form Dialogue Dataset}

To support research on the FFR task, this work constructs a large-scale multi-modal long-form dialogue retrieval dataset, MLDR, along with a real-world test set based on WeChat conversations. This section outlines the construction process and statistical analysis of both MLDR and WeChat test set.

\subsection{MLDR Construction}
As highlighted in Table~\ref{tab:1}, existing multi-modal dialogue datasets primarily consist of short, single-topic conversations, limiting their utility in modeling the multi-topic nature of real-world dialogues. To fill this gap, we construct a multi-modal long-form dialogue corpus and further develop the MLDR dataset. The pipeline is illustrated in Fig.\ref{fig:1} and summarized below (see Appendix.\ref{MLDR Construction} for details):

\noindent \textbf{Short Dialogue Collection \& Cleaning.} 
We begin with the DialogCC dataset~\citep{lee2024dialogcc}, which contains high-quality short dialogues sourced from various online platforms, covering a wide range of topics such as daily life and Wikipedia-style discussions. These dialogues provide semantically diverse and realistic foundations for generating coherent long-form conversations. To ensure data quality, we apply a multi-criteria cleaning pipeline to remove samples with poor image-text alignment, topic drift, unstructured turns, or low image quality.

\noindent \textbf{Dialogue Triplet Semantic Matching.}
To construct coherent long-form dialogues, we design a triplet-based matching strategy that ensures topic continuity and multi-modal alignment. For each short dialogue $D_A$, we first identify the Top-K semantically similar candidates using BERT-based sentence embeddings. We then refine this set by computing a weighted multi-modal similarity score with CLIP (0.7 for text, 0.3 for image), and select the Top-2 most aligned dialogues $D_B$. This process is repeated for each $D_B$ to retrieve two additional candidates $D_C$, forming four unique triplets of the form $D_A \rightarrow D_B \rightarrow D_C$, which serve as structural units for long-form synthesis.
\begin{align} \label{equation:1}
D_A \rightarrow D_B^i \rightarrow D_C^{i,j}, \quad i = 1,2; \quad j = 1,2
\end{align}

These triplets provide a structural foundation for generating extended dialogues. Notably, each dialogue \( D_A \), \( D_B \), and \( D_C \) in a triplet must be mutually disjoint in terms of both dialogue ID and image content, ensuring that each triplet represents unique and non-duplicate information.

\begin{table}[t]
\setlength{\abovecaptionskip}{1pt}
\setlength{\belowcaptionskip}{1pt}
\centering
\caption{Summary of main multi-modal dialogue datasets.} 
\renewcommand\arraystretch{1.25}
\resizebox{0.95\linewidth}{!}{
\begin{tabular}{cccc} 
\toprule
\textbf{Language}    & \textbf{Datasets}                                                      & \textbf{Avg. Turn}                                 & \textbf{Avg. Topic}                                \\ 
\hline
\multirow{6}{*}{ENG} & ImageChat~\citep{shuster2020image}                                                                                                    & 1.98~                                              & 1.00~                                              \\
& OpenViDial~\citep{meng2020openvidial}                                                                                      & 1.00~                                              & 1.00~                                              \\
& PhotoChat~\citep{zang2021photochat}                                                                                          & 12.74~                                             & 1.00~                                              \\
& MMDD~\citep{lee2021constructing}                                                                                                 & 11.56~                                             & 1.00~                                              \\
& MMDialog~\citep{feng2023mmdialog}                                                                                            & 4.56~                                              & 1.00~                                              \\
& DialogCC~\citep{lee2024dialogcc}                                                                                        & 8.20~                                              & 1.00~                                              \\ 
\hline
\multirow{6}{*}{CN} & MMChat~\citep{zheng2022mmchat}                                                                                               & 2.59~                                              & 1.00~                                              \\
& M3ED~\citep{zhao2022m3ed}                                                                                                       & 9.17~                                              & 1.00~                                              \\
& CPED~\citep{chen2022cped}                                                                                                 & 11.08~                                             & 1.00~                                              \\
& CMMA~\citep{zhang2023cmma}                                                                                                  & 7.27~                                              & 1.00~                                              \\
& TikTalk~\citep{lin2023tiktalk}                                                                                               & 2.25~                                              & 1.00~                                              \\
& {\cellcolor[rgb]{0.951,0.951,0.951}}\textbf{\textbf{MLDR (Ours)}} & {\cellcolor[rgb]{0.951,0.951,0.951}}\textbf{25.45} & {\cellcolor[rgb]{0.951,0.951,0.951}}\textbf{3.00}  \\
\bottomrule
\end{tabular}
}
\label{tab:1}
\end{table}

\noindent \textbf{Long-form Dialogue Generation.} 
Each matched triplet is converted into a coherent long-form dialogue utilizing the Qwen3-235B~\citep{yang2025qwen3} model with structured prompts. The generation process preserves multimodal semantics, ensures topical coherence through smooth transitions, and yields fluent, contextually grounded long-form dialogues, forming high-quality samples for downstream retrieval and reasoning tasks.

\noindent \textbf{Multi-Granularity Annotation.} 
To support fine-grained fragment retrieval, we adopt a two-level shared tagging scheme automatically generated by the Qwen3 under structured prompts. Each sentence and image caption is assigned a coarse-grained tag (e.g., domain) and a fine-grained tag (e.g., event), with each tag shared by at least two elements to form semantically consistent fragments. The tagging process is guided by explicit alignment rules, enabling efficient semantic disentanglement across multimodal dialogues 

\noindent \textbf{Task Construction for Retrieval.} 
We formulate a unified retrieval task over annotated long-form dialogues, where sampled coarse- or fine-grained tags serve as natural language queries, and corresponding utterances/images sharing the same tag form the retrieval fragments. To enhance diversity and robustness, we adopt a query-driven sampling strategy that generates four types of query-fragment pairs: (i) multimodal fragments; (ii) utterance-only fragments; (iii) image-only fragments; and (iv) negative samples constructed by replacing queries with unrelated tags. This formulation supports generalization across varied retrieval scenarios.

\subsection{Real-domain Evaluation on WeChat Dialogues}
To evaluate fine-grained retrieval in real-world scenarios, we construct a real-domain test set from naturally occurring multi-modal WeChat conversations. Unlike the synthesized MLDR data, these dialogues reflect authentic user interactions, with noisy inputs, informal expressions, and frequent topic shifts, posing greater challenges for robust retrieval.

We collect multi-turn image-utterance chats from 12 volunteers\footnotemark\footnotetext{Volunteers manually provided their own chat records from the application. We emphasize that all data were solely provided by volunteers, and no chat records were, or will ever be, obtained from WeChat's backend. More details are provided in Appendix.\ref{WeChat-Based Dataset Construction}.} and preprocess them by removing emojis, sensitive content, profanity, and semantically void utterances. Consecutive messages from the same speaker are merged, and only dialogues with at least two images are retained. This yields 270 long-form dialogues (avg. 145.38 turns), which are segmented if exceeding 100 turns, resulting in 580 coherent samples. All dialogues are manually annotated by a professionally trained team\footnotemark\footnotetext{All dialogues were annotated by 10 trained annotators following standardized guidelines on task definitions and labeling rules. Each dialogue was independently labeled and iteratively verified through multi-round consistency checks, ensuring the accuracy and reliability of the final benchmark.}, producing 1,250 query-dialogue pairs. Each sample includes a natural language query, its multi-modal dialogue (avg. 75.38 turns), and ground-truth utterance and image IDs as retrieval targets. The task format remains consistent with MLDR, enabling direct evaluation of model generalization in open-domain fragment retrieval.

\begin{figure}[t]
\setlength{\abovecaptionskip}{1pt}
\centering
\includegraphics[width=1.0\linewidth]{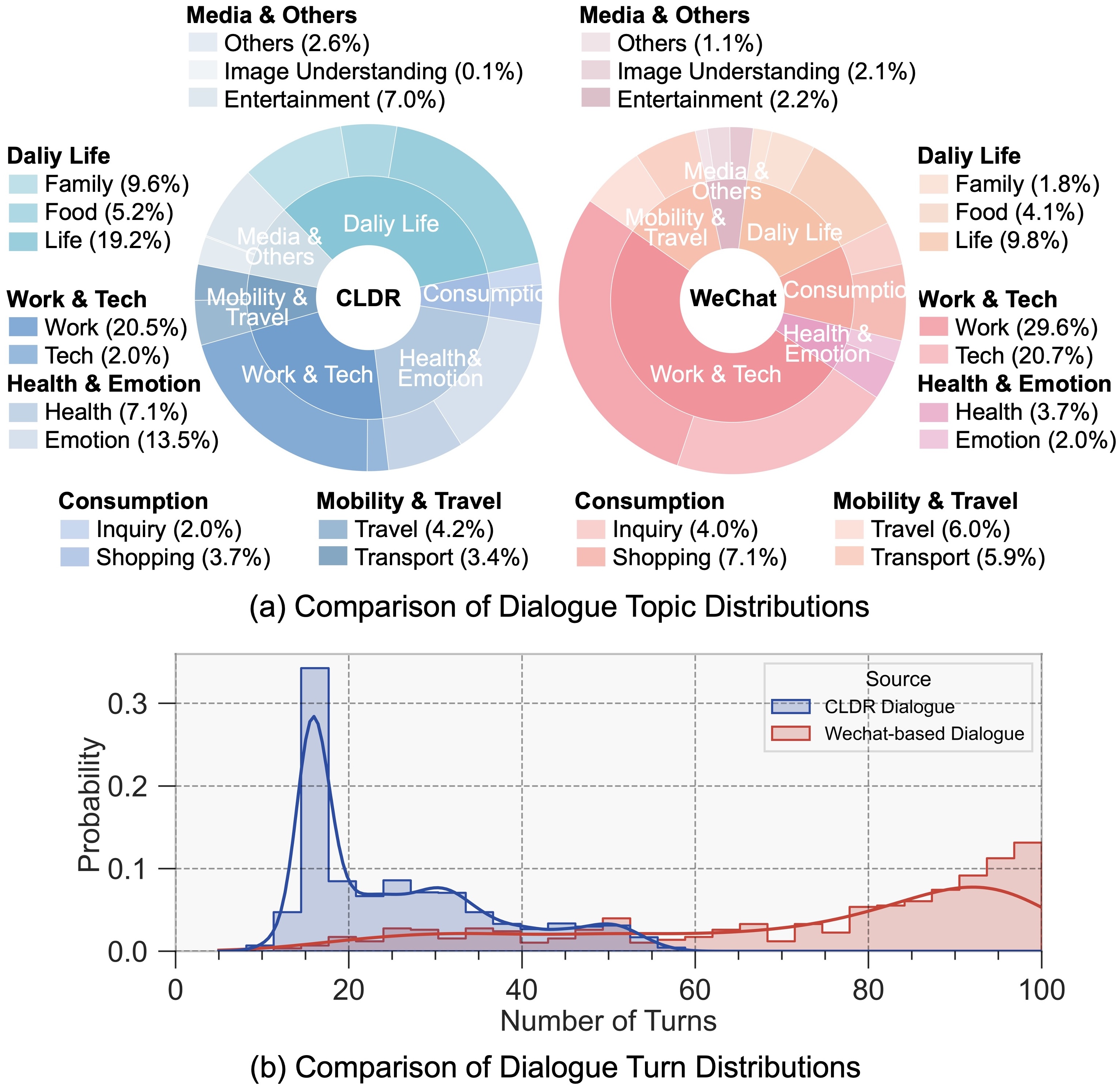}
\caption{Comparison between the MLDR and WeChat-based datasets in terms of dialogue topic and turn distributions.}
\label{fig:2}
\end{figure}

\subsection{Dataset Statistics and Analysis}
To better understand the characteristics of the MLDR dataset and its real-world generalization benchmark, we analyze topic coverage and turn-length distribution (more details are provided in Appendix.\ref{Dataset Statistics and Analysis}).

\noindent\textbf{Topic Diversity.}
Fig.\ref{fig:2}.a illustrates the dialogue topic distributions across six high-level domains and 14 fine-grained subtopics. Both MLDR and WeChat-based test sets demonstrate broad topical coverage, including daily life, work and technology, health and emotion, consumption, mobility and travel, and more. This diversity ensures that the dataset captures a wide range of user intents and real-world scenarios, providing a solid foundation for evaluating fragment-level retrieval in open-domain dialogues.

\noindent\textbf{Dialogue Turn Distribution.}
As illustrated in Fig.\ref{fig:2}.b, the MLDR and WeChat test sets exhibit distinctly different turn distributions. MLDR dialogues are tightly clustered around 15–25 turns, peaking near 20, which stems from its synthetic construction via controlled triplet concatenation. This design ensures topic continuity and manageable length, supporting stable training and evaluation. In contrast, WeChat dialogues display a broad, skewed distribution with many exceeding 60 turns and some approaching 100, reflecting the organic nature of real-world conversations—informal, multi-topic, and structurally irregular. This inherent complexity poses greater challenges for fragment retrieval, particularly in aligning sparse, semantically relevant content with user queries.

\section{Methodology}
This section introduces our proposed method for FFR from multi-modal long-form dialogues. We first introduce F$^2$RVLM, an MLLM-based retriever tailored for single-dialogue FFR. F$^2$RVLM is trained utilizing supervised learning and reinforcement learning to ensure that the retrieved fragments are both semantically precise and contextually coherent.
Next, we describe FFRS, a retrieval system designed for corpus-level FFR across large-scale dialogues. FFRS adopts an offline indexing and online dual-stage retrieval paradigm, combining embedding-based coarse-grained recall with fine-grained reasoning supported by F$^2$RVLM. This dual-stage framework enables accurate and efficient retrieval in real-world, corpus-level settings.

\begin{figure*}[t]
\setlength{\abovecaptionskip}{1pt}
\centering
\includegraphics[width=1.0\linewidth]{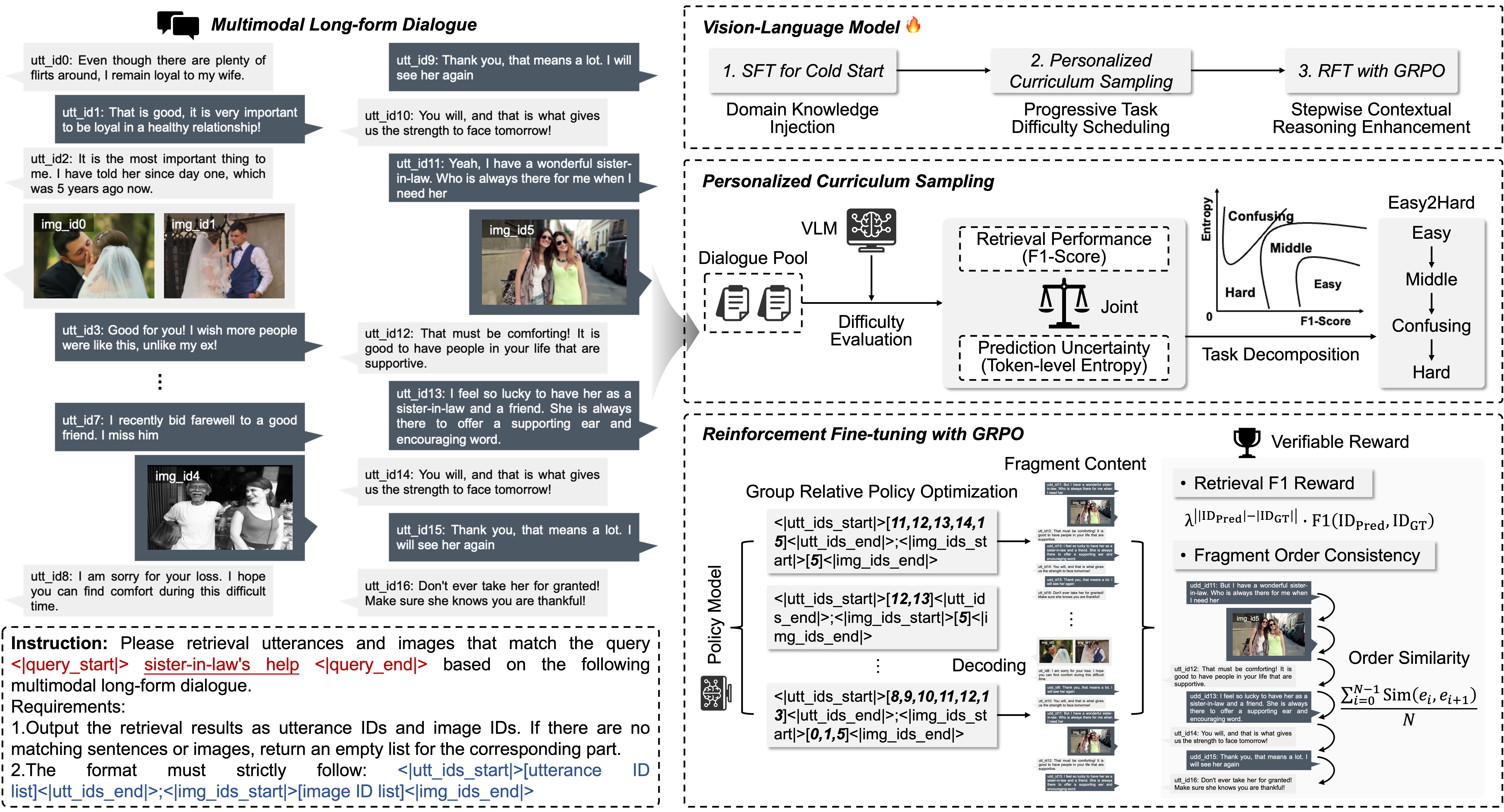}
\caption{\textbf{Overview of the F$^2$RVLM framework for Multi-modal Long-form Dialogue Fragment Retrieval.} It consists of supervised fine-tuning for fragment-level knowledge injection, personalized curriculum sampling based on retrieval difficulty, and GRPO-based reinforcement learning to jointly enhance retrieval accuracy and fragment semantic coherence.}
\label{fig:3}
  
\end{figure*}

\subsection{F$^2$RVLM: Fragment Retriever for Single-dialogue FFR}
\subsubsection{Framework Overview}
As illustrated in Fig.\ref{fig:3}, given a long-form dialogue $D$ and a query $q$, F$^2$RVLM generates structured outputs of relevant utterance and image IDs. To enable this behavior, the model is trained in two stages: first, supervised fine-tuning injects task-specific knowledge for fragment retrieval; then, GRPO-based reinforcement fine-tuning is conducted with a multi-objective reward that promotes semantic precision and contextual coherence. To stabilize training and enhance long-range reasoning, samples are organized into a difficulty-aware curriculum based on predicted F1 and uncertainty.

\subsubsection{Optimize Fragment Semantic Coherence by GRPO}
We adopt a rule-based variant of Group-Relative Policy Optimization (GRPO)~\citep{shao2024deepseekmath} to align model behavior with human-preferred retrieval patterns. To this end, we design three verifiable reward functions targeting format compliance, retrieval accuracy, and fragment order consistency, jointly guiding the model to generate structured, precise, and contextually coherent outputs.

\noindent \textbf{Preliminary of Group-Relative Policy Optimization}. Unlike prior methods~\citep{schulman2017proximal} that rely on an additional critic model to estimate absolute rewards, GRPO~\citep{shao2024deepseekmath} directly compares a group of candidate responses to estimate their relative quality, thereby significantly reducing computational costs. Given an input question $Q$, the model samples a group of $G$ candidate responses $o=\left \{ o_1, o_2,...,o_G \right \} $ from the policy model $\pi_{\theta_{old}}$. Each candidate is assessed by a rule-based and verifiable reward function $R(o_i)$, producing scores ${ r_1, r_2, \dots, r_G }$. GRPO then computes group-wise mean and standard deviation to derive the relative advantage of each response:
\begin{align} \label{equation:3}
A_i = \frac{r_i-\mathrm{mean}\left ( \left \{ r_1,r_2,...,r_G \right \}  \right )     }{\mathrm{std}\left ( \left \{ r_1,r_2,...,r_G \right \}  \right )  } 
\end{align}
where $A_i$ denotes the relative quality of the $i$-th response within the group. GRPO encourages the policy model to prefer responses with higher advantages. To prevent the optimized policy $\pi_{\theta}$ from deviating excessively from the original model $\pi_{ref}$, a KL-divergence term $\mathrm{D_{KL}}\left ( \cdot| \cdot  \right ) $ is incorporated as a regularization constraint in the training objective.

\noindent \textbf{Format Reward} $(R_{\text{Format}})$. 
$R_{\text{Format}}$ encourages the model to strictly adhere to the expected output format:
\begin{itemize}
    \item \texttt{<|utt\_ids\_start|>[...]<|utt\_ids\_end|>};
    \item \texttt{<|img\_ids\_start|>[...]<|img\_ids\_end|>}
\end{itemize}
This binary reward returns 1 only when the output fully matches the required format. A uniqueness constraint is also enforced: if any duplicate IDs are present, the reward is set to 0, even if the overall format appears correct.


\noindent \textbf{Retrieval F1 Reward} $(R_{\text{F1}})$. 
To encourage precise yet well-scoped retrieval, we design a reward function based on the F1 score with an exponential penalty on length deviation. It considers both utterance and image ID overlaps between predictions and ground truth, while explicitly penalizing over- and under-retrieval:
\begin{equation}
R_{\text{F1}} =  \sum_{m \in \{\text{utt}, \text{img}\}} \lambda_m \cdot \mathrm{F1}(I_m^{\text{pred}}, I_m^{\text{gt}}) \cdot \gamma^{|\,|I_m^{\text{pred}}| - |I_m^{\text{gt}}|\,|}
\end{equation}
where $I_{\text{*}}^{\text{pred}}$ and $I_{\text{*}}^{\text{gt}}$ denote the predicted and ground-truth ID sets, respectively. The weighting factors $\lambda$ balances modality importance. The penalty base $\gamma \in (0,1)$ modulates the severity of the length penalty, which increases exponentially with deviation in prediction length. 

\noindent \textbf{Fragment Order Consistency} $(R_{\text{Fragment}})$.
To encourage semantically coherent, contextually aligned fragment retrieval, we propose a reward based on cross-modal fragment order consistency. It evaluates whether the retrieved utterances and images maintain the natural progression of information, which is especially important in long conversations with intertwined modalities. 
Given the predicted sets of utterance and image IDs, we first locate the corresponding textual and visual elements in the original dialogue and encode them into a unified embedding space with CLIP. These embeddings are then interleaved and temporally ordered according to their original positions within the dialogue, forming a cross-modal sequence $\left \{ e_1,e_2,...,e_{N} \right \} $. The reward is defined as the average pairwise cosine similarity between adjacent embeddings in this sequence:
\begin{equation}
R_{\text{Fragment}} = \frac{1}{N-1} \sum_{i=1}^{N-1} \cos(\mathbf{e}_i, \mathbf{e}_{i+1})
\end{equation}
If the sequence length $N$ is less than 2, a fallback reward (i.e., 0.5) is returned.

\subsubsection{Difficulty-aware Curriculum Sampling}
Fragment structures in long-form dialogues vary significantly in the distribution of their internal elements: some fragments are tightly clustered within adjacent turns, while others span sparsely across distant positions. This structural variation substantially affects retrieval difficulty, clustered fragments are generally easier to localize, whereas dispersed ones pose greater challenges for alignment and consistency. To quantify this variability, we introduce a normalized span metric defined as: $\frac{\max(\text{turn\_id}) - \min(\text{turn\_id})}{L}$, where $L$ is the total number of turns in the dialogue. This metric captures how widely a relevant fragment is distributed relative to the overall dialogue length. Empirical statistics computed over our annotated benchmark reveal a wide range of dispersion: the normalized span values range from 0.5 to 13.43, with a mean of 1.08 and a standard deviation of 2.36. These results highlight the need for retrieval models to handle both tightly localized and sparsely distributed fragments under a unified framework.

To leverage this inherent difficulty hierarchy, we adopt a difficulty-aware curriculum sampling that dynamically quantifies instance difficulty based on predicted F1 scores and confidence. Training begins with high-confidence, high-F1 samples (i.e., easy cases) and gradually incorporates harder ones with lower scores. This progressive schedule enables the model to first master reasoning in dense contexts, then adapt to long-range, complex scenarios, mirroring the incremental nature of human learning.
Specifically, for each training sample $x_i$, we compute two indicators utilizing a cold-start model: (1) \textbf{Retrieval F1 Score} $f_i$: measures overlap between predicted and ground-truth utterance/image ID sets; and (2) \textbf{Prediction Entropy} $h_i$: quantifies uncertainty as the average entropy of predicted token distributions. Each instance is then assigned a difficulty level based on the following criteria:
\begin{equation}
d_i = 
\begin{cases}
\text{Easy}, & f_i \geq Q_{0.75}^{(f)} \ \text{and} \ h_i \leq Q_{0.25}^{(h)} \\
\text{Confusing}, & f_i \leq Q_{0.25}^{(f)} \ \text{and} \ h_i \geq Q_{0.75}^{(h)} \\
\text{Hard}, & f_i \leq Q_{0.25}^{(f)} \ \text{and} \ h_i \leq Q_{0.25}^{(h)} \\
\text{Medium}, & \text{otherwise}
\end{cases}
\end{equation}
Here, $Q_p^{(f)}$ and $Q_p^{(h)}$ denote the $p$-th percentiles of F1 and entropy distributions. “Easy” samples are confident and correct, “Confusing” are uncertain and incorrect, while “Hard” are incorrect yet overconfident. 

\begin{figure*}[t]
\setlength{\abovecaptionskip}{1pt}
\centering
\includegraphics[width=1.0\linewidth]{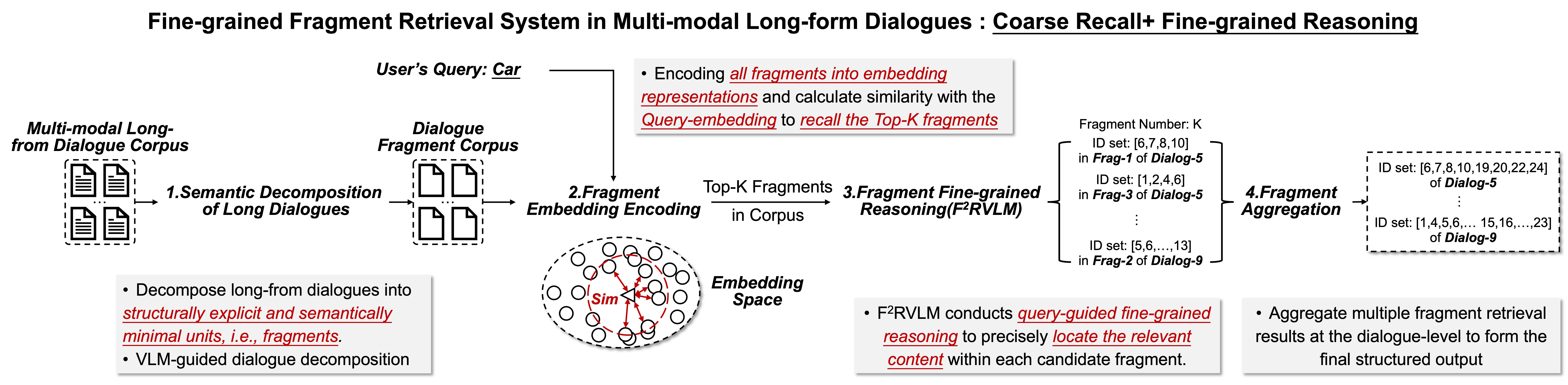}
\caption{\textbf{Overview of the proposed Fine-grained Fragment Retrieval System (FFRS) for Multi-modal Long-form Dialogues}, which consists of an offline indexing stage and an online retrieval stage.
\textbf{(1) Offline indexing.} 
First, long-form multi-modal dialogues are decomposed—under the guidance of a visual language model—into structurally explicit and semantically minimal fragments. 
Next, these fragments are encoded into dense vector representations and stored in a vector database to enable scalable, low-latency retrieval.
\textbf{(2) Online retrieval.} 
Given a user query, F$^2$RVLM performs query-guided fine-grained reasoning to precisely locate relevant content within each candidate fragment. 
Finally, retrieved fragments are aggregated at the dialogue level to produce a structured and contextually complete output.}
\label{fig:sys}
\end{figure*}

\subsection{FFRS: Fragment System for Corpus-level FFR}\label{FFRS}

\subsubsection{System Overview}
While F$^2$RVLM effectively addresses FFR within a single dialogue, real-world applications often demand retrieval across large-scale multi-dialogue corpora, where relevant fragments may be sparsely distributed over thousands of conversations. In such scenarios, relying solely on MLLM-based generation for fragment-level inference introduces significant scalability and latency challenges, the model must perform heavy contextual reasoning over each dialogue independently, leading to substantial computational overhead and inference latency that scales linearly with corpus size.

To overcome this bottleneck, we propose FFRS, a scalable two-stage retrieval system designed to balance semantic precision and retrieval efficiency for corpus-level FFR. As illustrated in Fig.\ref{fig:sys}, FFRS adopts an ``offline structured indexing and online dual-stage retrieval" paradigm.
\begin{enumerate}
    \item Offline phase. All multi-modal dialogues are first decomposed into fine-grained semantic units called fragments. Each fragment is encoded utilizing a dedicated Fragment Embedding Model (FEM) trained with a dual-level contrastive objective, and stored in a vector database. This structural indexing enables fast similarity search without repeatedly parsing full dialogues at runtime.
    \item Online phase. Upon receiving a user query, the system first conducts fast embedding-based retrieval to coarsely recall the Top-K candidate fragments utilizing FEM. These candidates are then further refined by F$^2$RVLM, which conducts more precise reasoning to locate sub-content that better matches the query semantics.
\end{enumerate}

This hybrid stage decouples heavy semantic modeling from the real-time path, allowing FFRS to scale to large corpora while retaining the nuanced understanding needed for fine-grained multi-modal retrieval.

\subsubsection{System Pipeline}
The overall system consists of five core steps, forming a closed-loop retrieval pipeline that spans from dialogue decomposition and fragment encoding to candidate retrieval and final result aggregation. We describe each step in detail below, with the pipeline illustrated in Algorithm \ref{FFRS pipeline}.

\begin{algorithm}[t]
\caption{FFRS: Fine-grained Fragment Retrieval System Pipeline} \label{FFRS pipeline}
\label{alg:ffrs}
\KwIn{User query $q$, Multi-modal dialogue corpus $\mathcal{D}$}
\KwOut{Retrieved fragment-level results $\mathcal{R}$}

\textbf{Offline Phase (Structured Indexing)}

\textbf{Step 1: Dialogue Decomposition}

\For{each dialogue $D \in \mathcal{D}$}{
Predict semantic fragment boundaries $\{f_1, f_2, \dots\}$ utilizing the Fragment Boundary Prediction Model;
Assign global fragment IDs and re-index local turn IDs;
}

\textbf{Step 2: Fragment Embedding \& Indexing}

\For{each fragment $f$}{
Encode $f$ utilizing the Fragment Embedding Model (FEM) with dual-level contrastive learning to obtain vector $\mathbf{e}_f$;

Store $(\mathbf{e}_f, \text{dialog-id}, \text{frag\_id})$ into vector database $\mathcal{I}$;
}

\textbf{Online Phase (Dual-stage Retrieval)}

\textbf{Step 3: Coarse Recall}

Encode query $q$ into embedding $\mathbf{e}_q$ via FEM;

Perform Top-$K$ vector similarity search over index $\mathcal{I}$ to obtain coarse candidates $\mathcal{C}$;

\textbf{Step 4: Fine-grained Reasoning}

\For{each candidate fragment $f_c \in \mathcal{C}$}{
Utilizing F$^{2}$RVLM to localize fine-grained turn indices within $f_c$:

$\hat{s}_{c} \leftarrow \text{F}^{2}\text{RVLM}(q, f_c)$;
}

\textbf{Step 5: Fragment Aggregation}

Group all $(f_c, \hat{s}_c)$ by their dialogue IDs;

Sort and merge overlapping spans to restore dialogue-level continuity;

Generate final structured retrieval output:
$\mathcal{R} \leftarrow \{ \text{dialog-id} : \text{merged spans} \}$;

\Return{$\mathcal{R}$}

\end{algorithm}

\noindent \textbf{(1) Dialogue Decomposition.}
Long-form dialogues, particularly those exceeding 100 turns and covering multiple distinct topics, pose significant challenges for direct semantic modeling. Without structural decomposition, feeding such raw dialogues into retrieval models typically leads to semantic dilution and a reduced ability to accurately match user queries with relevant responses. This issue is analogous to locating a specific passage in a book that lacks chapters or headings: in the absence of structural guidance, retrieval becomes inefficient and error-prone. To address this, FFRS employs a multi-modal semantic decomposition strategy that segments each dialogue into minimal, coherent semantic units, referred to as fragments. Each fragment encapsulates a distinct topic or intent (e.g., scheduling a meeting, describing a product, posing a question), enabling more precise content localization and mitigating interference from irrelevant dialogue turns.

This strategy is operationalized through a Fragment Boundary Prediction Model built on Qwen2.5-VL-3B~\citep{wang2024qwen2}, which is fine-tuned using instruction tuning on our annotated MLDR dataset. The model processes the entire dialogue and outputs a set of fragment intervals, e.g., ``[[1, 2, 3]", ``[5, 6, 7, 8]]", indicating that turns 1–3 form one fragment, and 5–8 form another. Each fragment is assigned a global dialogue-level ID (e.g., ``Frag-1 of Dialog-5") to track its position in the original conversation. Meanwhile, all utterances and images within the fragment are re-indexed with local turn-level IDs starting from 0. This structured representation supports fragment-level retrieval and enables subsequent re-aggregation at the dialogue level. By converting long, unstructured dialogues into structured sequences of fragments, this step lays a solid foundation for the embedding, indexing, and retrieval steps of the FFRS framework.

\noindent \textbf{(2) Fragment Embedding \& Offline Indexing.}
Once dialogues are decomposed into semantically independent fragments, each fragment is encoded into a fixed-dimensional dense vector to enable efficient indexing and retrieval. This step serves as a critical bridge between unstructured dialogue data and a structured, queryable memory space, supporting scalable vector-based search.

To capture fine-grained semantic information and ensure effective representation, we propose a dedicated Fragment Embedding Model (FEM) (see Sec.\ref{sec:FEM}), which is trained to encode localized intent and contextual clues at the fragment level. The resulting embeddings are stored in a vector database, along with metadata such as the corresponding dialogue ID and fragment ID. This storage format facilitates fast similarity computation using cosine distance and supports scalable retrieval in high-dimensional embedding space. The indexing pipeline is also designed to be incrementally updatable, allowing newly arriving dialogues to be decomposed, embedded, and inserted into the index without reprocessing the entire corpus. Overall, this embedding and indexing step lays the foundation for fast coarse retrieval and accurate downstream reasoning. By transforming decomposed multi-modal fragments into searchable vector representations while preserving semantic granularity, the FFRS framework ensures both retrieval scalability and semantic fidelity.

\noindent \textbf{(3) Coarse Recall via Similarity Search.}
In the online stage, FFRS begins with a fast coarse-grained recall step to narrow down the search space. Given a user query, often a single sentence or paragraph drawn from real-world task contexts, the system encodes it using the same FEM used in the offline stage, ensuring semantic space alignment. A Top-$K$ vector retrieval is then conducted using cosine similarity between the query embedding and all stored fragment vectors. This step rapidly selects the $K$ most semantically similar fragments from the corpus, substantially reducing the number of candidates passed to the next step. This coarse filtering brings two major advantages: (i) it transfers the bulk of retrieval computation to efficient vector operations, avoiding expensive inference over the entire corpus; and (ii) it ensures that system latency remains low even as the corpus scales to millions of fragments. The hyperparameter $K$ can be adjusted to balance recall and efficiency.

\noindent \textbf{(4) Fine-grained Reasoning with F$^2$RVLM.}
While coarse retrieval effectively narrows the corpus to a shortlist of semantically similar candidate fragments, these candidates may still contain false positives or lack fine-grained alignment with the user query. To ensure more precise content matching, we incorporate a fine-grained verification step utilizing the F$^2$RVLM model. Conditioned on the user query, F$^2$RVLM performs detailed reasoning over each candidate fragment to localize the specific subset of turns that directly address the query intent. As illustrated in Fig.\ref{fig:sys}, the model outputs a set of turn-level indices within each fragment, e.g., ``[6, 7, 8, 10] in Frag-1 of Dialog-5", indicating the most relevant dialogue spans. This enables the system to move beyond fragment-level retrieval toward sub-fragment localization, achieving more accurate and explainable results. By aligning the retrieved content at a finer granularity, this step significantly enhances the interpretability and reliability of the retrieval system, laying the foundation for structured aggregation and downstream multi-modal understanding.

\noindent \textbf{(5) Fragment Aggregation.}
The final stage of FFRS, Fragment Aggregation, is responsible for consolidating the reasoning outputs from F$^2$RVLM into structured, dialogue-level retrieval results. While fine-grained inference ensures precise localization within individual fragments, multiple candidate fragments may originate from the same long-form conversation. This stage reassembles semantically relevant content within each dialogue to restore contextual continuity and produce a unified retrieval output. 

Specifically, each verified fragment returned by F$^2$RVLM is accompanied by metadata, including the dialogue ID, fragment ID, and the localized set of turn indices (e.g., ``[6, 7, 8, 10]"). During aggregation, fragments are grouped according to their dialogue ID and sorted based on their original positions. Overlapping or adjacent spans are then merged to eliminate redundancy. This process results in a dialogue-level fragment set containing one or more semantically relevant, non-overlapping spans per dialogue, such as ``[1, 2, 3, 6, 7, 8, 10, 28, 29, 35]". The final output is thus structured at the granularity of entire dialogues: for each matched dialogue, FFRS returns a set of localized sub-fragments along with their associated text and visual content. This design not only enhances the interpretability and coherence of the retrieval results but also provides a standardized input format for downstream tasks, such as multi-modal dialogue understanding, summarization, or context-aware response generation.

\subsection{Fragment Embedding Model (FEM)}\label{sec:FEM}

To support scalable fragment-level retrieval in multi-modal long-form dialogues, we propose a tailored Fragment Embedding Model (FEM) that encodes each dialogue fragment into a fixed-dimensional dense representation. This model is central to the offline indexing process in FFRS, enabling coarse semantic filtering via vector similarity. 
\subsubsection{Limitations of Existing Embedding Models} 
Current multi-modal embedding models face two major limitations when applied to long-form, multi-turn dialogues: \textbf{(1) Coarse training supervision leading to insufficient granularity:} Most existing models are pre-trained on sentence-level or document-level data, lacking explicit fragment-level supervision. This results in a granularity mismatch between the query and candidate fragments, leading to unstable retrieval performance and poor semantic alignment in complex dialogue scenarios. \textbf{(2) Neglect of contextual and pragmatic dependencies:}  Existing models often treat dialogues or documents as concatenations of independent sentences. They lack mechanisms to model the natural dependencies that emerge across multiple turns in dialogue, such as question–answer pairs, elaboration chains, follow-up references, or image-text associations. As a result, the internal coherence of semantically unified fragments is frequently lost during embedding. This severely limits the model’s ability to represent a fragment as a single, cohesive semantic unit.

\subsubsection{The proposed FEM}
To address the limitations of existing multi-modal embedding models in long-form, multi-turn dialogue scenarios, we propose a Fragment Embedding Model (FEM) with a two-level contrastive learning strategy. The core objective is to ensure that: (1) semantically relevant query-fragment pairs are well-aligned at the global level, and (2) internal coherence and pragmatic structure within fragments are preserved via local structure modeling.

\noindent \textbf{Inter-Fragment Contrastive Loss.}
The global-level objective aligns semantically matched query-fragment pairs while pushing apart irrelevant ones. By optimizing the embedding space so that each query vector is close to its corresponding fragment and distant from others, the model learns to capture fine-grained semantic alignment. Given a batch of $N$ queries $\{q_i\}_{i=1}^N$ and their matched fragments $\{f_i\}_{i=1}^N$, the inter-fragment contrastive loss is defined as:
\begin{equation}
\mathcal{L}_{\text{inter}} = - \sum_{i=1}^{N} 
\log \frac{\exp(\text{sim}(q_i, f_i)/\tau)}
{\sum_{j=1}^{N} \exp(\text{sim}(q_i, f_j)/\tau)}
\end{equation}
where $\text{sim}(\cdot, \cdot)$ denotes cosine similarity, and $\tau$ is a temperature hyperparameter.

\noindent \textbf{Intra-Fragment Contrastive Loss.}
To more effectively capture the internal semantic structure and pragmatic dependencies within each fragment, we further decompose it into $K_i$ fine-grained question–answer (QA) pairs ${(q_i^k, a_i^k)}_{k=1}^{K_i}$. Each query $q_i^k$ reflects a localized communicative intent, such as subtopic initiation, clarification request, or visual reference, while the corresponding answer $a_i^k$ provides a contextually grounded elaboration or response. This decomposition allows the model to learn the latent discourse structure within a fragment, including QA dynamics, elaboration chains, and image–text grounding patterns. To preserve such intra-fragment coherence, we introduce a contrastive loss that explicitly aligns each query with its corresponding answer while pushing it away from other mismatched answers within the same fragment. The loss is formulated as:
\begin{equation}\label{eq:intra-frag}
\mathcal{L}_{\text{intra}} = - \sum_{i=1}^{N}\sum_{k=1}^{K_i}
\log \frac{\exp(\text{sim}(q_i^k, a_i^k)/\tau)}
{\sum_{m=1}^{K_i} \exp(\text{sim}(q_i^k, a_i^m)/\tau)}
\end{equation}
This objective encourages the model to internalize fragment-level cohesion by reinforcing fine-grained semantic bindings between interdependent units. As a result, the learned embeddings become more sensitive to intra-fragment structure, facilitating robust identification of self-contained, pragmatically coherent segments during downstream retrieval.

\noindent \textbf{Overall Loss.}
The final training objective combines the two losses with a weighting coefficient $\lambda$:
\begin{equation}
\mathcal{L}_{\text{FEM}} = \mathcal{L}_{\text{inter}} + \lambda \cdot \mathcal{L}_{\text{intra}}
\end{equation}
In our implementation, we set $\lambda=1.0$ and share the same backbone encoder across both contrastive objectives. This design allows FEM to effectively encode both inter-fragment semantic alignment and intra-fragment structural coherence, yielding expressive and retrieval-friendly embeddings for large-scale multi-modal dialogue systems.

\begin{table*}[t]
\setlength{\abovecaptionskip}{1pt}
\setlength{\belowcaptionskip}{1pt}
\caption{Comparison with popular VLMs on the MLDR validation set and WeChat test set. ``$\dagger$'' indicates zero-shot inference without MLDR fine-tuning. ``$\ast$'' indicates models limited by context length, evaluated via sliding-window inference (see Appendix.\ref{Sliding Window Inference for Long Contexts} for details).}
\centering
\renewcommand\arraystretch{1.25}
\resizebox{1.0\linewidth}{!}{
\begin{tabular}{r|cccc|cccc} 
\toprule
\multirow{2}{*}{Model}           & \multicolumn{4}{c}{In-domain Val Set (MLDR)}                      & \multicolumn{4}{c}{Real-domain Test Set (WeChat)}                  \\ 
\cline{2-9}
                                 & Precision(\%)  & Recall(\%)     & F1(\%)         & MCC(\%)        & Precision(\%)  & Recall(\%)     & F1(\%)         & MCC(\%)         \\ 
\hline
CLIP-Embedding$^\dagger$~\citep{radford2021learning}                    & 48.74          & 30.80          & 42.73          & 21.14          & 20.44          & 49.91          & 31.56          & 17.62           \\
BLIP2-Embedding$^\dagger$~\citep{li2023blip}                  & 31.88          & 2.96           & 23.47          & 0.00           & 15.22          & 52.74          & 25.05          & 4.94            \\
E5-V-Embedding$^\dagger$~\citep{jiang2024e5}                   & 53.85          & 48.83          & 51.82          & 28.92          & 30.33          & 47.13          & 36.99          & 24.40           \\
GME-Embedding$^\dagger$~\citep{zhang2024gme}                    & 62.27          & 23.75          & 35.52          & 24.17          & 29.63          & 53.11          & 38.22          & 26.68           \\
Qwen2.5-VL-7B$^\dagger$~\citep{bai2025qwen2}                    & 21.20          & 4.27           & 7.84           & 0.00           & 10.22          & 16.34          & 12.58          & 0.00            \\
MiMo-7B-RL$^\dagger$~\citep{coreteam2025mimovltechnicalreport}                       & 67.68          & 55.19          & 61.30          & 45.52          & 41.74          & 24.48          & 30.94          & 23.68           \\
Qwen2.5-VL-72B$^\dagger$~\citep{bai2025qwen2}                   & 61.11          & 67.14          & 64.09          & 44.61          & 32.55          & 36.95          & 36.60          & 25.26           \\
Doubao-Seed-1.6$^\dagger$~\citep{guo2025seed1}                  & 73.83          & 42.19          & 54.67          & 42.57          & 55.22          & 28.47          & 38.63          & 34.06           \\
Claude-Sonnet-4$^\dagger$~\citep{claude4.5}                  & 67.21          & 58.09          & 62.80          & 46.57          & 51.15          & 40.88          & 46.30          & 40.50           \\
GPT-4o$^\dagger$~\citep{jaech2024openai}                           & 70.43          & 52.49          & 60.32          & 45.91          & \underline{56.11}          & 41.82          & 48.89          & 42.85           \\
Gemini-2.5-Flash$^\dagger$~\citep{comanici2025gemini}                 & 70.18          & 69.30          & 69.87          & 54.66          & 51.89          & 49.80          & 53.21          & 48.22           \\ 
\hline

InternVL3-1B$^\ast$~\citep{chen2024internvl}           & 67.10          & 89.89          & 77.00          & 64.64          & 23.88          & 79.86          & 37.42          & 28.33           \\
InternVL3-2B$^\ast$~\citep{chen2024internvl}           & 75.73          & 91.86          & 83.07          & 74.11          & 31.59          & 73.57          & 44.32          & 33.34           \\
Ovis2-2B$^\ast$~\citep{lu2024ovis}               & 76.97          & 42.48          & 54.82          & 43.98          & 40.83          & 54.51          & 46.69          & 33.16           \\
DeepSeek-VL2-Tiny~\citep{wu2024deepseek}                & 62.97          & 84.25          & 72.23          & 56.59          & 29.83          & 21.12          & 24.73          & 16.79           \\
mPLUG-Owl3-2B~\citep{ye2024mplugowl3longimagesequenceunderstanding}                    & 74.86          & 83.18          & 78.88          & 67.59          & 18.35          & 45.99          & 26.37          & 11.61           \\
Qwen2-VL-2B~\citep{wang2024qwen2}                      & 74.97          & 92.99  & 83.07          & 74.18          & 23.20          & \underline{76.03}  & 36.65          & 26.55           \\
Qwen2.5-VL-3B~\citep{bai2025qwen2}                    & 80.35          & 91.41          & 85.57          & 77.98          & 33.75          & 75.82          & 47.60          & 39.56           \\
LLaVA-1.5-7B-hf$^\ast$~\citep{liu2023llava}        & 68.15          & 92.06          & 78.43          & 66.95          & 20.91          & \textbf{81.27} & 33.61          & 22.99           \\
MiMo-7B-SFT~\citep{coreteam2025mimovltechnicalreport}                      & 81.61          & \underline{93.25}  & 87.08          & 80.35          & 49.14          & 46.58          & 47.98          & 40.71           \\
MiMo-7B-RL~\citep{coreteam2025mimovltechnicalreport}                       & 80.90          & \textbf{93.46} & 86.71          & 79.78          & 49.46          & 48.22          & 49.05          & 41.84           \\
Qwen2-VL-7B~\citep{wang2024qwen2}                      & 83.15          & 90.72          & 86.81          & 79.92          & 42.80          & 69.60          & 53.54          & 45.75           \\
Qwen2.5-VL-7B~\citep{bai2025qwen2}
& 81.52          & 91.42          & 86.23          & 79.00          & 39.99          & 71.30          & 51.71          & 43.65           \\
InternVL3-8B$^\ast$~\citep{chen2024internvl}           & 80.67          & 92.98          & 86.43          & 79.35          & 34.08          & 75.52          & 47.07          & 36.65           \\
DeepSeek-VL2-Small-16B~\citep{wu2024deepseek}               & 77.52          & 91.13          & 83.82          & 75.23          & 47.43          & 19.48          & 27.91          & 24.51           \\
\rowcolor[rgb]{0.951,0.951,0.951}\textbf{F$^2$RVLM-Qwen2-VL-2B}   & 79.64          & 89.55          & 84.35          & 76.07          & 26.97          & 72.00          & 40.46          & 30.90           \\
\rowcolor[rgb]{0.951,0.951,0.951}\textbf{F$^2$RVLM-Qwen2.5-VL-3B} & 82.74          & 91.65          & 87.00          & 80.19          & 45.24          & 70.86          & 55.60          & 48.09           \\
\rowcolor[rgb]{0.951,0.951,0.951}\textbf{F$^2$RVLM-Qwen2-VL-7B}   & \textbf{84.02} & 90.67          & \textbf{87.25} & \textbf{80.60} & \textbf{57.21} & 67.46          & \textbf{62.07} & \textbf{55.46}  \\
\rowcolor[rgb]{0.951,0.951,0.951}\textbf{F$^2$RVLM-Qwen2.5-VL-7B} & \underline{83.24}  & 91.60          & \underline{87.24}  & \underline{80.55}  & 54.83  & 65.16          & \underline{59.60}  & \underline{52.39}   \\
\bottomrule
\end{tabular}
}
\label{tab:2}
\end{table*}

\begin{table*}[t]
\setlength{\abovecaptionskip}{1pt}
\setlength{\belowcaptionskip}{1pt}
\renewcommand\arraystretch{1.25}
\centering
\caption{Comparison with popular VLMs on the MLDR validation set. We report Precision, Recall, F1, and MCC for utterance, image, and joint predictions. ``$\dagger$'' indicates zero-shot inference without MLDR fine-tuning.}
\label{tab:A2}
\resizebox{1.0\linewidth}{!}{
\begin{tabular}{r|cccc|cccc|cccc} 
\toprule
\multirow{2}{*}{Model}                                          & \multicolumn{4}{c|}{Utterance Retrieval}                          & \multicolumn{4}{c|}{Image Retrieval}                              & \multicolumn{4}{c}{Joint Retrieval}                                \\ 
\cline{2-13}
& Precision      & Recall         & F1             & MCC            & Precision      & Recall         & F1             & MCC            & Precision      & Recall         & F1             & MCC             \\ 
\hline

CLIP-Embedding$^\dagger$                    & 45.21          & 20.42          & 28.13          & 12.18          & 52.86          & 62.63          & 57.33          & 30.11          & 48.74          & 30.80          & 42.73          & 21.14           \\
BLIP2-Embedding$^\dagger$                   & 30.54          & 78.97          & 44.05          & 0.00           & 33.33          & 1.51           & 2.89           & 0.00           & 31.88          & 2.96           & 23.47          & 0.00            \\
E5-V-Embedding$^\dagger$                    & 54.39          & 40.16          & 46.21          & 27.31          & 53.32          & 62.25          & 57.44          & 30.54          & 53.85          & 48.83          & 51.82          & 28.92           \\
GME-Embedding$^\dagger$                     & 56.96          & 33.09          & 41.86          & 26.09          & 68.66          & 18.52          & 29.18          & 22.25          & 62.27          & 23.75          & 35.52          & 24.17           \\
Qwen2.5-VL-7B$^\dagger$                     & 23.29          & 3.08           & 5.45           & 0.00           & 19.46          & 6.95           & 10.24          & 0.00           & 21.20          & 4.27           & 7.84           & 0.00            \\
MiMo-7B-RL$^\dagger$                        & 70.18          & 46.23          & 55.74          & 42.74          & 65.35          & 68.44          & 66.86          & 48.30          & 67.68          & 55.19          & 61.30          & 45.52           \\
Qwen2.5-VL-72B$^\dagger$                    & 62.42          & 60.67          & 61.54          & 44.27          & 59.85          & 75.15          & 66.63          & 44.95          & 61.11          & 67.14          & 64.09          & 44.61           \\
Doubao-Seed-1.6$^\dagger$                   & 79.69          & 33.69          & 47.36          & 40.79          & 68.78          & 56.42          & 61.99          & 44.36          & 73.83          & 42.19          & 54.67          & 42.57           \\
Claude-Sonnet-4$^\dagger$                   & 69.88          & 48.51          & 57.26          & 43.83          & 64.74          & 72.37          & 68.34          & 49.31          & 67.21          & 58.09          & 62.80          & 46.57           \\
GPT-4o$^\dagger$                            & 73.90          & 46.51          & 57.09          & 46.11          & 67.28          & 60.24          & 63.56          & 45.70          & 70.43          & 52.49          & 60.32          & 45.91           \\ 
Gemini-2.5-Flash$^\dagger$                  & 68.85          & 64.85          & 66.79          & 52.27          & 71.56          & 74.42          & 72.96          & 57.06          & 70.18          & 69.30          & 69.87          & 54.66           \\
\hline
InternVL3-1B                                                    & 62.49          & 89.16          & 73.48          & 60.64          & 72.45          & 90.63          & 80.53          & 68.65          & 67.10          & 89.89          & 77.00          & 64.64           \\
InternVL3-2B                                                    & 72.68          & 91.31          & 80.94          & 71.84          & 79.04          & 92.41          & 85.20          & 76.38          & 75.73          & 91.86          & 83.07          & 74.11           \\
Ovis2-2B                                                        & 73.85          & 46.27          & 56.90          & 45.33          & 80.37          & 39.26          & 52.75          & 42.63          & 76.97          & 42.48          & 54.82          & 43.98           \\
mPLUG-Owl3-2B                                                   & 71.70          & 81.80          & 76.42          & 64.93          & 78.32          & 84.60          & 81.34          & 70.25          & 74.86          & 83.18          & 78.88          & 67.59           \\
Qwen2-VL-2B                                                     & 71.88          & \underline{92.66}  & 80.96          & 71.97          & 78.35          & 93.32          & 85.18          & 76.38          & 74.97          & 92.99          & 83.07          & 74.18           \\
Qwen2.5-VL-3B                                                   & 77.21          & 91.22          & 83.64          & 75.85          & 83.75          & 91.60          & 87.50          & 80.12          & 80.35          & 91.41          & 85.57          & 77.98           \\
DeepSeek-VL2-Tiny-3B                                            & 58.54          & 83.69          & 68.89          & 52.87          & 68.12          & 84.81          & 75.56          & 60.30          & 62.97          & 84.25          & 72.23          & 56.59           \\
\rowcolor[rgb]{0.951,0.951,0.951} \textbf{F$^2$RVLM-Qwen2-VL-2B}   & 76.84          & 88.86          & 82.42          & 73.97          & 82.64          & 90.25          & 86.28          & 78.16          & 79.64          & 89.55          & 84.35          & 76.07           \\
\rowcolor[rgb]{0.951,0.951,0.951} \textbf{F$^2$RVLM-Qwen2.5-VL-3B} & 80.71          & 90.49          & 85.32          & 78.35          & 84.88          & 92.84          & 88.68          & 82.03          & 82.74          & 91.65          & 87.00          & 80.19           \\ 
\hline
LLaVa-1.5-7B-hf                                                 & 63.99          & 91.93          & 75.46          & 63.72          & 72.88          & 92.19          & 81.41          & 70.18          & 68.15          & 92.06          & 78.43          & 66.95           \\

MiMo-7B-SFT                                                     & 79.01          & 92.55          & 85.25          & 78.28          & 84.38          & \underline{93.97}  & \underline{88.92}  & \underline{82.41}  & 81.61          & \underline{93.25}  & 87.08          & 80.35           \\
MiMo-7B-RL                                                      & 78.17          & \textbf{92.85} & 84.88          & 77.75          & 83.63          & \textbf{94.08} & 88.55          & 81.81          & 80.90          & \textbf{93.46} & 86.71          & 79.78           \\
Qwen2-VL-7B                                                     & 80.60          & 89.60          & 84.86          & 77.66          & 85.86          & 91.87          & 88.76          & 82.18          & 83.15          & 90.72          & 86.81          & 79.92           \\
Qwen2.5-VL-7B                                                   & 78.74          & 90.88          & 84.37          & 76.93          & 84.51          & 91.98          & 88.09          & 81.07          & 81.52          & 91.42          & 86.23          & 79.00           \\
InternVL3-8B                                                    & 77.74          & 92.48          & 84.47          & 77.14          & 83.82          & 93.48          & 88.39          & 81.56          & 80.67          & 92.98          & 86.43          & 79.35           \\
DeepSeek-VL2-Small-16B                                          & 74.93          & 90.56          & 82.01          & 73.40          & 80.29          & 91.71          & 85.62          & 77.05          & 77.52          & 91.13          & 83.82          & 75.23           \\
\rowcolor[rgb]{0.951,0.951,0.951} \textbf{F$^2$RVLM-Qwen2-VL-7B}   & \textbf{82.19} & 88.99          & \underline{85.46}  & \underline{78.58}  & \textbf{85.93} & 92.41          & \textbf{89.05} & \textbf{82.63} & \textbf{84.02} & 90.67          & \textbf{87.25} & \textbf{80.60}  \\
\rowcolor[rgb]{0.951,0.951,0.951} \textbf{F$^2$RVLM-Qwen2.5-VL-7B} & \underline{81.64}  & 90.49          & \textbf{85.84} & \textbf{79.12} & \underline{84.91}  & 92.73          & 88.65          & 81.97          & \underline{83.24}  & 91.60          & \underline{87.24}  & \underline{80.55}   \\
\bottomrule
\end{tabular}
}
\end{table*}

\begin{table*}[t]
\setlength{\abovecaptionskip}{1pt}
\setlength{\belowcaptionskip}{1pt}
\renewcommand\arraystretch{1.25}
\centering
\caption{Comparison with popular VLMs on the WeChat test set. We report Precision, Recall, F1, and MCC for utterance, image, and joint predictions. ``$\dagger$'' indicates zero-shot inference without MLDR fine-tuning. ``$\ast$'' indicates models limited by context length, evaluated via sliding-window inference.}
\label{tab:A3}
\resizebox{1.0\linewidth}{!}{
\begin{tabular}{r|cccc|cccc|cccc} 
\toprule
\multirow{2}{*}{Model}                                         & \multicolumn{4}{c|}{Utterance Retrieval}                                         & \multicolumn{4}{c|}{Image Retrieval}                                        & \multicolumn{4}{c}{Joint Retrieval}                                          \\ 
\cline{2-13}
& Precision      & Recall         & F1             & MCC            & Precision      & Recall         & F1             & MCC            & Precision      & Recall         & F1             & MCC             \\ 
\hline
CLIP-Embedding$^\dagger$                    & 16.44          & 35.99          & 22.57          & 9.15           & 27.01          & 81.37          & 40.55          & 26.08          & 20.44          & 49.91          & 31.56          & 17.62           \\
BLIP2-Embedding$^\dagger$                   & 11.00          & 71.41          & 19.07          & 0.00           & 24.68          & 41.81          & 31.04          & 11.97          & 15.22          & 52.74          & 25.05          & 4.94            \\
E5-V-Embedding$^\dagger$                    & 34.73          & 35.82          & 35.27          & 26.80          & 26.92          & 68.89          & 38.72          & 22.00          & 30.33          & 47.13          & 36.99          & 24.40           \\
GME-Embedding$^\dagger$                     & 32.54          & 39.12          & 35.52          & 26.51          & 27.19          & 82.67          & 40.93          & 26.86          & 29.63          & 53.11          & 38.22          & 26.68           \\
Qwen2.5-VL-7B$^\dagger$                                     & 13.95          & 11.38          & 12.53          & 2.77           & 8.07           & 28.98          & 12.62          & 0.00           & 10.22          & 16.34          & 12.58          & 0.00            \\
MiMo-7B-RL$^\dagger$                                     & 41.63          & 22.70          & 29.38          & 24.49           & 41.85           & 26.55          & 32.49          & 22.88           & 41.74          & 24.48          & 30.94          & 23.68            \\
Qwen2.5-VL-72B$^\dagger$                                    & 28.25          & 27.90          & 28.07          & 18.88          & 38.41          & 54.70          & 45.13          & 31.63          & 32.55          & 36.95          & 36.60          & 25.26           \\ 
Doubao-Seed-1.6$^\dagger$                                    & 61.83          & 21.81          & 32.24          & 32.49          & 49.89          & 41.01          & 45.02          & 35.62          & 55.22          & 28.47          & 38.63          & 34.06           \\
Claude-Sonnet-4$^\dagger$                                    & 60.38          & 29.89          & 39.98          & 37.69          & 44.36          & 64.68          & 52.63          & 43.31          & 51.15          & 40.88          & 46.30          & 40.50           \\
GPT-4o$^\dagger$                                             & 60.43          & 32.19          & 42.01          & 39.18          & 52.37          & 59.65          & 55.77          & 46.52          & \underline{56.11}          & 41.82          & 48.89          & 42.85           \\
Gemini-2.5-Flash$^\dagger$                                   & 48.65          & 36.86          & 41.94          & 35.92          & 55.60          & 76.76          & \underline{64.49}          & \textbf{60.52}          & 51.89          & 49.80          & 53.21          & 48.22           \\

\hline

InternVL3-1B$^\ast$                                                  & 19.97          & \underline{86.85} & 32.48          & 15.27          & 29.69          & \underline{73.92} & 42.37          & 41.39          & 23.88          & \underline{79.86} & 37.42          & 28.33           \\
InternVL3-2B$^\ast$                                                  & 28.34          & 81.31          & 42.02          & 22.41          & 35.70          & 67.18          & 46.62          & 44.27  & 31.59          & 73.57          & 44.32          & 33.34           \\
Ovis2-2B$^\ast$                                                      & 39.72  & 56.87          & 46.77  & 23.54          & 42.00          & 52.34          & 46.60          & 42.79          & 40.83          & 54.51          & 46.69          & 33.16           \\
mPLUG-Owl3-2B                                                        & 20.65          & 45.01          & 28.31          & 16.78          & 16.51          & 47.01          & 24.44          & 6.44           & 18.35          & 45.99          & 26.37          & 11.61           \\
Qwen2-VL-2B                                                   & 18.38          & 86.49  & 30.32          & 23.64          & 31.46          & 67.83          & 42.98          & 29.45          & 23.20          & 76.03  & 36.65          & 26.55           \\
Qwen2.5-VL-3B                                                 & 27.77          & 79.03          & 41.11          & 35.70  & 43.02          & 72.86  & 54.09  & 43.41          & 33.75          & 75.82          & 47.60  & 39.56   \\
DeepSeek-VL2-Tiny-3B                                    & 21.95          & 27.96          & 24.59          & 13.69          & 46.52          & 16.97          & 24.87          & 19.89          & 29.83          & 21.12          & 24.73          & 16.79           \\
\rowcolor[rgb]{0.951,0.951,0.951} \textbf{F$^2$RVLM-Qwen2-VL-2B}   & 20.99          & 82.21          & 33.45          & 27.11          & 37.72          & 64.04          & 47.48          & 34.70          & 26.97          & 72.00          & 40.46          & 30.90           \\
\rowcolor[rgb]{0.951,0.951,0.951} \textbf{F$^2$RVLM-Qwen2.5-VL-3B} & 39.54          & 71.98          & 51.04 & 45.31 & 52.87  & 69.78          & 60.16 & 50.88 & 45.24  & 70.86          & 55.60 & 48.09  \\ 
\hline
LLaVA-1.5-7B-hf$^\ast$                                               & 18.16          & \textbf{89.80} & 30.21          & 9.78           & 24.66          & \textbf{74.21} & 37.02          & 36.20          & 20.91          & \textbf{81.27} & 33.61          & 22.99           \\
MiMo-7B-SFT                                                 & 40.06          & 52.03          & 45.27          & 37.60          & \underline{63.55}  & 42.16          & 50.69          & 43.82          & 49.14          & 46.58          & 47.98          & 40.71           \\
MiMo-7B-RL                                                  & 40.05          & 53.53          & 45.82          & 38.23          & \textbf{64.66} & 43.87          & 52.28          & 45.46          & 49.46          & 48.22          & 49.05          & 41.84           \\
Qwen2-VL-7B                                                   & 36.28          & 71.65          & 48.17          & 42.20          & 52.17          & 67.65          & 58.91          & 49.30          & 42.80          & 69.60          & 53.54          & 45.75           \\
Qwen2.5-VL-7B                                                 & 34.74          & 71.70          & 46.80          & 40.74          & 47.11          & 70.90  & 56.61          & 46.57          & 39.99          & 71.30          & 51.71          & 43.65           \\
InternVL3-8B$^\ast$                                                  & 31.05          & 80.78  & 44.86          & 25.98          & 37.76          & 70.90  & 49.28          & 47.33          & 34.08          & 75.52  & 47.07          & 36.65           \\
DeepSeek-VL2-Small-16B                                  & 40.13 & 24.92          & 30.75          & 24.93          & 57.98 & 16.00          & 25.07          & 24.09          & 47.43 & 19.48          & 27.91          & 24.51           \\
\rowcolor[rgb]{0.951,0.951,0.951} \textbf{F$^2$RVLM-Qwen2-VL-7B}   & \underline{52.36}  & 67.51          & \textbf{58.98} & \textbf{53.50} & 63.05          & 67.42          & \textbf{65.16} & \textbf{57.41} & \textbf{57.21} & 67.46          & \textbf{62.07} & \textbf{55.46}  \\
\rowcolor[rgb]{0.951,0.951,0.951} \textbf{F$^2$RVLM-Qwen2.5-VL-7B} & \textbf{52.44} & 64.51          & \underline{57.85}  & \underline{52.16}  & 57.46          & 65.82          & 61.36  & 52.63  & 54.83  & 65.16          & \underline{59.60}  & \underline{52.39}   \\
\bottomrule
\end{tabular}
}
\end{table*}

\section{Experiments}
This section systematically evaluates FFR under both single-dialogue and corpus-level settings.

\subsection{F$^2$RVLM for Single-Dialogue FFR}
\subsubsection{Experimental Settings}

\noindent \textbf{Models \& Details.} 
We implement F$^2$RVLM based on the ms-swift~\citep{zhao2024swiftascalablelightweightinfrastructure} framework, using Qwen-VL-series as the backbone with 2B, 3B, and 7B parameters. Fine-tuning is conducted on the MLDR dataset, with 1,000 samples reserved for validation and the remaining for training. Among the training data, 25\% is used for SFT cold start, while the remaining is reserved for RFT. We adopt LoRA~\citep{hu2022lora} for parameter-efficient adaptation during both the SFT and RFT stages. During GRPO fine-tuning in RFT, we sample $G = 8$ candidate responses for each instance. The reward weights are set to $\lambda_{\text{utt}} = \lambda_{\text{img}} = 0.5$, and the exponential length penalty base is $\gamma = 0.95$. Curriculum sampling begins with 10\% of the easy and medium-level instances and gradually incorporates confusing and hard instances as training progresses. To enhance training stability and efficiency, we adopt the dynamic sampling strategy from DAPO~\citep{yu2025dapo}. Gradient checkpointing and FlashAttention are utilized to enhance training efficiency. All models are trained for 1 epoch on 8$\times$A100 GPUs.

For evaluation, we compare F$^2$RVLM against a comprehensive set of MLLMs, including proprietary models such as GPT-4o~\citep{jaech2024openai}, Gemini-2.5~\citep{comanici2025gemini}, Doubao-Seed-1.6~\citep{guo2025seed1}, and Claude-Sonnet-4~\citep{claude4.5}; open-source models including LLaVa~\citep{liu2023llava}, Qwen-VL-series~\citep{wang2024qwen2}, DeepSeek-VL2-series~\citep{wu2024deepseek}, InternVL3-series~\citep{chen2024internvl}, Ovis2~\citep{lu2024ovis}, Owl3~\citep{ye2024mplugowl3longimagesequenceunderstanding}, and Mimo-VL~\citep{coreteam2025mimovltechnicalreport}; as well as embedding models such as CLIP~\citep{radford2021learning}, BLIP-2~\citep{li2023blip}, E5-V~\citep{jiang2024e5}, and GME~\citep{zhang2024gme}. Open-source models are fine-tuned on MLDR using SFT, while proprietary and embedding models are evaluated in inference-only mode.

\noindent \textbf{Metrics.}
We evaluate fragment-level retrieval performance utilizing four metrics: Precision, Recall, F1 Score, and Matthews Correlation Coefficient (MCC), computed separately for utterance IDs and image IDs. To obtain unified scores, we average the F1 and MCC values across both modalities and calculate the harmonic mean of Precision and Recall to reflect joint retrieval performance.

\subsubsection{Comparison with Popular VLMs}

\noindent \textbf{Overall Retrieval Performance.}
Table~\ref{tab:2} summarizes fragment-level retrieval results on the in-domain MLDR validation set and the real-domain WeChat test set. Key observations include: 
(i) F$^2$RVLM achieves SOTA performance in both domains. The 7B variant tops MLDR with 87.25\% F1, and leads on WeChat with 62.07\% F1. Its 3B version also outperforms larger competitors such as MiMo-7B-RL, and GPT-4o.
(ii) MLDR Fine-tuning significantly boosts cross-domain retrieval.
Qwen2.5-VL-7B improves from 12.58\% (zero-shot) to 51.71\% F1 on WeChat after MLDR tuning, demonstrating MLDR's value as a supervised resource for fragment retrieval. Moreover, F$^2$RVLM offers less performance degradation from MLDR to WeChat compared to other models, indicating stronger generalization in real-world long-form dialogues.
(iii) F$^2$RVLM consistently outperforms baselines with comparable or larger parameter scales, validating the effectiveness of our GRPO-based training with curriculum sampling and multi-objective reward optimization.
(iv) Qualitative analysis provided in Appendix.\ref{Experimental Results in Single-Dialogue FFR} further verifies the robustness of F$^2$RVLM. Case studies show that the model accurately identifies semantically relevant fragments, even when key evidence is distributed across multiple turns, highlighting its strong alignment with real user intent.

\noindent \textbf{Detailed Retrieval Performance.} Tables~\ref{tab:A2} and \ref{tab:A3} further report the evaluation metrics assessing the sentence and image elements contained within each predicted fragment. Precision, recall, F1, and MCC are computed separately for utterance IDs and image IDs, and their averages form the unified fragment-level scores, as shown in Table~\ref{tab:2}. This design enables a clearer assessment of how well a model captures the internal composition of a fragment rather than merely identifying its boundary. Key observations include: 
\begin{enumerate}
    \item Utterance correctness remains the most challenging component.
Many baselines retrieve fragments that contain correct images but inaccurate or redundant sentences, leading to a persistent gap between utterance-level and image-level F1. F$^2$RVLM substantially reduces this gap, achieving 85.32\% F1 on MLDR and 58.98\% on WeChat. This demonstrates that fragment-level rewards and curriculum sampling effectively strengthen fine-grained textual grounding.
\item Image selection benefits from fragment-aware grounding rather than simple visual–text matching.
Although several VLMs reach high image recall, they often include visually similar yet irrelevant frames. F$^2$RVLM maintains a more balanced precision–recall profile (e.g., 88.68\% on MLDR; 65.16\% on WeChat), indicating improved selectivity in identifying images that genuinely belong to the target fragment.
\end{enumerate}

Overall, these results show that F$^2$RVLM not only identifies the correct fragment boundaries but also provides more faithful intra-fragment composition, aligning both textual and visual elements more accurately with the ground truth.

\noindent \textbf{Discussion about Recall Metric.} While models like LLaVA achieve higher recall, they often sacrifice precision by retrieving many irrelevant fragments. In contrast, our model incorporates a penalty term in the $R_{\text{F1}}$ to suppress over-retrieval, encouraging the selection of fewer but more semantically consistent fragments, better aligned with human preferences and fragment-level retrieval objectives.

\begin{figure}[t]
\setlength{\abovecaptionskip}{1pt}
\centering
\includegraphics[width=1.0\linewidth]{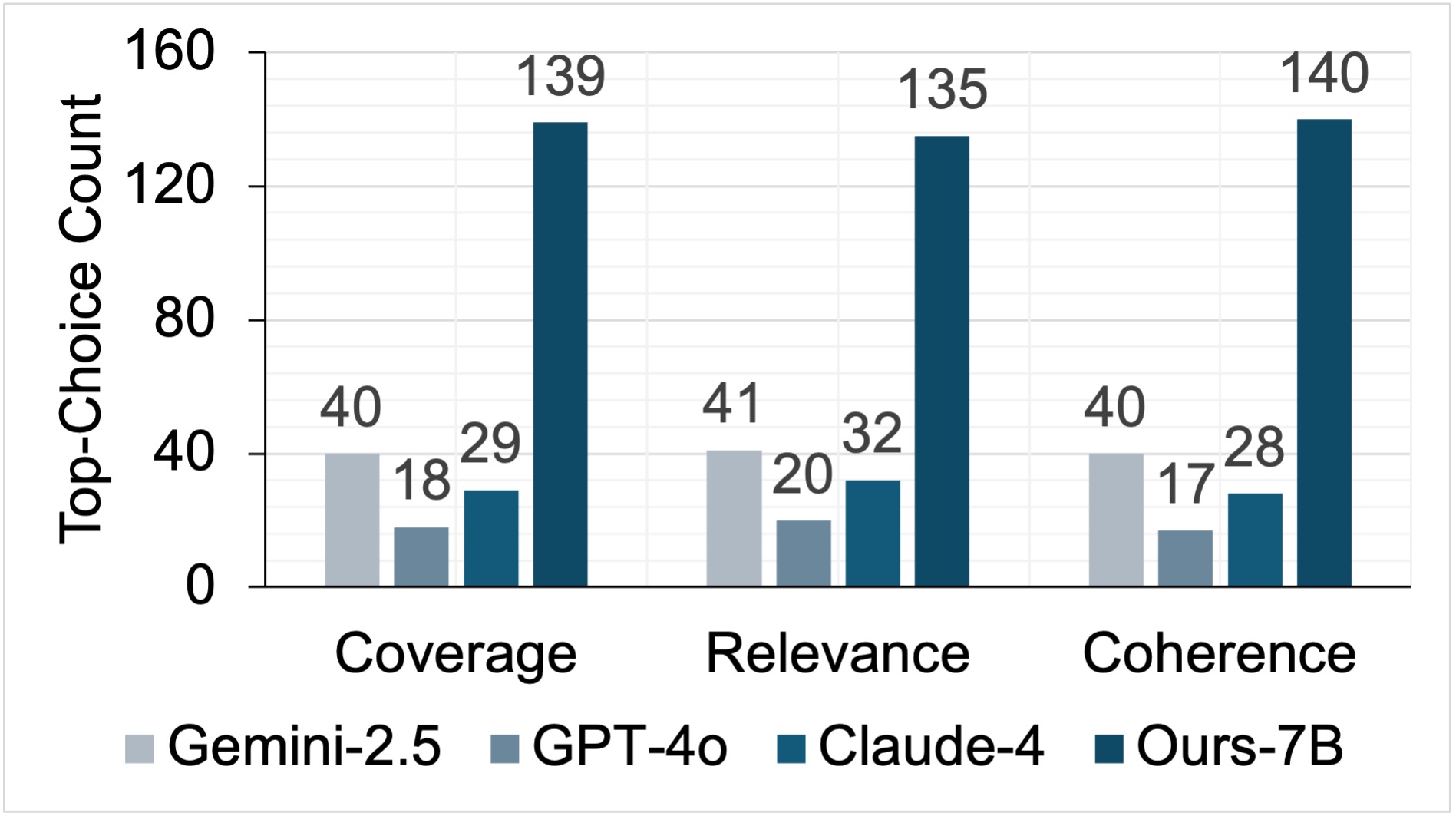}
\caption{\textbf{Human subjective evaluation results on the WeChat test set.} We randomly sample 200 dialogues and ask expert annotators to select the top-performing model across three criteria: Coverage, Relevance, and Coherence.}
\label{fig:5}
\end{figure}

\begin{table}[t]
\setlength{\abovecaptionskip}{1pt}
\setlength{\belowcaptionskip}{1pt}
\centering
\caption{Comparison with popular VLMs on the real-domain WeChat test set, evaluated by Fragment Order Consistency and Query-Fragment Similarity. ``$\dagger$'' indicates zero-shot inference without MLDR fine-tuning.} \label{tab:3}
\renewcommand\arraystretch{1.25}
\resizebox{1.0\linewidth}{!}{
\begin{tabular}{rccc} 
\toprule
Model                                                & \#Params                                                                 & Fragment Consis. & Query Sim.      \\ 
\hline
Doubao-Seed-1.6$^\dagger$                                      & \multirow{4}{*}{\begin{tabular}[c]{@{}c@{}}Closed\\-Source\end{tabular}} & 27.89            & 31.06           \\
Gemini-2.5-Flash$^\dagger$                                     &                                                                          & 37.67            & 41.44           \\
Claude-Sonnet-4$^\dagger$                                      &                                                                          & 42.29            & 46.64           \\
GPT-4o$^\dagger$                                               &                                                                          & 43.17            & 48.21           \\
Qwen2.5-VL$^\dagger$                                           & 72B                                                                      & 45.98            & 48.49           \\ 
\hline
DeepSeek-VL2-Small                                   & 16B                                                                      & 12.52            & 13.67           \\
MiMo-VL-RL                                           & 7B                                                                       & 34.69            & 36.69           \\
Qwen2-VL                                             & 7B                                                                       & \underline{55.28}    & \underline{59.49}   \\
Qwen2.5-VL                                           & 7B                                                                       & 54.81            & 58.39           \\
\rowcolor[rgb]{0.951,0.951,0.951} \textbf{F$^2$RVLM} & 3B                                                                       & \textbf{60.53}   & \textbf{61.18}  \\
\bottomrule
\end{tabular}}
\end{table}

\begin{figure}[t]
\centering
\includegraphics[width=1.0\linewidth]{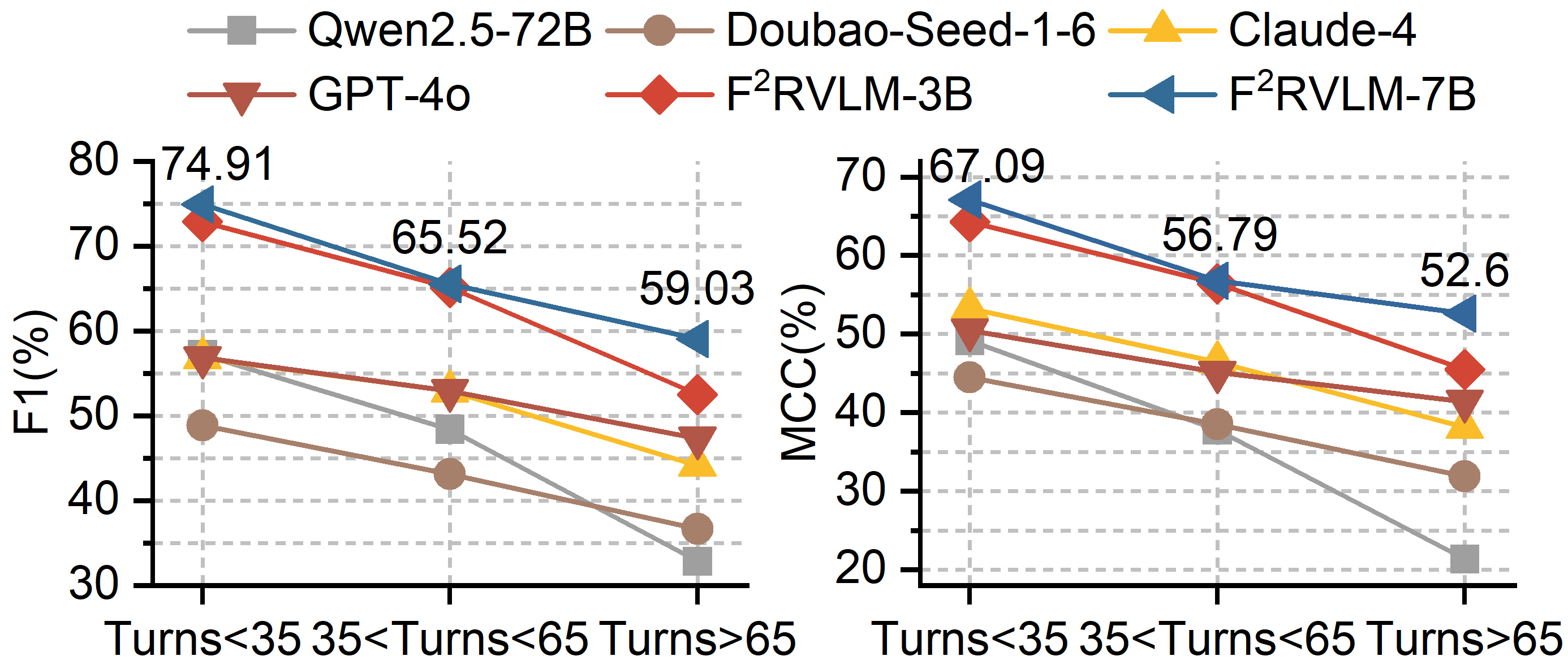}
\caption{\textbf{Performance comparison across dialogue turn groups on the real-domain WeChat test set.} We report F1 and MCC across three dialogue turn ranges: short ($<$35 turns), medium (35–65 turns), and long ($>$65 turns).}
\label{fig:4}
\end{figure}

\begin{table}[t]
\setlength{\abovecaptionskip}{1pt}
\setlength{\belowcaptionskip}{1pt}
\centering
\caption{Ablation study on reward components and curriculum sampling on MLDR validation and WeChat test sets.} \label{tab:5}
\renewcommand\arraystretch{1.25}
\resizebox{1.0\linewidth}{!}{
\begin{tabular}{ccc|cccc} 
\toprule
\multirow{2}{*}{$R_\text{F1}$}         & \multirow{2}{*}{$R_\text{Fragment}$}   & \multirow{2}{*}{Curriculum}            & \multicolumn{2}{c}{In-domain} & \multicolumn{2}{c}{Real-domain}  \\
&                                        &                                        & F1(\%)         & MCC(\%)             & F1(\%)         & MCC(\%)                 \\ 
\hline
\textcolor[rgb]{0.702,0.702,0.702}{\ding{55}} & \textcolor[rgb]{0.702,0.702,0.702}{\ding{55}} & \ding{51}                                     & 86.02          & 78.69               & 53.39          & 46.07                   \\
\ding{51}                                     & \textcolor[rgb]{0.702,0.702,0.702}{\ding{55}} & \ding{51}                                     & \underline{86.39}  & \underline{79.24}       & \underline{54.68}  & \underline{47.41}           \\
\ding{51}                                     & \ding{51}                                     & \textcolor[rgb]{0.702,0.702,0.702}{\ding{55}} & 84.69          & 77.53               & 50.65          & 43.71                   \\
\ding{51}                                     & \ding{51}                                     & \ding{51}                                     & \textbf{87.00} & \textbf{80.19}      & \textbf{55.60} & \textbf{48.09}          \\
\bottomrule
\end{tabular}}
\end{table}

\begin{table}[t]
\setlength{\abovecaptionskip}{1pt}
\setlength{\belowcaptionskip}{1pt}
\centering
\caption{Ablation study of different SFT and GRPO data ratios in F$^2$RVLM on MLDR validation and WeChat test sets.} \label{tab:5_1}
\renewcommand\arraystretch{1.25}
\resizebox{1.0\linewidth}{!}{
\begin{tabular}{c|cc|cc} 
\toprule
\multirow{2}{*}{Training Strategy} & \multicolumn{2}{c|}{In-domain} & \multicolumn{2}{c}{Real-domain}  \\ 
\cline{2-5}
& F1(\%)    & MCC(\%)                              & F1(\%)    & MCC(\%)                               \\ 
\hline
All SFT                            & 85.57 & 77.98                            & 47.60 & 39.56                             \\
All GRPO(0.1\% SFT)                & 82.14      & 73.29                                 & 45.47      & 37.81                                   \\
\hline
5\%SFT+95\%GRPO                    & 84.58      & 78.65                                & 53.64      & 42.83                                   \\
10\%SFT+90\%GRPO                   & 86.44       & 79.26                                 & 54.51      &  47.18                                 \\
25\%SFT+75\%GRPO                   & \underline{87.00} & \underline{80.19}                            & \textbf{55.60} & \textbf{48.09}                             \\
50\%SFT+50\%GRPO                   & \textbf{87.37}      & \textbf{81.05}                                  & \underline{54.32}      & \underline{47.33}                                  \\
\bottomrule
\end{tabular}
}
\end{table}

\noindent \textbf{Fragment-level Consistency and Alignment.}
To assess VLMs’ ability to capture dialogue structure and semantic alignment, we introduce two metrics on the WeChat test set: Fragment Order Consistency (average cosine similarity between adjacent retrieved elements) and Query-Fragment Similarity (average similarity between the query and each retrieved element), as reported in Table~\ref{tab:3}. F$^2$RVLM achieves the highest scores on both, outperforming larger open- and closed-source models. This demonstrates its superior ability to retrieve coherent, semantically aligned fragments, enabled by the order-consistency reward in GRPO training.

\noindent \textbf{Discussion about Dialogue Turns.}
Fig.\ref{fig:4} reports model performance on the WeChat test set across short ($<$35 turns), medium (35–65 turns), and long ($>$65 turns) dialogues. F$^2$RVLM-7B (Qwen2) consistently achieves the highest F1 and MCC scores as dialogue turn increases, with minimal performance degradation compared to other models, which demonstrates its robustness in long-context reasoning.

\noindent \textbf{Subjective Results.}
We conduct a human evaluation on 200 dialogues from the WeChat test set, comparing our 7B model against Gemini-2.5, GPT-4o, and Claude-4 across three criteria: coverage, relevance, and fragment coherence (Details are provided in Appendix.\ref{Experimental Results in Single-Dialogue FFR}). As depicted in Fig.\ref{fig:5}, our model is consistently preferred by annotators, receiving the highest top-choice counts in all aspects, demonstrating superior semantic completeness, alignment, and fluency in real-world dialogue retrieval.

\subsubsection{Ablation Results}

\noindent \textbf{Effectiveness of Reward Functions and Curriculum Learning.}  
This section presents ablation studies to evaluate the effectiveness of the proposed reward functions and personalized curriculum learning, focusing on joint F1 and MCC metrics (Table~\ref{tab:5}).  
(i) Replacing standard accuracy with $R_{\text{F1}}$ significantly improves performance, while incorporating fragment order consistency $R_{\text{Fragment}}$ further enhances cross-domain F1 and MCC, highlighting the role of intra-fragment semantic consistency.  
(ii) Integrating difficulty-aware curriculum sampling into GRPO yields consistent gains: in-domain F1 improves from 84.69\% to 87.00\%, and real-domain F1 from 50.65\% to 55.60\%, validating the effectiveness of progressive learning from easier to harder samples.  

\noindent \textbf{Training Ratio Between SFT and GRPO-based RFT.}  
Table~\ref{tab:5_1} explores the impact of different proportions of SFT and GRPO-based RFT on model performance.  
We observe the following:
(i) Using GRPO alone underperforms compared to using only SFT, highlighting the indispensable role of supervised learning for cold-start in providing stable and accurate training signals;  
(ii) As the proportion of GRPO increases, overall performance gradually surpasses that of pure SFT, demonstrating the effectiveness of RFT for dialogue fragment retrieval;  
(iii) A 25\% SFT and 75\% GRPO ratio achieves the best results on the real-domain test set, while increasing SFT to 50\% slightly improves in-domain validation performance but compromises generalization to real-world scenarios.

\subsection{FFRS for Corpus-level FFR}
This section first evaluate the performance of key components proposed in each step, including the fragment boundary prediction model and the fragment embedding model, followed by the overall system performance.

\subsubsection{Evaluation of Fragment Boundary Prediction Model}

\noindent \textbf{Models \& Details.} 
We fine-tune the pretrained MLLM Qwen2.5-VL-3B~\citep{bai2025qwen2} on our curated MLDR dataset, which contains 23,584 multi-modal long-form dialogues with manually annotated fragment boundaries. To efficiently adapt the model while preserving its general conversational and reasoning abilities, we apply parameter-efficient LoRA tuning and implement the entire training workflow using the ms-swift framework. During training, the model receives the full multi-turn dialogue as input and is supervised to output a set of fragment boundaries, formatted as clusters of utterance IDs (e.g., [[0,1,2], [3,4,5]]). This supervision encourages the model to identify coherent discourse units that align with natural conversational segmentation.

For evaluation, we reserve 10\% of the MLDR dataset as a held-out test set. The predicted fragment clusters are compared against the ground-truth annotations to assess how accurately the fine-tuned model captures semantic boundaries within long-form dialogues and generalizes to unseen conversations.

\noindent \textbf{Metrics.} 
We evaluate the quality of predicted dialogue fragmentations using two clustering-based metrics: Pairwise Precision/Recall/F1~\citep{shi2018face} and B-cubed Precision/Recall/F1~\citep{gunther2025jina}. Both metrics compute precision, recall, and F1-score, which are harmonic means of precision and recall. The key difference between them lies in the granularity of the evaluation (see Appendix.\ref{Metrics of Fragment Boundary Prediction Model} for details).

\noindent \textbf{Performance Comparison.} 
The evaluation results of our model and the baseline models (Qwen2.5-VL-72B and GPT-4o) on the MLDR dataset in long-form dialogue decomposition, as shown in Table~\ref{tab:decom}, demonstrate that our model outperforms the baseline models across all Pairwise and B-cubed metrics. Specifically, our model achieves 81.2\% precision, 92.2\% recall, and 83.6\% F1 score for Pairwise metrics, and 83.8\% precision, 94.2\% recall, and 86.7\% F1 score for B-cubed metrics. These results indicate the effectiveness of our model in decomposing long-form dialogues into semantically coherent fragments, ensuring high precision and recall in both global and local content alignment. Such accurate fragmentation provides clean, well-structured semantic units, which serve as a reliable foundation for the subsequent embedding, indexing, and fine-grained retrieval stages of the FFRS pipeline.

\begin{table*}[t]
\setlength{\abovecaptionskip}{1pt}
\setlength{\belowcaptionskip}{1pt}
\centering
\caption{Performance comparison of different models in long-form dialogue decomposition in terms of Pairwise and B-cubed precision, recall, and F1 scores.} \label{tab:decom}
\renewcommand\arraystretch{1.25}
\resizebox{0.8\linewidth}{!}{
\begin{tabular}{r|ccc|ccc} 
\toprule
\multirow{2}{*}{Model} & \multicolumn{3}{c|}{Pairwise}                 & \multicolumn{3}{c}{B-cubed}                    \\
& Precision(\%)     & Recall(\%)        & F1(\%)            & Precision(\%)     & Recall(\%)        & F1(\%)             \\ 
\hline
Qwen2.5-VL-72B~\citep{bai2025qwen2}         & 58.4          & 66.9          & 60.3          & 60.5          & 68.1          & 61.1           \\
GPT-4o~\citep{jaech2024openai}                 & 77.5          & 84.1          & 78.6          & 80.2          & 92.5          & 83.9           \\ 
\rowcolor[rgb]{0.951,0.951,0.951} \textbf{Our model-3B}  & \textbf{81.2} & \textbf{92.2} & \textbf{83.6} & \textbf{83.8} & \textbf{94.2} & \textbf{86.7}  \\
\bottomrule
\end{tabular}}
\end{table*}

\subsubsection{Evaluation of Fragment Embedding Model (FEM)}

\noindent \textbf{Models \& Details.}  
We adopt GME-Qwen2-VL-2B~\citep{zhang2024gme} as the backbone for our Fragment Embedding Model (FEM) and fine-tune it on our custom-built MLDR dataset. Each dialogue in MLDR contains multiple annotated ground-truth fragments, and each fragment is paired with a corresponding query. Based on this structure, we extract all fragment–query pairs, resulting in 62,654 Query–Fragment instances for training FEM’s global semantic alignment. Furthermore, to enable FEM to learn intra-fragment dependencies (see Eq.\ref{eq:intra-frag}), we decompose each fragment into multiple Question–Answer (QA) pairs, which serve as training examples for the local contrastive objective.

To rigorously evaluate the semantic representation capability of FEM, we construct a high-quality Query–Fragment test set specifically designed to measure the model’s ability to retrieve the correct fragment within the top-K search results. The construction process follows three steps: (i) Sentence Vector Extraction:
We encode every query using BERT to obtain high-dimensional semantic vectors that capture their underlying meanings. (ii) Topic Clustering: To ensure semantic diversity in the test set, we apply K-means clustering on all query vectors and set the number of clusters to 20. This guarantees that the test set covers a wide range of semantic themes. (iii) Semantic Diversity Sampling: We proportionally sample 500 Query–Fragment pairs across the clusters, ensuring that the test set reflects diverse semantic contexts and allows evaluation of the model’s robustness across different dialogue topics and structures.

We compare FEM against several mainstream multi-modal embedding models, including Jina-embeddings-v4~\citep{gunther2025jina}, VLM2Vec~\citep{jiang2024vlm2vec}, LamRA~\citep{liu2025lamra}, MonoQwen-ReRanker~\citep{MonoQwen}, and GME-Qwen2-VL-2B~\citep{zhang2024gme}.

\noindent \textbf{Evaluation Protocol \& Metrics.}
To evaluate the proposed embedding model, we adopt a Query--Fragment retrieval setting. Given a query $q$ and its ground-truth fragment $f^\ast$, the model encodes $q$ into an embedding vector $\mathbf{e}_q$, and encodes all candidate fragments into $\{\mathbf{e}_{f_j}\}$.  
We compute cosine similarity scores between the query embedding and each fragment embedding, and rank all fragments accordingly. Retrieval is considered successful if the ground-truth fragment $f^\ast$ appears in the Top-$K$ ranked results. The evaluation metric is Recall@$K$, where $K \in \{1, 5, 10, 15, 20\}$. This metric directly evaluates both the semantic fidelity of the learned embeddings—whether relevant queries and fragments are mapped closely in the embedding space—and the practical retrieval accuracy of the model under different levels of search depth.

\begin{table*}[t]
\setlength{\abovecaptionskip}{1pt}
\setlength{\belowcaptionskip}{1pt}
\centering
\caption{Performance comparison of various multi-modal embedding models on the Query–Fragment retrieval task. Results are reported in terms of Recall@K. ``$\dagger$'' indicates zero-shot inference without Query–Fragment data fine-tuning.
} \label{tab:embed}
\renewcommand\arraystretch{1.25}
\resizebox{1.0\linewidth}{!}{
\begin{tabular}{r|ccccc} 
\toprule
Multi-modal Embedding Model                                                      & Recall@1(\%) & Recall@5(\%) & Recall@10(\%) & Recall@15(\%) & Recall@20(\%)  \\ 
\hline
GME-Qwen2-VL(2B)$^\dagger$~\citep{zhang2024gme}         & 38.8                         & 65.0                         & 74.4                          & 79.4                          & 83.4                           \\
MonoQwen-ReRanker(2B)$^\dagger$~\citep{MonoQwen}   & 27.6                         & 56.0                         & 70.6                          & 76.8                          & 80.4                           \\
Jina-embeddings-v4(3B)$^\dagger$~\citep{gunther2025jina} & 21.2                         & 48.8                         & 57.6                          & 64.3                          & 69.0                           \\
VLM2Vec(7B)$^\dagger$~\citep{jiang2024vlm2vec}                 & 29.8                         & 47.8                         & 54.5                          & 62.1                          & 67.5                           \\
LamRA(7B)$^\dagger$~\citep{liu2025lamra}               & 28.2                         & 47.6                         & 53.2                          & 61.3                          & 66.8                           \\ 
\hline
GME-Qwen2-VL w/ Inter-Fragment Contrastive               & \underline{58.0}                 & \underline{87.8}                 & \underline{92.4}                  & \underline{96.0}                  & \underline{97.2}                   \\
GME-Qwen2-VL w/ Intra-Fragment Contrastive               & 52.6                         & 83.4                         & 91.0                          & 93.8                          & 95.8                           \\
\rowcolor[rgb]{0.951,0.951,0.951} \textbf{FEM (Ours)}                                        & \textbf{62.8}                & \textbf{89.6}                & \textbf{94.2}                 & \textbf{97.4}                 & \textbf{98.6}                  \\
\bottomrule
\end{tabular}}
\end{table*}

\begin{table}[t]
\setlength{\abovecaptionskip}{1pt}
\setlength{\belowcaptionskip}{1pt}
\centering
\caption{Comparison of average cosine similarity scores between query embeddings and their paired fragment embeddings across different multi-modal embedding models, where higher similarity indicates better semantic alignment.} \label{tab:cos}
\renewcommand\arraystretch{1.25}
\resizebox{1.0\linewidth}{!}{
\begin{tabular}{c|cccc} 
\toprule
Model & GME    & Jina-embeddings-v4 & LamRA  & FEM              \\ 
\hline
Cosine Similarity & 0.3916 & 0.3711             & 0.2937 & \textbf{0.4374}  \\
\bottomrule
\end{tabular}}
\end{table}

\noindent \textbf{Performance Comparison.}
Table \ref{tab:embed} reports the results of our Fragment Embedding Model (FEM) and several representative multi-modal embedding baselines on the Query–Fragment retrieval task. The baselines (marked with $^\dagger$) are evaluated in a zero-shot setting, without any task-specific fine-tuning. Across all Recall@$K$ metrics, our FEM substantially outperforms the zero-shot embedding models as well as the ablated variants. Overall, FEM achieves 62.8\% Recall@1, outperforming the strongest zero-shot baseline (GME-Qwen2-VL-2B) by +24.0 percentage points, and surpassing other models by an even larger margin (e.g., +41.6 over Jina-embeddings-v4). At larger retrieval depths, the advantage becomes more pronounced: FEM reaches 89.6\% Recall@5, 94.2\% Recall@10, and exceeds 98\% performance at Recall@20, indicating that nearly all ground-truth fragments are successfully recalled within the top retrieved candidates.

In addition to retrieval accuracy, we further analyze the intrinsic quality of fragment embeddings by measuring the average cosine similarity between each query and its ground-truth fragment embedding. As shown in Table~\ref{tab:cos}, existing multi-modal embedding models, such as GM~\citep{zhang2024gme}, Jina-embeddings-v4~\citep{gunther2025jina}, and LamRA~\citep{liu2025lamra}, exhibit relatively low semantic alignment between Query–Fragment pairs. In contrast, our FEM achieves the highest similarity score of 0.4374, indicating that the learned embeddings preserve stronger semantic correspondence even before retrieval ranking is applied. This higher intrinsic alignment provides a complementary perspective to the Recall@$K$ results in Table~\ref{tab:embed}, further demonstrating that FEM produces more discriminative and semantically coherent fragment representations.

\noindent \textbf{Ablation Study.}
We further evaluate two ablated variants that isolate the contributions of the proposed contrastive objectives. The inter-fragment contrast improves global semantic alignment (58.0\% Recall@1), while the intra-fragment contrast enhances local structural coherence (52.6\% Recall@1). However, neither alone matches the full FEM, demonstrating that both global and local contrastive signals are essential and complementary. Their combination yields the best performance across all settings. These results collectively indicate that our dual-level contrastive learning enables FEM to (i) learn highly discriminative fragment-level embeddings, (ii) better preserve discourse structure, and (iii) robustly generalize across diverse dialogue topics. This strong retrieval capability is critical for the downstream stages of FFRS, where accurate fragment recall directly influences fine-grained reasoning quality.

\subsubsection{Evaluation of Overall System}

\noindent\textbf{System Setup.}
Based on the components described in Sec.\ref{FFRS}, we deploy a full multi-modal fine-grained fragment retrieval system that integrates (i) multi-modal long-dialogue decomposition, (ii) fragment embedding and offline indexing, (iii) online coarse recall, (iv) fine-grained reasoning with F$^2$RVLM, and (v) dialogue-level fragment aggregation. The goal is to assess the system’s end-to-end capability in performing accurate and efficient fragment retrieval over large-scale real-world corpora.

\noindent\textbf{Evaluation Corpus \& Deployment.}
To evaluate the practical performance of the proposed FFRS system, we utilize a real-world multi-modal dialogue corpus from our previously constructed WeChat test set. The corpus contains 537 authentic long-form conversations, each with more than 100 utterances on average and covering diverse daily-life and workplace scenarios with rich multi-modal content.  

The entire system is deployed on 8$\times$NVIDIA H20 GPUs (PCIe, 96GB). For each test case, a user-issued query and the full multi-modal corpus are jointly processed by the FFRS pipeline, which retrieves and ranks semantically relevant fragments across all dialogues in the corpus.

\noindent\textbf{Efficiency Comparison.}
We compare the retrieval latency of FFRS with a brute-force baseline that applies the generative multi-modal model F$^{2}$RVLM directly to the entire corpus, performing inference on each long dialogue one by one without any coarse filtering. It represents the naive but commonly adopted strategy where a generative VLM is applied directly for fragment-level reasoning without retrieval. As summarized in Table~\ref{tab:system}, FFRS achieves an average end-to-end query latency of only 18 seconds, whereas F$^{2}$RVLM requires approximately 720 seconds under the same hardware conditions—yielding a 40$\times$ speedup. These results demonstrate that the proposed two-stage pipeline (coarse retrieval followed by fine-grained reasoning) substantially improves retrieval efficiency in large-scale multi-modal dialogue corpora while maintaining strong performance.

\begin{table}[t]
\setlength{\abovecaptionskip}{1pt}
\setlength{\belowcaptionskip}{1pt}
\centering
\caption{Retrieval latency comparison between FFRS and a brute-force generative baseline in a large-scale corpus.}
\label{tab:system}
\renewcommand\arraystretch{1.25}
\resizebox{0.65\linewidth}{!}{
\begin{tabular}{cc}
\toprule
Model & Avg. Latency per Query (s) \\
\hline
F$^{2}$RVLM & 720 \\
FFRS & 18 \\
\bottomrule
\end{tabular}}
\end{table}

\begin{table}[t]
\centering
\caption{Human evaluation of retrieval quality on 100 real user queries. Each criterion is assessed using a five-point Likert scale (1 = very poor, 5 = excellent).}
\label{tab:human}
\renewcommand\arraystretch{1.25}
\resizebox{0.9\linewidth}{!}{
\begin{tabular}{ccc} 
\toprule
Evaluation Criterion      & Mean Score & Std. Dev.  \\ 
\midrule
Query–Fragment Relevance  & 4.22       & 0.41       \\
Dialogue-Level Coverage   & 4.09       & 0.47       \\
Overall Retrieval Quality & 4.16       & 0.39       \\
\bottomrule
\end{tabular}}
\end{table}

\noindent\textbf{Human Evaluation of Retrieval Quality.}
Since the WeChat dialogue corpus contains a large number of conversations and user queries are issued freely, it is not feasible to obtain definitive ground-truth fragments for each query. To address this, we conducted a controlled human evaluation to quantitatively assess the semantic correctness and coverage of the retrieved fragments. Specifically, we recruited 10 volunteers, each asked to provide 10 real information-seeking queries, resulting in 100 query instances. For each query, FFRS returned several semantically relevant fragments from the corpus. A trained annotation team then evaluated the retrieval results according to two criteria: (i) Query–Fragment Relevance: Whether each retrieved fragment accurately reflects the semantic intent of the user query. (ii) Dialogue-Level Coverage: Whether the retrieved fragments capture the major query-relevant content within the source dialogue.

Both criteria were assessed utilizing a five-point Likert scale (1 = very poor, 5 = excellent), with higher scores indicating better performance. The final results are summarized in Table~\ref{tab:human}. Overall, the proposed system demonstrates strong semantic relevance and satisfactory coverage across diverse real-world queries, indicating that FFRS is capable of reliably returning high-quality fragments that align with user intent.

Beyond quantitative metrics, qualitative inspection further confirms the effectiveness of the proposed system. As illustrated in the qualitative examples provided in the Appendix.\ref{Qualitative Results of FFRS in Corpus-level FFR}, the fragments retrieved by FFRS exhibit clear semantic alignment with user-issued queries, accurately capturing both the explicit intent and the underlying contextual cues. These examples highlight that FFRS consistently locates the most relevant portions of long multi-modal conversations, even when the supporting evidence is dispersed across multiple turns. Taken together, the qualitative findings underscore that FFRS not only achieves substantial efficiency improvements but also preserves high retrieval fidelity across large-scale, real-world multi-modal dialogue corpora.

\section{Conclusion}
This work introduces the task of Fine-grained Fragment Retrieval (FFR), which targets retrieving semantically coherent multi-turn fragments, including both utterances and images, from long-form multi-modal dialogues. We explore two key settings: FFR within a single dialogue and FFR across a dialogue corpus, reflecting both focused and open-domain retrieval needs. To address these, we develop tailored solutions, F$^2$RVLM for intra-dialogue retrieval and FFRS for corpus-level retrieval. We also construct the MLDR dataset and a real-world WeChat test set to support this task. Extensive experiments demonstrate the effectiveness of our models, setting a solid foundation for future research on multi-modal dialogue understanding and retrieval. Moreover, the proposed methods show strong potential for integration into real-world multi-modal dialogue systems, enabling more context-aware and user-aligned information access.

\section*{Data Availability Statement}
The datasets generated in this study are publicly available at \url{https://github.com/HanboBizl/FFRS.github.io}.

\section*{Acknowledgments}
This work was part of Hanbo Bi's research during his internship at Tencent WXG, under the guidance of Zhiqiang Yuan, and both made equivalent contributions. 
We also acknowledge the use of the publicly available multi-modal dialogue datasets DialogCC~\citep{lee2024dialogcc} and MMDialog~\citep{feng2023mmdialog}.

\bibliographystyle{spbasic}      
\bibliography{3dfsl_main}   

\begin{thebibliography}{69}
\providecommand{\natexlab}[1]{#1}
\providecommand{\url}[1]{{#1}}
\providecommand{\urlprefix}{URL }
\expandafter\ifx\csname urlstyle\endcsname\relax
  \providecommand{\doi}[1]{DOI~\discretionary{}{}{}#1}\else
  \providecommand{\doi}{DOI~\discretionary{}{}{}\begingroup
  \urlstyle{rm}\Url}\fi
\providecommand{\eprint}[2][]{\url{#2}}

\bibitem[{Achiam et~al.(2023)Achiam, Adler, Agarwal, Ahmad, Akkaya, Aleman,
  Almeida, Altenschmidt, Altman, Anadkat et~al.}]{achiam2023gpt}
Achiam J, Adler S, Agarwal S, Ahmad L, Akkaya I, Aleman FL, Almeida D,
  Altenschmidt J, Altman S, Anadkat S, et~al. (2023) Gpt-4 technical report.
  arXiv:230308774

\bibitem[{{Anthropic}(2025)}]{claude4.5}
{Anthropic} (2025) Introducing claude sonnet 4.5.
  \url{https://www.anthropic.com/news/claude-sonnet-4-5}

\bibitem[{Bai et~al.(2025{\natexlab{a}})Bai, Chen, Liu, Wang, Ge, Song, Dang,
  Wang, Wang, Tang et~al.}]{bai2025qwen2}
Bai S, Chen K, Liu X, Wang J, Ge W, Song S, Dang K, Wang P, Wang S, Tang J,
  et~al. (2025{\natexlab{a}}) Qwen2. 5-vl technical report. arXiv preprint
  arXiv:250213923

\bibitem[{Bai et~al.(2025{\natexlab{b}})Bai, Ji, Cao, Wang, and
  Ye}]{bai2025chat}
Bai Y, Ji Y, Cao M, Wang J, Ye M (2025{\natexlab{b}}) Chat-based person
  retrieval via dialogue-refined cross-modal alignment. In: Proceedings of the
  Computer Vision and Pattern Recognition Conference, pp 3952--3962

\bibitem[{BehnamGhader et~al.(2024)BehnamGhader, Adlakha, Mosbach, Bahdanau,
  Chapados, and Reddy}]{behnamghader2024llm2vec}
BehnamGhader P, Adlakha V, Mosbach M, Bahdanau D, Chapados N, Reddy S (2024)
  Llm2vec: Large language models are secretly powerful text encoders. arXiv
  preprint arXiv:240405961

\bibitem[{Bi et~al.(2025)Bi, Yuan, Jia, Zhang, Li, Luo, Deng, Duan, and
  Zhang}]{bi2025f2rvlm}
Bi H, Yuan Z, Jia Z, Zhang J, Li C, Luo P, Deng Y, Duan X, Zhang J (2025)
  F2rvlm: Boosting fine-grained fragment retrieval for multi-modal long-form
  dialogue with vision language model. arXiv preprint arXiv:250817714

\bibitem[{Bischoff and Graefe(2002)}]{bischoff2002dependable}
Bischoff R, Graefe V (2002) Dependable multimodal communication and interaction
  with robotic assistants. In: Proceedings. 11th IEEE International Workshop on
  Robot and Human Interactive Communication, IEEE, pp 300--305

\bibitem[{Chaffin and Lac(2024)}]{MonoQwen}
Chaffin A, Lac A (2024) Monoqwen: Visual document reranking.
  \urlprefix\url{https://huggingface.co/lightonai/MonoQwen2-VL-v0.1}

\bibitem[{Chen et~al.(2024{\natexlab{a}})Chen, Song, Zuo, Wei, Nie, and
  Chua}]{chen2024domain}
Chen X, Song X, Zuo J, Wei Y, Nie L, Chua TS (2024{\natexlab{a}}) Domain-aware
  multimodal dialog systems with distribution-based user characteristic
  modeling. ACM Transactions on Multimedia Computing, Communications and
  Applications 21(2):1--22

\bibitem[{Chen et~al.(2022)Chen, Fan, Xing, Pang, Huang, Han, Tie, and
  Xu}]{chen2022cped}
Chen Y, Fan W, Xing X, Pang J, Huang M, Han W, Tie Q, Xu X (2022) {CPED}: A
  large-scale chinese personalized and emotional dialogue dataset for
  conversational ai. arXiv:220514727
  \urlprefix\url{https://arxiv.org/abs/2205.14727}

\bibitem[{Chen et~al.(2024{\natexlab{b}})Chen, Wu, Wang, Su, Chen, Xing, Zhong,
  Zhang, Zhu, Lu et~al.}]{chen2024internvl}
Chen Z, Wu J, Wang W, Su W, Chen G, Xing S, Zhong M, Zhang Q, Zhu X, Lu L,
  et~al. (2024{\natexlab{b}}) Internvl: Scaling up vision foundation models and
  aligning for generic visual-linguistic tasks. In: Proceedings of the IEEE/CVF
  Conference on Computer Vision and Pattern Recognition, pp 24185--24198

\bibitem[{Comanici et~al.(2025)Comanici, Bieber, Schaekermann, Pasupat,
  Sachdeva, Dhillon, Blistein, Ram, Zhang, Rosen et~al.}]{comanici2025gemini}
Comanici G, Bieber E, Schaekermann M, Pasupat I, Sachdeva N, Dhillon I,
  Blistein M, Ram O, Zhang D, Rosen E, et~al. (2025) Gemini 2.5: Pushing the
  frontier with advanced reasoning, multimodality, long context, and next
  generation agentic capabilities. arXiv:250706261

\bibitem[{Feng et~al.(2023)Feng, Sun, Xu, Zhao, Yang, Tao, Zhao, and
  Lin}]{feng2023mmdialog}
Feng J, Sun Q, Xu C, Zhao P, Yang Y, Tao C, Zhao D, Lin Q (2023) Mmdialog: A
  large-scale multi-turn dialogue dataset towards multi-modal open-domain
  conversation. In: Proceedings of the 61st Annual Meeting of the Association
  for Computational Linguistics (Volume 1: Long Papers), pp 7348--7363

\bibitem[{Feng et~al.(2025)Feng, Gong, Li, Guo, Wang, Peng, Wu, Zhang, Wang,
  and Yue}]{feng2025video}
Feng K, Gong K, Li B, Guo Z, Wang Y, Peng T, Wu J, Zhang X, Wang B, Yue X
  (2025) Video-r1: Reinforcing video reasoning in mllms. arXiv:250321776

\bibitem[{G{\"u}nther et~al.(2025)G{\"u}nther, Sturua, Akram, Mohr, Ungureanu,
  Wang, Eslami, Martens, Werk, Wang et~al.}]{gunther2025jina}
G{\"u}nther M, Sturua S, Akram MK, Mohr I, Ungureanu A, Wang B, Eslami S,
  Martens S, Werk M, Wang N, et~al. (2025) jina-embeddings-v4: Universal
  embeddings for multimodal multilingual retrieval. In: Proceedings of the 5th
  Workshop on Multilingual Representation Learning (MRL 2025), pp 531--550

\bibitem[{Guo et~al.(2025{\natexlab{a}})Guo, Wu, Zhu, Leng, Shi, Chen, Fan,
  Wang, Jiang, Wang et~al.}]{guo2025seed1}
Guo D, Wu F, Zhu F, Leng F, Shi G, Chen H, Fan H, Wang J, Jiang J, Wang J,
  et~al. (2025{\natexlab{a}}) Seed1. 5-vl technical report. arXiv:250507062

\bibitem[{Guo et~al.(2025{\natexlab{b}})Guo, Yang, Zhang, Song, Zhang, Xu, Zhu,
  Ma, Wang, Bi et~al.}]{guo2025deepseek}
Guo D, Yang D, Zhang H, Song J, Zhang R, Xu R, Zhu Q, Ma S, Wang P, Bi X,
  et~al. (2025{\natexlab{b}}) Deepseek-r1: Incentivizing reasoning capability
  in llms via reinforcement learning. arXiv:250112948

\bibitem[{Hu et~al.(2022)Hu, Shen, Wallis, Allen-Zhu, Li, Wang, Wang, Chen
  et~al.}]{hu2022lora}
Hu EJ, Shen Y, Wallis P, Allen-Zhu Z, Li Y, Wang S, Wang L, Chen W, et~al.
  (2022) Lora: Low-rank adaptation of large language models. ICLR 1(2):3

\bibitem[{Huang et~al.(2025)Huang, Jia, Zhai, Cao, Ye, Zhao, Xu, Hu, and
  Lin}]{huang2025vision}
Huang W, Jia B, Zhai Z, Cao S, Ye Z, Zhao F, Xu Z, Hu Y, Lin S (2025)
  Vision-r1: Incentivizing reasoning capability in multimodal large language
  models. arXiv:250306749

\bibitem[{Jaech et~al.(2024)Jaech, Kalai, Lerer, Richardson, El-Kishky, Low,
  Helyar, Madry, Beutel, Carney et~al.}]{jaech2024openai}
Jaech A, Kalai A, Lerer A, Richardson A, El-Kishky A, Low A, Helyar A, Madry A,
  Beutel A, Carney A, et~al. (2024) Openai o1 system card. arXiv:241216720

\bibitem[{Jia et~al.(2021)Jia, Yang, Xia, Chen, Parekh, Pham, Le, Sung, Li, and
  Duerig}]{jia2021scaling}
Jia C, Yang Y, Xia Y, Chen YT, Parekh Z, Pham H, Le Q, Sung YH, Li Z, Duerig T
  (2021) Scaling up visual and vision-language representation learning with
  noisy text supervision. In: International conference on machine learning,
  PMLR, pp 4904--4916

\bibitem[{Jiang et~al.(2024{\natexlab{a}})Jiang, Song, Zhang, Huang, Deng, Sun,
  Zhang, Wang, and Zhuang}]{jiang2024e5}
Jiang T, Song M, Zhang Z, Huang H, Deng W, Sun F, Zhang Q, Wang D, Zhuang F
  (2024{\natexlab{a}}) E5-v: Universal embeddings with multimodal large
  language models. arXiv:240712580

\bibitem[{Jiang et~al.(2024{\natexlab{b}})Jiang, Meng, Yang, Yavuz, Zhou, and
  Chen}]{jiang2024vlm2vec}
Jiang Z, Meng R, Yang X, Yavuz S, Zhou Y, Chen W (2024{\natexlab{b}}) Vlm2vec:
  Training vision-language models for massive multimodal embedding tasks.
  arXiv:241005160

\bibitem[{Lee et~al.(2021)Lee, Shin, Choo, Choi, and
  Myaeng}]{lee2021constructing}
Lee N, Shin S, Choo J, Choi HJ, Myaeng SH (2021) Constructing multi-modal
  dialogue dataset by replacing text with semantically relevant images. In:
  Proceedings of the 59th Annual Meeting of the Association for Computational
  Linguistics and the 11th International Joint Conference on Natural Language
  Processing (Volume 2: Short Papers), pp 897--906

\bibitem[{Lee et~al.(2024)Lee, Ko, Kim, Hyeon, and Choi}]{lee2024dialogcc}
Lee YJ, Ko B, Kim HG, Hyeon J, Choi HJ (2024) Dialogcc: An automated pipeline
  for creating high-quality multi-modal dialogue dataset. In: Proceedings of
  the 2024 Conference of the North American Chapter of the Association for
  Computational Linguistics: Human Language Technologies (Volume 1: Long
  Papers), pp 1938--1963

\bibitem[{Li et~al.(2023)Li, Li, Savarese, and Hoi}]{li2023blip}
Li J, Li D, Savarese S, Hoi S (2023) Blip-2: Bootstrapping language-image
  pre-training with frozen image encoders and large language models. In:
  International conference on machine learning, PMLR, pp 19730--19742

\bibitem[{Lin et~al.(2023)Lin, Ruan, Xia, Liu, Wen, Xu, Hu, Song, Zhao, Jin
  et~al.}]{lin2023tiktalk}
Lin H, Ruan L, Xia W, Liu P, Wen J, Xu Y, Hu D, Song R, Zhao WX, Jin Q, et~al.
  (2023) Tiktalk: a video-based dialogue dataset for multi-modal chitchat in
  real world. In: Proceedings of the 31st ACM International Conference on
  Multimedia, pp 1303--1313

\bibitem[{Lin et~al.(2025)Lin, Lee, Shoeybi, Lin, Catanzaro, and Ping}]{linmm}
Lin SC, Lee C, Shoeybi M, Lin J, Catanzaro B, Ping W (2025) Mm-embed: Universal
  multimodal retrieval with multimodal llms. In: The Thirteenth International
  Conference on Learning Representations

\bibitem[{Liu et~al.(2023{\natexlab{a}})Liu, Li, Wu, and Lee}]{liu2023visual}
Liu H, Li C, Wu Q, Lee YJ (2023{\natexlab{a}}) Visual instruction tuning.
  Advances in neural information processing systems 36:34892--34916

\bibitem[{Liu et~al.(2023{\natexlab{b}})Liu, Li, Wu, and Lee}]{liu2023llava}
Liu H, Li C, Wu Q, Lee YJ (2023{\natexlab{b}}) Visual instruction tuning. In:
  NeurIPS

\bibitem[{Liu et~al.(2025{\natexlab{a}})Liu, Zhang, Cai, Jiang, Hu, Yao, Wang,
  and Xie}]{liu2025lamra}
Liu Y, Zhang Y, Cai J, Jiang X, Hu Y, Yao J, Wang Y, Xie W (2025{\natexlab{a}})
  Lamra: Large multimodal model as your advanced retrieval assistant. In:
  Proceedings of the Computer Vision and Pattern Recognition Conference, pp
  4015--4025

\bibitem[{Liu et~al.(2022)Liu, Xiong, Lv, Liu, and Yu}]{liuuniversal}
Liu Z, Xiong C, Lv Y, Liu Z, Yu G (2022) Universal vision-language dense
  retrieval: Learning a unified representation space for multi-modal retrieval.
  In: The Eleventh International Conference on Learning Representations

\bibitem[{Liu et~al.(2025{\natexlab{b}})Liu, Sun, Zang, Dong, Cao, Duan, Lin,
  and Wang}]{liu2025visual}
Liu Z, Sun Z, Zang Y, Dong X, Cao Y, Duan H, Lin D, Wang J (2025{\natexlab{b}})
  Visual-rft: Visual reinforcement fine-tuning. arXiv:250301785

\bibitem[{Lu et~al.(2024)Lu, Li, Chen, Xu, Luo, Zhang, and Ye}]{lu2024ovis}
Lu S, Li Y, Chen QG, Xu Z, Luo W, Zhang K, Ye HJ (2024) Ovis: Structural
  embedding alignment for multimodal large language model. arXiv:240520797

\bibitem[{Meng et~al.(2024)Meng, Liu, Joty, Xiong, Zhou, and
  Yavuz}]{meng2024sfr}
Meng R, Liu Y, Joty SR, Xiong C, Zhou Y, Yavuz S (2024) Sfr-embedding-2:
  Advanced text embedding with multi-stage training. Last accessed: Nov 24th

\bibitem[{Meng et~al.(2025)Meng, Jiang, Liu, Su, Yang, Fu, Qin, Chen, Xu, Xiong
  et~al.}]{meng2025vlm2vec}
Meng R, Jiang Z, Liu Y, Su M, Yang X, Fu Y, Qin C, Chen Z, Xu R, Xiong C,
  et~al. (2025) Vlm2vec-v2: Advancing multimodal embedding for videos, images,
  and visual documents. arXiv preprint arXiv:250704590

\bibitem[{Meng et~al.(2020)Meng, Wang, Han, Sun, Wu, Yan, and
  Li}]{meng2020openvidial}
Meng Y, Wang S, Han Q, Sun X, Wu F, Yan R, Li J (2020) Openvidial: A
  large-scale, open-domain dialogue dataset with visual contexts.
  arXiv:201215015

\bibitem[{Mostafazadeh et~al.(2017)Mostafazadeh, Brockett, Dolan, Galley, Gao,
  Spithourakis, and Vanderwende}]{mostafazadeh2017image}
Mostafazadeh N, Brockett C, Dolan WB, Galley M, Gao J, Spithourakis G,
  Vanderwende L (2017) Image-grounded conversations: Multimodal context for
  natural question and response generation. In: Proceedings of the Eighth
  International Joint Conference on Natural Language Processing (Volume 1: Long
  Papers), pp 462--472

\bibitem[{Nie et~al.(2021)Nie, Wang, and Xiong}]{nie2021research}
Nie J, Wang Q, Xiong J (2021) Research on intelligent service of customer
  service system. Cognitive Computation and Systems 3(3):197--205

\bibitem[{Peng et~al.(2025)Peng, Zhang, Zhang, You, Liu, Zhu, Yang, Xu, Geng,
  and Yang}]{peng2025lmm}
Peng Y, Zhang G, Zhang M, You Z, Liu J, Zhu Q, Yang K, Xu X, Geng X, Yang X
  (2025) Lmm-r1: Empowering 3b lmms with strong reasoning abilities through
  two-stage rule-based rl. arXiv:250307536

\bibitem[{Radford et~al.(2021)Radford, Kim, Hallacy, Ramesh, Goh, Agarwal,
  Sastry, Askell, Mishkin, Clark et~al.}]{radford2021learning}
Radford A, Kim JW, Hallacy C, Ramesh A, Goh G, Agarwal S, Sastry G, Askell A,
  Mishkin P, Clark J, et~al. (2021) Learning transferable visual models from
  natural language supervision. In: International conference on machine
  learning, PmLR, pp 8748--8763

\bibitem[{Schulman et~al.(2017)Schulman, Wolski, Dhariwal, Radford, and
  Klimov}]{schulman2017proximal}
Schulman J, Wolski F, Dhariwal P, Radford A, Klimov O (2017) Proximal policy
  optimization algorithms. arXiv:170706347

\bibitem[{Shao et~al.(2024)Shao, Wang, Zhu, Xu, Song, Bi, Zhang, Zhang, Li, Wu
  et~al.}]{shao2024deepseekmath}
Shao Z, Wang P, Zhu Q, Xu R, Song J, Bi X, Zhang H, Zhang M, Li Y, Wu Y, et~al.
  (2024) Deepseekmath: Pushing the limits of mathematical reasoning in open
  language models. arXiv:240203300

\bibitem[{Shen et~al.(2025)Shen, Liu, Li, Fang, Ma, Liao, Shen, Zhang, Zhao,
  Zhang et~al.}]{shen2025vlm}
Shen H, Liu P, Li J, Fang C, Ma Y, Liao J, Shen Q, Zhang Z, Zhao K, Zhang Q,
  et~al. (2025) Vlm-r1: A stable and generalizable r1-style large
  vision-language model. arXiv:250407615

\bibitem[{Shi et~al.(2018)Shi, Otto, and Jain}]{shi2018face}
Shi Y, Otto C, Jain AK (2018) Face clustering: representation and pairwise
  constraints. IEEE Transactions on Information Forensics and Security
  13(7):1626--1640

\bibitem[{Shuster et~al.(2020)Shuster, Humeau, Bordes, and
  Weston}]{shuster2020image}
Shuster K, Humeau S, Bordes A, Weston J (2020) Image-chat: Engaging grounded
  conversations. In: Proceedings of the 58th Annual Meeting of the Association
  for Computational Linguistics, pp 2414--2429

\bibitem[{Su et~al.(2023)Su, Shi, Kasai, Wang, Hu, Ostendorf, Yih, Smith,
  Zettlemoyer, and Yu}]{su2023one}
Su H, Shi W, Kasai J, Wang Y, Hu Y, Ostendorf M, Yih Wt, Smith NA, Zettlemoyer
  L, Yu T (2023) One embedder, any task: Instruction-finetuned text embeddings.
  In: Findings of the Association for Computational Linguistics: ACL 2023, pp
  1102--1121

\bibitem[{Tan et~al.(2025)Tan, Ji, Hao, Lin, Wang, Wang, and
  Zhang}]{tan2025reason}
Tan H, Ji Y, Hao X, Lin M, Wang P, Wang Z, Zhang S (2025) Reason-rft:
  Reinforcement fine-tuning for visual reasoning. arXiv:250320752

\bibitem[{Tomar and Kakkar(2014)}]{tomar2014maturity}
Tomar A, Kakkar A (2014) Maturity model for features of social messaging
  applications. In: Proceedings of 3rd International Conference on Reliability,
  Infocom Technologies and Optimization, IEEE, pp 1--6

\bibitem[{Walnycky et~al.(2015)Walnycky, Baggili, Marrington, Moore, and
  Breitinger}]{walnycky2015network}
Walnycky D, Baggili I, Marrington A, Moore J, Breitinger F (2015) Network and
  device forensic analysis of android social-messaging applications. Digital
  Investigation 14:S77--S84

\bibitem[{Wang et~al.(2025)Wang, Qu, Huang, Chu, Lin, and Chen}]{wang2025vl}
Wang H, Qu C, Huang Z, Chu W, Lin F, Chen W (2025) Vl-rethinker: Incentivizing
  self-reflection of vision-language models with reinforcement learning.
  arXiv:250408837

\bibitem[{Wang et~al.(2024{\natexlab{a}})Wang, Yang, Huang, Yang, Majumder, and
  Wei}]{wang2024improving}
Wang L, Yang N, Huang X, Yang L, Majumder R, Wei F (2024{\natexlab{a}})
  Improving text embeddings with large language models. In: Proceedings of the
  62nd Annual Meeting of the Association for Computational Linguistics (Volume
  1: Long Papers), pp 11897--11916

\bibitem[{Wang et~al.(2024{\natexlab{b}})Wang, Bai, Tan, Wang, Fan, Bai, Chen,
  Liu, Wang, Ge et~al.}]{wang2024qwen2}
Wang P, Bai S, Tan S, Wang S, Fan Z, Bai J, Chen K, Liu X, Wang J, Ge W, et~al.
  (2024{\natexlab{b}}) Qwen2-vl: Enhancing vision-language model's perception
  of the world at any resolution. arXiv:240912191

\bibitem[{Wei et~al.(2024)Wei, Chen, Chen, Hu, Zhang, Fu, Ritter, and
  Chen}]{wei2024uniir}
Wei C, Chen Y, Chen H, Hu H, Zhang G, Fu J, Ritter A, Chen W (2024) Uniir:
  Training and benchmarking universal multimodal information retrievers. In:
  European Conference on Computer Vision, Springer, pp 387--404

\bibitem[{Wu et~al.(2024{\natexlab{a}})Wu, Chen, Pan, Liu, Liu, Dai, Gao, Ma,
  Wu, Wang et~al.}]{wu2024deepseek}
Wu Z, Chen X, Pan Z, Liu X, Liu W, Dai D, Gao H, Ma Y, Wu C, Wang B, et~al.
  (2024{\natexlab{a}}) Deepseek-vl2: Mixture-of-experts vision-language models
  for advanced multimodal understanding. arXiv:241210302

\bibitem[{Wu et~al.(2024{\natexlab{b}})Wu, She, and Zhou}]{wu2024intelligent}
Wu Z, She Q, Zhou C (2024{\natexlab{b}}) Intelligent customer service system
  optimization based on artificial intelligence. Journal of Organizational and
  End User Computing (JOEUC) 36(1):1--27

\bibitem[{Xiaomi(2025)}]{coreteam2025mimovltechnicalreport}
Xiaomi LCT (2025) Mimo-vl technical report.
  \urlprefix\url{https://arxiv.org/abs/2506.03569}, \eprint{2506.03569}

\bibitem[{Yang et~al.(2025{\natexlab{a}})Yang, Li, Yang, Zhang, Hui, Zheng, Yu,
  Gao, Huang, Lv et~al.}]{yang2025qwen3}
Yang A, Li A, Yang B, Zhang B, Hui B, Zheng B, Yu B, Gao C, Huang C, Lv C,
  et~al. (2025{\natexlab{a}}) Qwen3 technical report. arXiv:250509388

\bibitem[{Yang et~al.(2025{\natexlab{b}})Yang, Peng, Gao, Wang, Huang, and
  Deng}]{yang2025deep}
Yang Z, Peng T, Gao C, Wang C, Huang H, Deng Y (2025{\natexlab{b}}) A deep dive
  into retrieval-augmented generation for code completion: Experience on
  wechat. arXiv preprint arXiv:250718515

\bibitem[{Ye et~al.(2024)Ye, Xu, Liu, Hu, Yan, Qian, Zhang, Huang, and
  Zhou}]{ye2024mplugowl3longimagesequenceunderstanding}
Ye J, Xu H, Liu H, Hu A, Yan M, Qian Q, Zhang J, Huang F, Zhou J (2024)
  mplug-owl3: Towards long image-sequence understanding in multi-modal large
  language models. \urlprefix\url{https://arxiv.org/abs/2408.04840},
  \eprint{2408.04840}

\bibitem[{Yin et~al.(2024)Yin, Hui, Yang, Huang, and Li}]{yin2024dialclip}
Yin Z, Hui B, Yang M, Huang F, Li Y (2024) Dialclip: Empowering clip as
  multi-modal dialog retriever. In: ICASSP 2024-2024 IEEE International
  Conference on Acoustics, Speech and Signal Processing (ICASSP), IEEE, pp
  12421--12425

\bibitem[{Yu et~al.(2025)Yu, Zhang, Zhu, Yuan, Zuo, Yue, Dai, Fan, Liu, Liu
  et~al.}]{yu2025dapo}
Yu Q, Zhang Z, Zhu R, Yuan Y, Zuo X, Yue Y, Dai W, Fan T, Liu G, Liu L, et~al.
  (2025) Dapo: An open-source llm reinforcement learning system at scale.
  arXiv:250314476

\bibitem[{Yue et~al.(2025)Yue, Yuan, Yu, Zuo, Zhu, Xu, Chen, Wang, Fan, Du
  et~al.}]{yue2025vapo}
Yue Y, Yuan Y, Yu Q, Zuo X, Zhu R, Xu W, Chen J, Wang C, Fan T, Du Z, et~al.
  (2025) Vapo: Efficient and reliable reinforcement learning for advanced
  reasoning tasks. arXiv:250405118

\bibitem[{Zang et~al.(2021)Zang, Liu, Wang, Song, Zhang, and
  Chen}]{zang2021photochat}
Zang X, Liu L, Wang M, Song Y, Zhang H, Chen J (2021) Photochat: A human-human
  dialogue dataset with photo sharing behavior for joint image-text modeling.
  In: Proceedings of the 59th Annual Meeting of the Association for
  Computational Linguistics and the 11th International Joint Conference on
  Natural Language Processing (Volume 1: Long Papers), pp 6142--6152

\bibitem[{Zhang et~al.(2024)Zhang, Zhang, Xie, Li, Dai, Long, Xie, Zhang, Li,
  and Zhang}]{zhang2024gme}
Zhang X, Zhang Y, Xie W, Li M, Dai Z, Long D, Xie P, Zhang M, Li W, Zhang M
  (2024) Gme: Improving universal multimodal retrieval by multimodal llms.
  arXiv preprint arXiv:241216855

\bibitem[{Zhang et~al.(2023)Zhang, Yu, Guo, Wang, Zhao, Uprety, Song, Li, and
  Qin}]{zhang2023cmma}
Zhang Y, Yu Y, Guo Q, Wang B, Zhao D, Uprety S, Song D, Li Q, Qin J (2023)
  Cmma: benchmarking multi-affection detection in chinese multi-modal
  conversations. Advances in Neural Information Processing Systems
  36:18794--18805

\bibitem[{Zhao et~al.(2022)Zhao, Zhang, Hu, Liu, Jin, Wang, and
  Li}]{zhao2022m3ed}
Zhao J, Zhang T, Hu J, Liu Y, Jin Q, Wang X, Li H (2022) M3ed: Multi-modal
  multi-scene multi-label emotional dialogue database. In: Proceedings of the
  60th Annual Meeting of the Association for Computational Linguistics (Volume
  1: Long Papers), pp 5699--5710

\bibitem[{Zhao et~al.(2024)Zhao, Huang, Hu, Wang, Mao, Zhang, Jiang, Wu, Ai,
  Wang, Zhou, and Chen}]{zhao2024swiftascalablelightweightinfrastructure}
Zhao Y, Huang J, Hu J, Wang X, Mao Y, Zhang D, Jiang Z, Wu Z, Ai B, Wang A,
  Zhou W, Chen Y (2024) Swift:a scalable lightweight infrastructure for
  fine-tuning. \urlprefix\url{https://arxiv.org/abs/2408.05517},
  \eprint{2408.05517}

\bibitem[{Zheng et~al.(2022)Zheng, Chen, Liu, and Sun}]{zheng2022mmchat}
Zheng Y, Chen G, Liu X, Sun J (2022) Mmchat: Multi-modal chat dataset on social
  media. In: Proceedings of the Thirteenth Language Resources and Evaluation
  Conference, pp 5778--5786

\end{thebibliography}

\clearpage

\appendix

\section{Appendix Overview}

The Supplementary Materials offer additional insights and experimental details that support the main paper. The content is organized as follows:

\begin{itemize}
\item \textbf{Dataset Construction and Analysis}
\begin{itemize}
\item \textbf{MLDR Construction:} Describes the full pipeline for building the MLDR dataset.
\item \textbf{WeChat-Based Dataset Construction:} Details the data source, annotation protocol, and licensing for the WeChat test set.
\item \textbf{Dataset Statistics and Analysis:} Presents topic bias analysis and representative qualitative examples.
\end{itemize}

\item \textbf{Implementation Details}
\begin{itemize}
\item \textbf{Sliding Window Inference for Long Contexts:} Explains inference strategies for models that cannot process long dialogues directly under the single-dialogue FFR setting.
\item \textbf{Metrics of Fragment Boundary Prediction Model:} Provides a detailed explanation of the two clustering-based metrics, Pairwise Precision/Recall/F1 and B-cubed Precision/Recall/F1, used to evaluate the fragment boundary prediction model in FFRS for decomposing multi-modal dialogues.
\end{itemize}

\item \textbf{Experimental Results and Analysis}
\begin{itemize}
\item \textbf{Experimental Results in Single-Dialogue FFR:} Includes typical qualitative retrieval examples comparing different models, evaluates performance on an English-translated version of the WeChat test set, and defines the coverage, relevance, and consistency metrics used in the human evaluation.
\item \textbf{Qualitative Results of FFRS in Corpus-level FFR:} Presents qualitative results of the proposed FFRS system on the WeChat corpus.
\end{itemize}
\end{itemize}

\section{Dataset Construction and Analysis}

\subsection{MLDR Construction}\label{MLDR Construction}
This section provides detailed elaboration on the key steps involved in the construction of the MLDR dataset.

\noindent \textbf{Data Cleaning.}
To ensure high-quality long-form construction, we conduct a rigorous cleaning process based on four key filtering standards: (i) Image-Text Consistency: We encode visual and textual content utilizing CLIP and discard samples with low cross-modal similarity. (ii) Topic Coherence: Sentence embeddings are obtained utilizing BERT, and K-means clustering is applied to filter out dialogues that deviate from a central topic. (iii) Dialogue Turn Structure: Each dialogue must contain at least three turns in a ``User1–User2–User1" pattern to ensure basic conversational flow. (iv) Image Quality: We filter out images with resolutions below 500 pixels or with extreme aspect ratios (greater than 7.5), ensuring visual quality.

\begin{table*}[t]
\setlength{\abovecaptionskip}{1pt}
\setlength{\belowcaptionskip}{1pt}
\renewcommand\arraystretch{1.25}
\caption{Summary of main multimodal dialogue datasets. Datasets are categorized by modality (audio~(a), visual~(v), and textual~(t)), dialogue type (image/video-grounded vs. image-sharing), source, and language. Our MLDR dataset features significantly longer dialogues and richer topic transitions, better reflecting real-world human communication scenarios.} \label{tab:A1}
\centering
\resizebox{1.0\linewidth}{!}{
\begin{tabular}{c|cccc|ccccc} 
\toprule
\textbf{Datasets}                                      & \textbf{Modalites} & \textbf{Dialogue Type} & \textbf{Dialogue Source} & \textbf{Language} & \textbf{Dialogues} & \textbf{Images} & \textbf{Turns} & \textbf{Turn/Dialog} & \textbf{Topic/Dialog}  \\ 
\hline
ImageChat~\citep{shuster2020image}                                              & v, t               & image-grounded         & crowdsourcing            & English           & 201,779            & 201,779         & 400,853        & 1.98~                & 1.00~                  \\
OpenViDial~\citep{meng2020openvidial}                                             & v, t               & image-grounded         & moviesTVs                & English           & 1,100,000          & 1,100,000       & 1,100,000      & 1.00~                & 1.00~                  \\
PhotoChat~\citep{zang2021photochat}                                              & v, t               & image-sharing          & crowdsourcing            & English           & 11,820             & 10,479          & 150,138        & 12.74~               & 1.00~                  \\
MMDD~\citep{lee2021constructing}                                                   & v, t               & image-sharing          & text datasets            & English           & 17,679             & 13,288          & 187,421        & 11.56~               & 1.00~                  \\
MMDialog~\citep{feng2023mmdialog}                                               & v, t               & image-sharing          & social media             & English           & 1,079,117          & 1,556,868       & 4,920,000      & 4.56~                & 1.00~                  \\
DialogCC~\citep{lee2024dialogcc}                                               & v, t               & image-sharing          & text datasets            & English           & 83,209             & 129,802         & 676,181        & 8.20~                & 1.00~                  \\
\hline
MMChat~\citep{zheng2022mmchat}                                                 & v, t               & image-grounded         & social media             & Chinese           & 120,840            & 204,320         & 314,130        & 2.59~                & 1.00~                  \\

M3ED~\citep{zhao2022m3ed}                                                   & a, v, t            & video-grounded         & TVs                      & Chinese           & 990                & -               & 9082           & 9.17~                & 1.00~                  \\
CPED~\citep{chen2022cped}                                                   & a, v, t            & video-grounded         & TVs                      & Chinese           & 12,000             & -               & 133,000        & 11.08~               & 1.00~                  \\
CMMA~\citep{zhang2023cmma}                                                   & a, v, t            & video-grounded         & TVs                      & Chinese           & 3,000              & -               & 21,795         & 7.27~                & 1.00~                  \\
TikTalk~\citep{lin2023tiktalk}                                                & a, v, t            & video-grounded         & social media             & Chinese           & 367,670            & 38,703 videos   & 826,752        & 2.25~                & 1.00~                  \\ 
\hline
\rowcolor[rgb]{0.951,0.951,0.951} \textbf{Our Dataset} & v, t               & image-sharing          & text datasets            & Chinese           & 37,030             & 194,543         & 942,414        & \textbf{25.45}       & \textbf{3.00}          \\
\bottomrule
\end{tabular}}
\end{table*}

\begin{figure*}[htbp]
\centering
\includegraphics[width=1\linewidth]{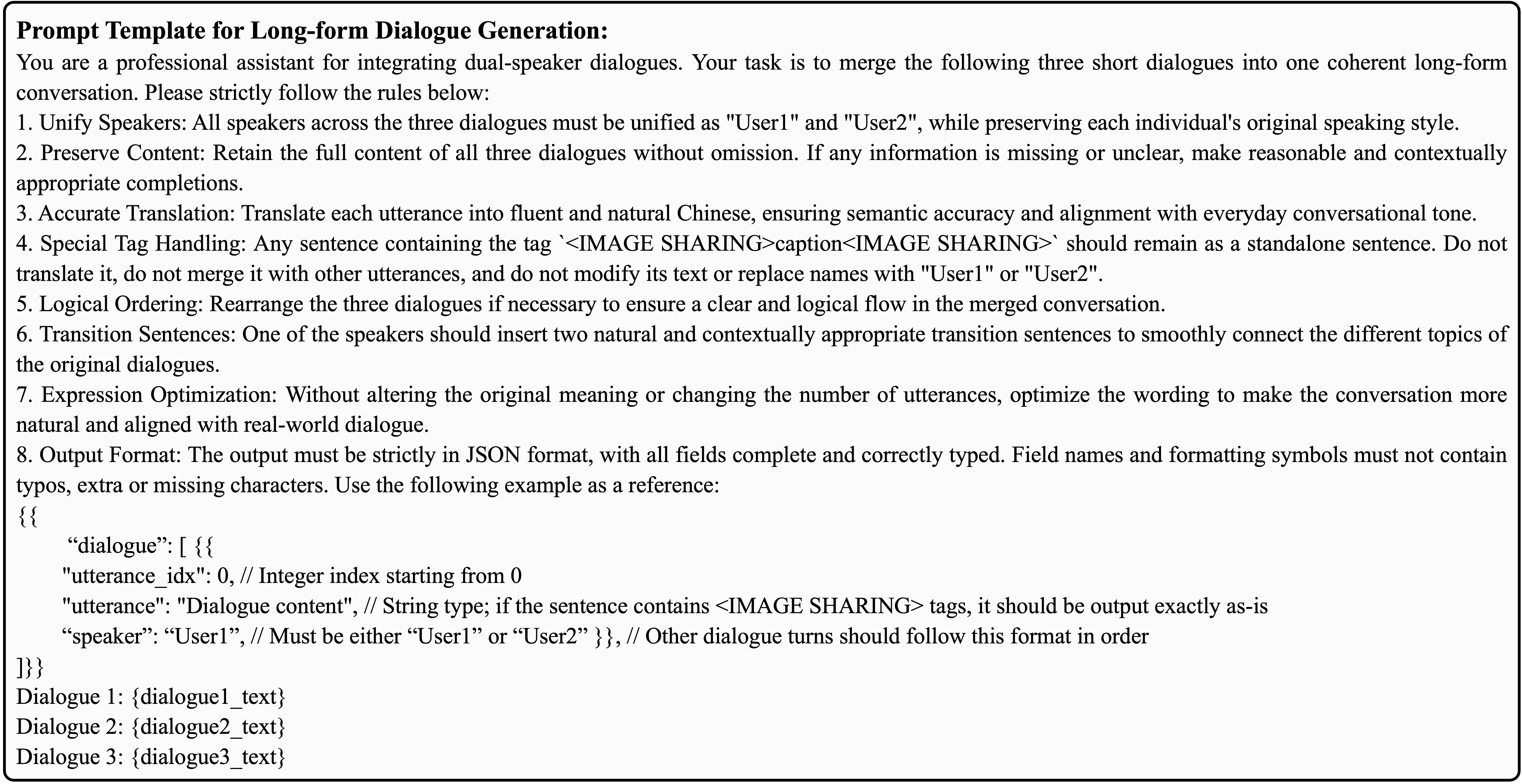}
\caption{\textbf{Prompt Template for Long-form Dialogue Generation.}}

\label{fig:A1}
\end{figure*}

\begin{figure*}[!ht]
\centering
\includegraphics[width=1.0\linewidth]{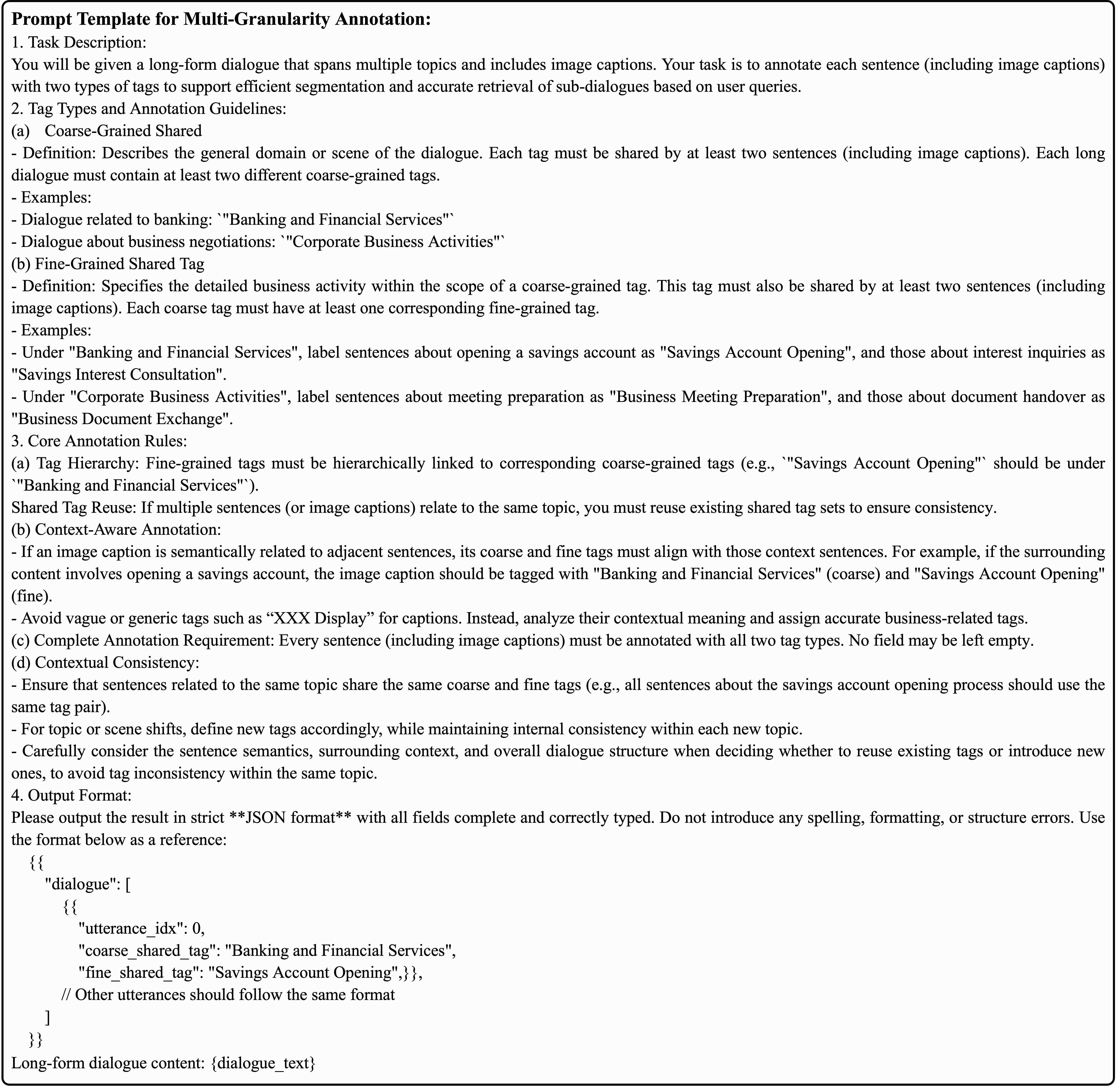}
\caption{\textbf{Prompt Template for Multi-Granularity Annotation.}}
\label{fig:A2}
\end{figure*}

\noindent \textbf{Dialogue Triplet Semantic Matching.}
To generate coherent long-form dialogues from multiple short dialogues, we propose a three-step triplet-based semantic matching strategy. This process ensures both topic continuity and multimodal alignment, and includes the following steps:
\begin{enumerate}
\item Trial Screening of Text Semantic: For each cleaned short dialogue \( D_A \), we first perform textual semantic screening. We encode all textual content (including image captions) using a pretrained BERT model and compute the mean sentence embedding for each dialogue. Cosine similarity is then calculated between \( D_A \)’s embedding and the embeddings of all other candidate dialogues. Based on this initial similarity score, we select the Top-\(K\) (where \(K = 50\)) most semantically relevant dialogues.
\item Refined Screening of Multimodal Semantic: Next, we perform multimodal semantic refinement for the Top-\(K\) dialogues selected in the first step. Using CLIP, we separately encode both textual and visual content, calculating the cosine similarity for each modality. These similarities are then combined with weighted scores (i.e., 0.7 for text and 0.3 for image) to obtain the final multimodal similarity score. From this, we select the Top-2 dialogues \( \{D_B^1, D_B^2\} \), based on this refined multimodal score.
\item Triplet Construction: For each selected dialogue \( D_B^i \) (\(i = 1, 2\)), we repeat the above two-stage matching process to retrieve the Top-2 most semantically and multimodally relevant dialogues, denoted as \( \{D_C^{i,1}, D_C^{i,2}\} \). This results in 4 distinct triplets: 
\begin{align} \label{equation:5}
D_A \rightarrow D_B^i \rightarrow D_C^{i,j}, \quad i = 1,2; \quad j = 1,2
\end{align}

These triplets provide a structural foundation for generating extended dialogues, ensuring both semantic and multimodal coherence. Notably, each dialogue \( D_A \), \( D_B \), and \( D_C \) in a triplet must be mutually disjoint in terms of both dialogue ID and image content, ensuring that each triplet represents unique and non-duplicate information.
\end{enumerate}

\noindent \textbf{Long-form Dialogue Generation.} 
To convert semantically aligned triplets into coherent long-form dialogues, we employ the Qwen3 256B~\citep{yang2025qwen3} large language model under a structured prompt template (Fig.\ref{fig:A1}). The generation process adheres to the following principles: (i) Preserve the original multimodal content and semantics while enhancing fluency and readability. (ii) Insert natural transition utterances to ensure coherent topic progression and stylistic consistency across dialogue segments. (iii) Translating all content into fluent and contextually appropriate Chinese, aligned with real-world conversational norms. This process enables the synthesis of high-quality, semantically coherent, and structurally complete Chinese multimodal long-form dialogues, which form the foundation for subsequent retrieval and understanding tasks.

\noindent \textbf{Multi-Granularity Annotation.} 
To enable fine-grained semantic retrieval over long-form multimodal dialogues, we design a two-level shared tagging scheme and implement it automatically using a Qwen3 model under structured prompts (see Fig.\ref{fig:A2}). Each sentence and image caption is annotated with (i) a coarse-grained shared tag indicating the high-level domain or scenario (e.g., “banking services”, “travel planning”), and (ii) a fine-grained shared tag specifying the detailed activity within the domain (e.g., “open savings account”). Both types of tags are required to be shared by at least two utterances or image captions to ensure topic consistency. To support high-quality tagging, Qwen3 is guided by explicit rules including tag reuse, topic-aligned grouping, and context-aware alignment across dialogue turns. This automated annotation process enables efficient semantic disentanglement and facilitates precise retrieval in complex multimodal conversations.

\subsection{WeChat-Based Dataset Construction}\label{WeChat-Based Dataset Construction}
\textbf{Data Source and Licensing Statement.}
The real-domain WeChat test set is constructed from naturally occurring image-text conversations voluntarily contributed by 12 participants. These dialogues are not extracted from any internal WeChat databases, but are shared with explicit informed consent from the contributors for academic research purposes. Prior to inclusion, all conversations were thoroughly anonymized and sanitized, with personal identifiers, sensitive content, and inappropriate language removed to ensure privacy protection and ethical compliance. To support responsible use, the dataset will be released under a research-only license. It is strictly prohibited to use the data for commercial applications, product development, or any non-academic purposes. Access to the dataset requires agreement to the accompanying license terms.

\noindent \textbf{Annotation Protocol and Quality Control.}
To ensure high-quality supervision for real-world fragment retrieval, we recruited 10 trained annotators to label the WeChat test set. Prior to annotation, all annotators received standardized training covering task definitions, multi-granularity labeling rules, and edge case handling. Each dialogue was independently annotated by all annotators to ensure consistency and comprehensive coverage. Following the initial round, we conducted rigorous quality checks to identify inconsistencies and labeling errors. Annotators were required to revise their annotations based on feedback. This iterative process continued until all annotations passed final validation, resulting in a reliable and well-curated benchmark for evaluating fine-grained multimodal retrieval.

\subsection{Dataset Statistics and Analysis}\label{Dataset Statistics and Analysis}
\textbf{Dataset Cases.}
To illustrate the characteristics of different evaluation scenarios, we present representative cases from both the MLDR and the WeChat test set (as reported in Fig.\ref{fig:A4}-\ref{fig:A6}). The MLDR set contains dialogues with fewer turns and a cleaner structure, while the WeChat test set features long, real-world conversations with many turns and mixed topics, posing greater challenges for retrieval models in maintaining relevance and coherence.

\noindent \textbf{Topic Distribution Bias in the WeChat Test Set.}
As illustrated in Fig.\ref{fig:2} in the main body, the WeChat test set exhibits a pronounced skew toward Work \& Tech topics, which account for nearly 50\% of all annotated dialogues. This bias can be primarily attributed to two factors. (i) The dataset is constructed from real-world conversations voluntarily contributed by 12 participants, most of whom are doctoral students or corporate employees. Their daily communication naturally revolves around professional and technical discussions, resulting in a topic distribution that overrepresents work-related content. (ii) In compliance with ethical and privacy considerations, we intentionally filtered out sensitive or highly private conversations (e.g., involving family, personal health, or intimate relationships). This further reduces the proportion of daily-life and emotion-related content.

We acknowledge that this topic bias represents a limitation of the current dataset. While it provides valuable insights into real-world retrieval performance under noisy and task-oriented scenarios, it may not fully capture the diversity of open-domain multimodal interactions. In future iterations, we plan to expand the contributor pool to include participants from more diverse demographics and occupations, and to adopt privacy-preserving mechanisms (e.g., local anonymization) that allow for the safe inclusion of a wider range of topics. This will support the construction of a more balanced and comprehensive real-domain benchmark.

\section{Implementation Details}

\subsection{Sliding Window Inference for Long Contexts} \label{Sliding Window Inference for Long Contexts}
Some open-source MLLMs (e.g., Ovis2-2B, LLaVA-1.5-7B, and InternVL3-series) are constrained by limited context length and cannot directly process long-form dialogues in a single pass. To address this limitation, we adopt a sliding window inference strategy during evaluation. Specifically, the dialogue is segmented into overlapping windows of 35 turns, with an overlap of 15 turns between adjacent windows. Each window is independently processed by the model, and the fragment-level predictions are collected. For overlapping segments, we take the union of predicted results across windows to ensure comprehensive coverage and mitigate boundary truncation effects. This approach enables fair evaluation of models with limited context capacity while maintaining retrieval completeness.

\subsection{Metrics of Fragment Boundary Prediction Model} \label{Metrics of Fragment Boundary Prediction Model}
We evaluate the quality of predicted dialogue fragmentations using two clustering-based metrics: Pairwise Precision/Recall/F1~\citep{shi2018face} and B-cubed Precision/Recall/F1~\citep{gunther2025jina}. Both metrics compute precision, recall, and F1-score, which are harmonic means of precision and recall. The key difference between them lies in the granularity of the evaluation.

\noindent \textbf{Pairwise Precision/Recall/F1.} 
This metric evaluates the overall consistency by considering all possible utterance pairs in the dialogue. Let $R(i,j) = 1$ if utterances $i$ and $j$ belong to the same ground-truth fragment, and $\hat{R}(i,j) = 1$ if they are predicted to belong to the same fragment. From this, we compute true positives (TP), false positives (FP), and false negatives (FN), and derive the precision, recall, and F1 scores:
\begin{equation}
\begin{aligned}
 &   \text{TP} = \sum_{i<j}[R(i,j) = 1 \wedge \hat{R}(i,j) = 1] \\
 &   \text{FP} = \sum_{i<j}[R(i,j) = 0 \wedge \hat{R}(i,j) = 1] \\
 &   \text{FN} = \sum_{i<j}[R(i,j) = 1 \wedge \hat{R}(i,j) = 0] \\
\end{aligned}
\end{equation}
The final precision, recall, and F1 scores are computed as:
\begin{equation}
\begin{aligned}
& \text{Precision}_{\text{pair}} = \frac{\text{TP}}{\text{TP} + \text{FP}}, \quad
\text{Recall}_{\text{pair}} = \frac{\text{TP}}{\text{TP} + \text{FN}} \\
& \text{F1}_{\text{pair}} = \frac{2 \cdot \text{Precision}_{\text{pair}} \cdot \text{Recall}_{\text{pair}}}{\text{Precision}_{\text{pair}} + \text{Recall}_{\text{pair}}}
\end{aligned}
\end{equation}

\noindent \textbf{B-cubed Precision/Recall/F1.} 
This metric measures local consistency for each utterance. Let $C(i)$ and $\hat{C}(i)$ denote the ground-truth and predicted clusters for utterance $i$, respectively:
\begin{equation}
\begin{aligned}
    P_i = \frac{|C(i) \cap \hat{C}(i)|}{|\hat{C}(i)|}, \quad
    R_i = \frac{|C(i) \cap \hat{C}(i)|}{|C(i)|}.
\end{aligned}
\end{equation}
The B-cubed precision, recall, and F1 scores are averaged over all utterances in the dialogue:
\begin{equation}
\begin{aligned}
&    \text{Precision}_B = \frac{1}{N} \sum_i P_i, \quad
\text{Recall}_B = \frac{1}{N} \sum_i R_i \\
&    \text{F1}_B = \frac{2 \cdot \text{Precision}_B \cdot \text{Recall}_B}{\text{Precision}_B + \text{Recall}_B}
\end{aligned}
\end{equation}
This metric evaluates local consistency by ensuring that each utterance is correctly grouped with its corresponding cluster, both in ground-truth and predicted groupings.

\begin{table*}[t]
\renewcommand\arraystretch{1.25}
\setlength{\abovecaptionskip}{1pt}
\setlength{\belowcaptionskip}{1pt}
\centering
\caption{Evaluation results on the English-translated WeChat test set. We report precision, recall, F1, and MCC for utterance, image, and joint retrieval. ``$\dagger$'' indicates zero-shot inference without MLDR fine-tuning.}
\resizebox{1.0\linewidth}{!}{
\begin{tabular}{r|cccc|cccc|cccc} 
\toprule
\multirow{2}{*}{Model}                                                  & \multicolumn{4}{c|}{Utterance Retrieval}                          & \multicolumn{4}{c|}{Image Retrieval}                              & \multicolumn{4}{c}{Joint Retrieval}                                \\ 
\cline{2-13}
& Precision      & Recall         & F1             & MCC            & Precision      & Recall         & F1             & MCC            & Precision      & Recall         & F1             & MCC             \\ 
\hline
Qwen2.5-VL-72B$^\dagger$                                                          & 30.31          & 32.04          & 31.15          & 21.57          & 34.23          & \underline{73.56}  & 46.72          & 36.60          & 32.15          & 44.64          & 38.94          & 29.08           \\
Claude-4$^\dagger$                                                                 & \underline{59.09}  & 39.46          & \underline{47.32}          & \underline{42.82}  & 59.05          & 59.62          & \underline{59.33}  & \underline{51.83}  & \textbf{59.07} & 47.49          & \underline{53.32}  & \underline{47.32}   \\
GPT-4o$^\dagger$                                                                   & 64.18          & 34.54          & 44.91          & 42.23          & 53.69          & 62.98          & 57.96          & 49.74          & \underline{58.47}  & 44.61          & 51.43          & 45.99           \\
Gemini-2.5$^\dagger$                                                               & \textbf{59.94} & 26.56          & 36.81          & 35.00          & 52.25          & 47.21          & 49.60          & 41.13          & 55.83          & 33.99          & 43.20          & 38.06           \\ 
\hline
mPLUG-Owl3-2B                                                                 & 22.32          & 16.92          & 19.25          & 10.14          & 16.71          & 29.81          & 21.42          & 5.13           & 19.11          & 21.59          & 20.33          & 7.64            \\
Qwen2-VL-2B                                                            & 15.90          & \textbf{86.34} & 26.86          & 16.69          & 27.57          & \textbf{81.25} & 41.17          & 30.58          & 20.17          & \textbf{83.72} & 34.01          & 23.64           \\
Qwen2.5-VL-3B                                                          & 23.78          & 76.42          & 36.27          & 28.87          & 34.47          & 72.60          & 46.75          & 36.50          & 28.15          & 74.46          & 41.51          & 32.68           \\
Mimo-7B-RL                                                           & 37.44          & 40.92          & 39.10          & 30.49          & \textbf{64.75} & 37.98          & 47.88          & 43.04          & 47.44          & 39.39          & 43.49          & 36.77           \\
Qwen2-VL-7B                                                            & 38.85          & 56.66          & 46.09          & 38.14          & 47.47          & 58.65          & 52.47          & 43.00          & 42.73          & 57.64          & 49.28          & 40.57           \\
Qwen2.5-VL-7B                                                          & 34.82          & 63.52          & 44.98          & 37.27          & 44.90          & 67.79          & 54.02          & 44.91          & 39.22          & 65.59          & 49.50          & 41.09           \\
DeepSeek-VL2-Small-16B                                                   & 26.43          & 14.66          & 18.86          & 11.97          & 55.88          & 9.69           & 16.52          & 18.72          & 35.89          & 11.67          & 17.69          & 15.35           \\
\rowcolor[rgb]{0.951,0.951,0.951} \textbf{\textbf{F$^2$RVLM-Qwen2-VL-2B}}   & 17.03          & \underline{86.27}  & 28.45          & 19.51          & 29.80          & 78.37          & 43.18          & 32.78          & 21.68          & \underline{82.13}  & 35.81          & 26.14           \\
\rowcolor[rgb]{0.951,0.951,0.951} \textbf{\textbf{F$^2$RVLM-Qwen2.5-VL-3B}} & 30.25          & 70.18          & 42.28          & 35.00          & 42.11          & 73.08          & 53.43          & 44.60          & 35.21          & 71.60          & 47.85          & 39.80           \\
\rowcolor[rgb]{0.951,0.951,0.951} \textbf{F$^2$RVLM-Qwen2-VL-7B}            & 50.19          & 53.81          & \textbf{51.94} & \textbf{45.23} & \underline{62.15}  & 63.94          & \textbf{63.03} & \textbf{56.15} & 55.54          & 58.44          & \textbf{57.49} & \textbf{50.69}  \\
\bottomrule
\end{tabular}
}
\label{tab:A4}
\end{table*}

\section{Experimental Results and Analysis}
\subsection{Experimental Results in Single-Dialogue FFR} \label{Experimental Results in Single-Dialogue FFR}

\begin{figure*}[!ht]
\centering
\includegraphics[width=0.85\linewidth]{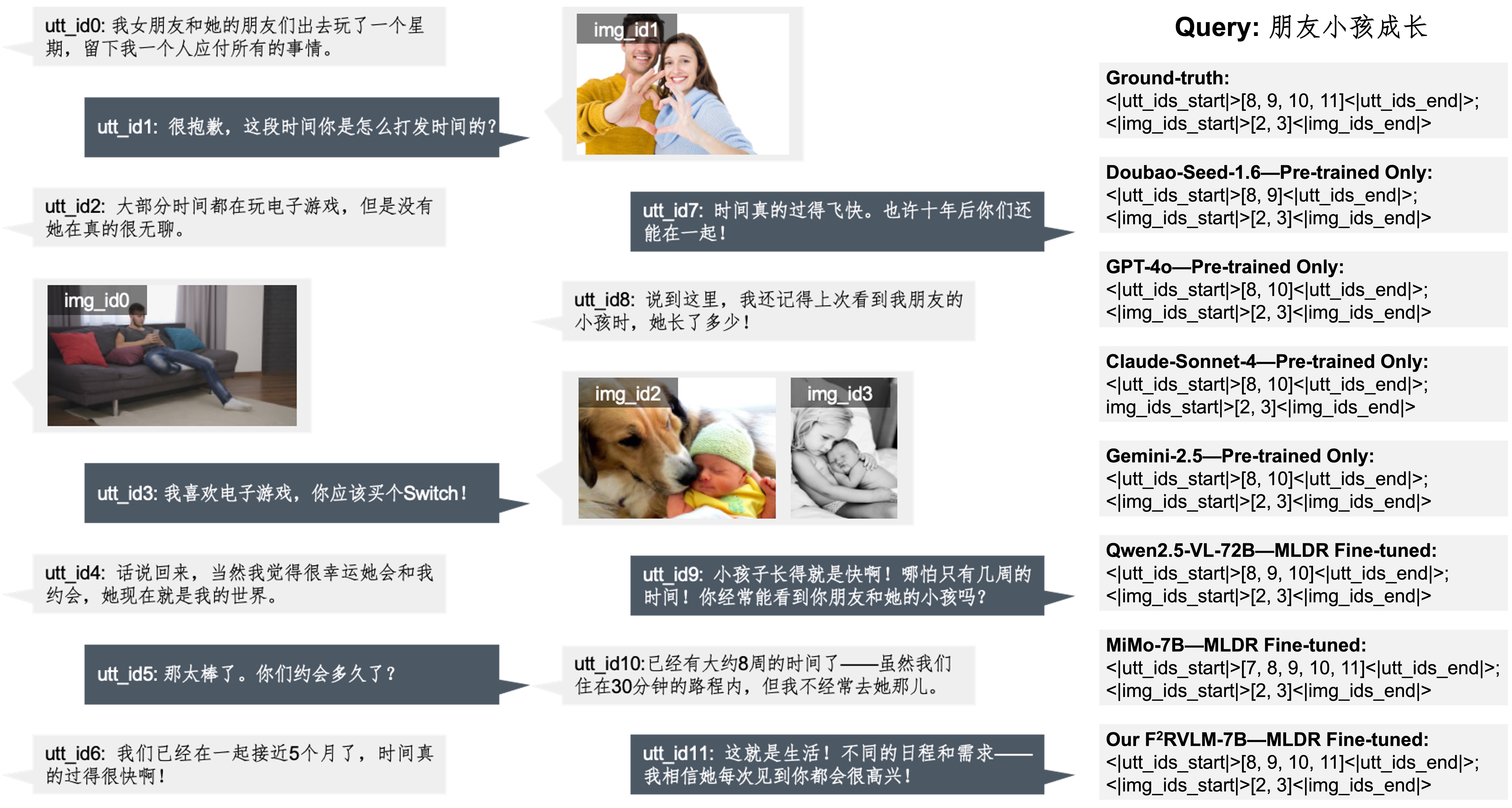}
\caption{Qualitative comparison on the MLDR validation set. Given a user query, we visualize one representative case of retrieved fragments from various models. Pre-trained models often retrieve semantically relevant but incomplete or disjointed fragments. MLDR fine-tuned models improve alignment but may still miss contextual boundaries. Our F$^2$RVLM-7B achieves the most coherent and complete retrieval, accurately aligning both utterances and images with the intended semantics.}
\label{fig:A4}
\end{figure*}

\begin{figure*}[!ht]
\centering
\includegraphics[width=1.0\linewidth]{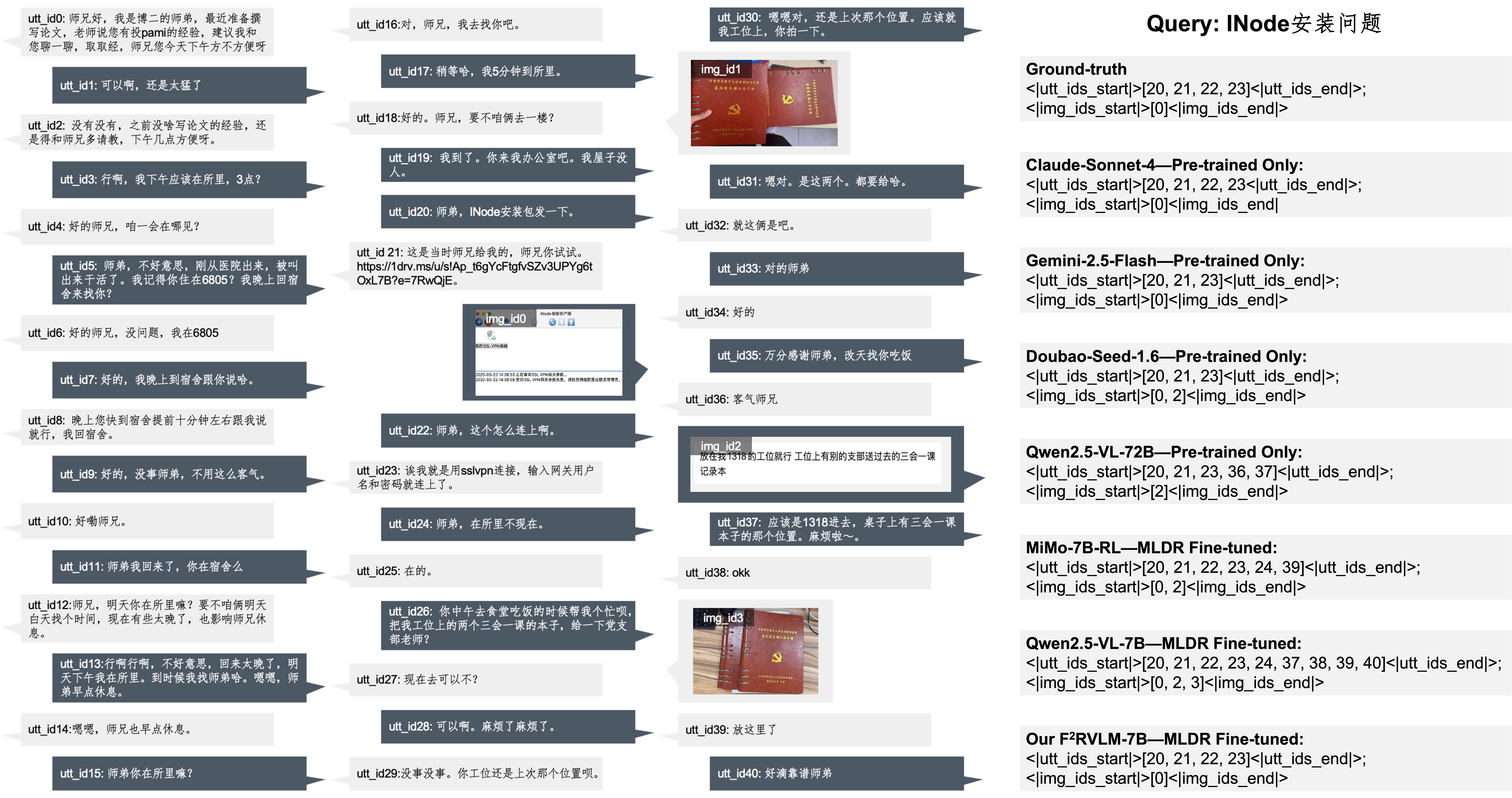}
\vspace{-15pt}  
\caption{Case 1 of qualitative comparison on the WeChat test set, focusing on a personal-life dialogue scenario.}
\label{fig:A5}
\end{figure*}

\begin{figure*}[!ht]
\centering
\includegraphics[width=1.0\linewidth]{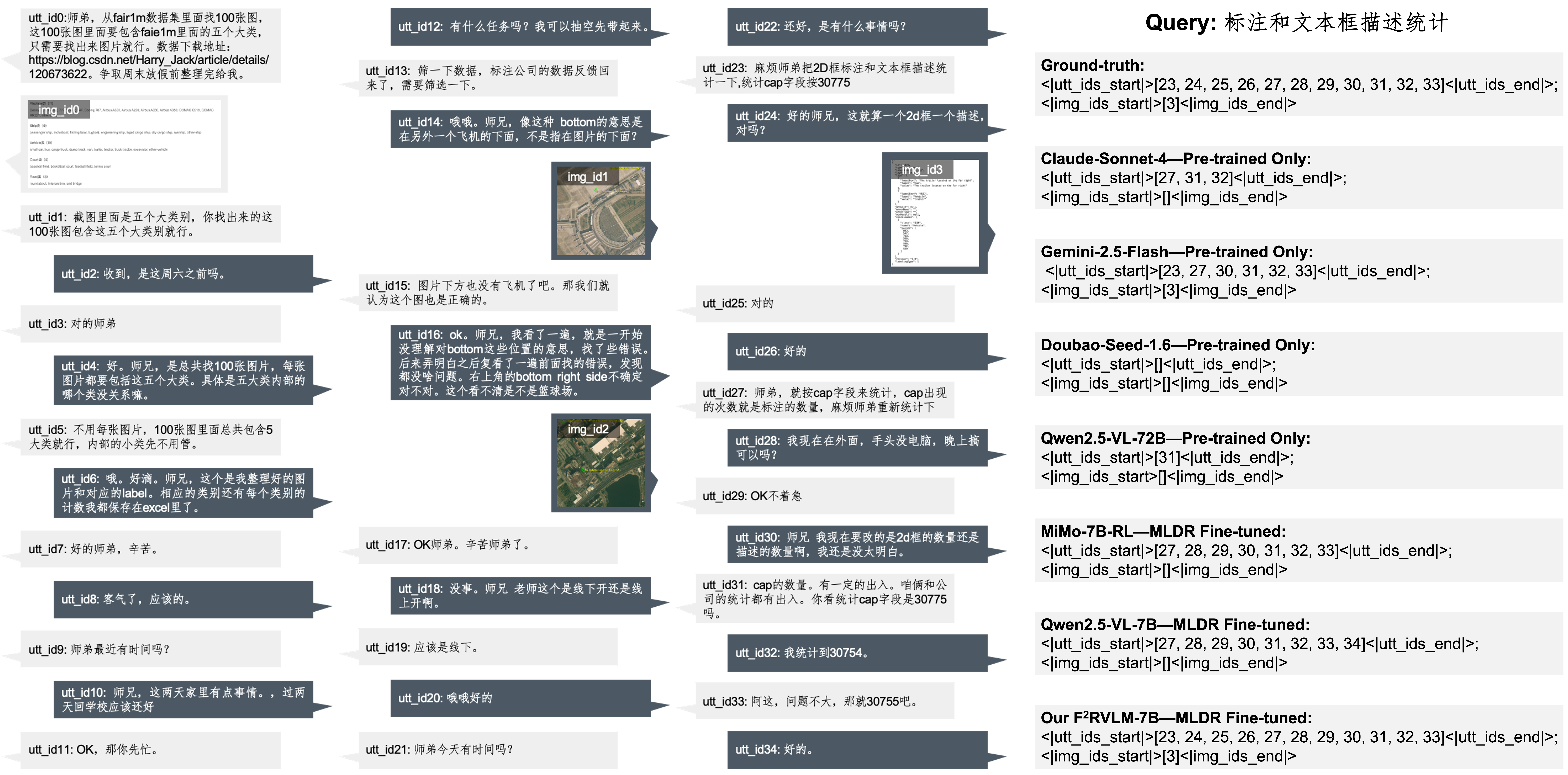}
\vspace{-15pt}  
\caption{Case 2 of qualitative comparison on the WeChat test set, illustrating fragment retrieval in a technical-support dialogue scenario.}
\label{fig:A6}
\end{figure*}

\noindent \textbf{Comparison of Qualitative Results.}
We provide qualitative comparisons on the MLDR validation set and the WeChat test set to illustrate fragment-level retrieval performance. On the MLDR validation set (Fig.\ref{fig:A4}), our F$^2$RVLM accurately retrieves semantically complete and coherent fragments that align well with human annotations. It outperforms closed-source or open-source large models that often miss key utterances or include off-topic content. On the WeChat test set (Fig.\ref{fig:A5}–\ref{fig:A6}), our model demonstrates better robustness in informal, multi-topic dialogues. While closed-source and open-source models tend to produce partial or noisy results, F$^2$RVLM captures more complete, relevant, and coherent fragments, effectively combining verbal and visual cues even in complex real-world cases. These results further validate the model’s superior capability in real-world fragment-level retrieval.

\noindent \textbf{Comparison of Cross-Lingual Generalization.}
To assess the cross-lingual robustness, we evaluate various VLMs on an English-translated version of the WeChat test set. Specifically, we randomly sample 200 dialogues from the original set and translate each utterance into English using DeepL Translator to preserve semantic fidelity. The translated set maintains the original structure and multimodal content, enabling fair comparison across languages. As depicted in Table~\ref{tab:A4}, even though models like Claude-4 and GPT-4o exhibit strong zero-shot performance due to extensive pre-training on English corpora, our F$^2$RVLM-7B achieves the highest overall joint F1 (57.49\%) and MCC (50.69\%). Even the 3B variant of F$^2$RVLM outperforms larger models like MiMo-7B and Qwen2.5-VL-7B, confirming its strong generalization with fewer parameters. Importantly, unlike some VLMs that exhibit skewed retrieval behavior (e.g., high precision but low recall, or vice versa), F$^2$RVLM achieves a more balanced trade-off between precision and recall. This balance results in consistently higher F1 and MCC scores, highlighting the model’s ability to perform fine-grained fragment retrieval across languages. The above results suggest that the reward-guided fine-tuning strategy in F$^2$RVLM effectively transfers to different linguistic settings, supporting fine-grained fragment alignment and reasoning beyond Chinese-language contexts.

\noindent \textbf{Human Subjective Evaluation Criteria.}
To assess the human-perceived quality of fragment-level retrieval, we randomly sample 200 dialogues and ask expert annotators to compare model outputs under identical input contexts. Each annotator is instructed to select the best-performing model according to the following three criteria:
\begin{enumerate}
    \item \textbf{Coverage,} which measures whether the retrieved fragment fully captures all key information relevant to the user query. It emphasizes completeness, particularly the inclusion of essential contextual elements. Fragments that contain more content can still receive high coverage scores as long as they thoroughly address the query without omitting important details.
    \item \textbf{Relevance,} which evaluates how well the retrieved content semantically aligns with the user query. Every utterance and image in the fragment should directly relate to the query’s core intent. Even if some parts are on-topic, the presence of off-topic or irrelevant content will reduce the relevance score.
    \item \textbf{Coherence,} which assesses the logical flow and clarity of the retrieved fragment itself. It focuses on whether the sentences and images are naturally connected, well-organized, and easy to understand when read as a self-contained unit. This evaluation does not consider the broader dialogue context, but strictly the internal consistency of the retrieved segment.
\end{enumerate}
If multiple models produce identical and top-performing outputs for a given dialogue, annotators will select all such models. As a result, the total number of top-choice selections may exceed the number of dialogues evaluated.

\subsection{Qualitative Results of FFRS in Corpus-level FFR} \label{Qualitative Results of FFRS in Corpus-level FFR}
Beyond quantitative metrics, qualitative inspection further confirms the effectiveness of the proposed system. As showcased in our demo\footnote{\url{https://sprproxy-1258344707.cos.ap-shanghai.myqcloud.com/hanbobi/Pipeline_demo.html}}, the fragments retrieved by FFRS exhibit clear semantic alignment with user-issued queries, accurately capturing both the explicit intent and the underlying contextual cues. These examples highlight that FFRS consistently locates the most relevant portions of long multi-modal conversations, even when the supporting evidence is dispersed across multiple turns. Taken together, the qualitative findings underscore that FFRS not only achieves substantial efficiency improvements but also preserves high retrieval fidelity across large-scale, real-world multi-modal dialogue corpora.

\end{sloppypar}  
\end{document}